\documentclass[runningheads]{llncs}

\usepackage{eccv}

\usepackage{graphicx}
\usepackage{tikz}
\usetikzlibrary{calc}
\usepackage{comment}
\usepackage{amsmath,amssymb} 
\usepackage{color}
\usepackage{xcolor, colortbl}
\usepackage{xspace}
\usepackage{booktabs}
\usepackage{multirow}
\usepackage{pdfpages}
\usepackage{wrapfig}
\usepackage[normalem]{ulem} 
\usepackage{url}

\usepackage{hyperref}

\usepackage{enumitem}  
\usepackage{sidecap} 
\usepackage{orcidlink}

\usepackage[accsupp]{axessibility}  

\usepackage{eccvabbrv}

\renewcommand{\paragraph}[1]{\vspace{0pt}\noindent\textbf{#1}\hspace{5pt}}

\newcounter{ablationrow}
\renewcommand{\theablationrow}{\Alph{ablationrow}}
\newcommand{\row}[1]{\refstepcounter{ablationrow}\label{#1}\theablationrow}

\definecolor{firstcolor}{rgb}{1, 0.6, 0.6}
\definecolor{secondcolor}{rgb}{1, 0.8, 0.6}
\definecolor{thirdcolor}{rgb}{1,1, 0.6}

\definecolor{revisit}{RGB}{0, 121, 255}
\definecolor{incremental}{RGB}{0, 190, 62}
\definecolor{offline}{RGB}{255, 0, 96}

\newcommand{\incremental}{\hyperref[sec:sources_of_geometry]{\textcolor{incremental}{{incremental}}}\@\xspace}
\newcommand{\incrementally}{\hyperref[sec:sources_of_geometry]{\textcolor{incremental}{{incrementally}}}\@\xspace}
\newcommand{\revisit}{\hyperref[sec:sources_of_geometry]{\textcolor{revisit}{revisit}}\@\xspace}
\newcommand{\offline}{\hyperref[sec:sources_of_geometry]{\textcolor{offline}{offline}}\@\xspace}

\definecolor{bluearrow}{RGB}{10, 40, 255}

\definecolor{ms_note}{RGB}{0, 181, 190}
\definecolor{burgundy}{RGB}{150, 0, 0}
\definecolor{pink}{rgb}{1,0.0,1.0}

\begin{document}

\pagestyle{headings}
\mainmatter

\title{DoubleTake: \\Geometry Guided Depth Estimation} 

\titlerunning{DoubleTake: Geometry Guided Depth Estimation}
%
\newcommand{\gap}{\hspace{10pt}}
\author{Mohamed Sayed\inst{1}\orcidlink{0000-0002-4074-3314} \gap Filippo Aleotti\inst{1}\orcidlink{0000-0002-8911-3241} \gap Jamie Watson\inst{1,2}\orcidlink{0000-0001-7461-5663
} \\ 
Zawar Qureshi\inst{1}\orcidlink{0009-0003-9917-4827} \gap Guillermo Garcia-Hernando\inst{1}\orcidlink{0000-0003-3215-7857} \gap 
 Gabriel Brostow\inst{1,2}\orcidlink{0000-0001-8472-3828}  \\ Sara Vicente\inst{1}\orcidlink{0009-0001-3108-5030} \gap Michael Firman\inst{1}\orcidlink{0000-0002-0833-2021}
}

\authorrunning{M.~Sayed et al.}
%

\institute{ $^{1}$Niantic \hspace{30pt}$^{2}$UCL \\ \url{https://nianticlabs.github.io/doubletake/}}
\maketitle

\begin{abstract}

Estimating depth from a sequence of posed RGB images is a fundamental computer vision task, with applications in augmented reality, path planning etc. 
Prior work typically makes use of previous frames in a multi view stereo framework,
relying on matching textures in a local neighborhood. 
In contrast, our model leverages historical predictions by giving the latest 3D geometry data as an extra input to our network. 
This self-generated geometric hint can encode information from areas of the scene not covered by the keyframes and it is more regularized when compared to individual predicted depth maps for previous frames. 
We introduce a \emph{Hint MLP} which combines cost volume features with a hint of the prior geometry, rendered as a depth map from the current camera location, together with a measure of the confidence in the prior geometry.
We demonstrate that our method, which can run at interactive speeds, achieves state-of-the-art estimates of depth and 3D scene reconstruction in both offline and incremental evaluation scenarios.


\end{abstract}

\section{Introduction}

\begin{figure}[t]
    \centering
    \includegraphics[width=0.95\linewidth]{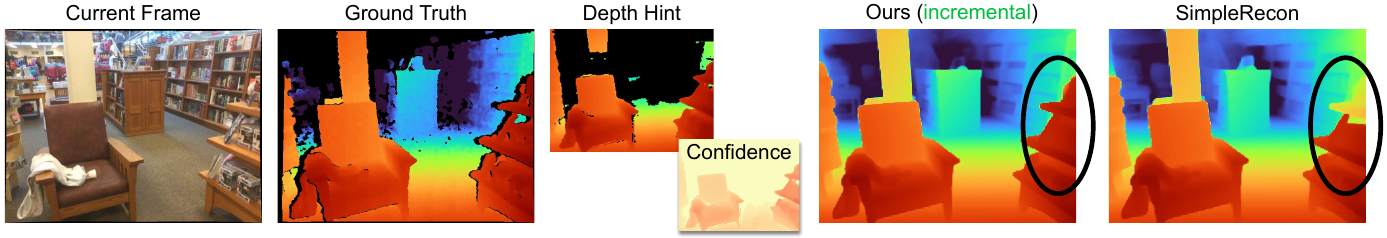}
    \caption{
        A depth hint and confidence are rendered from a prior estimate of geometry and given as input to our method. This enables it to correctly predict the depth for ambiguous parts of the scene.
    }
    \label{fig:teaser}
\end{figure}

High quality depth estimations have been shown to be effective for virtual occlusions \cite{valentin2018depth}, path planning \cite{du2020depthlab}, object avoidance and a variety of augmented reality (AR) effects \cite{du2020depthlab}.
For all these applications, we need per-frame depths to be generated at interactive speeds.
While the best quality depth maps can be achieved via offline methods \eg \cite{izquierdo2023sfm,yariv2021volume}, these are not suitable for interactive applications. 
For interactive use, depths are typically estimated via a multi-view stereo (MVS) approach, where a network takes as input a target frame together with nearby posed and matchable source frames to build a cost volume \cite{kendall2017end} and estimate a depth map as the output of a neural network \cite{sayed2022simplerecon,chang2018pyramid,Zhang2019GANet,zhang2019domaininvariant,cheng2019learning}.

Matchability needs the texture on the visible surfaces to be similar and visible in both source and target frames, though this is often not possible \eg due to occlusions or distance from the surface. 
We make the case that each time we estimate depth for a location, it is unlikely that this is the first time we have seen this place. 
We might revisit a location in the short term, for example turning to look at a kitchen appliance we were last looking at a few seconds ago.
Alternatively we may revisit a location after a period of time has passed, for example entering a room we last visited a week prior. 
In this work, we argue that these short and long-term geometric `snapshots' of the scene can be a crucial source of information which can be used to gain higher quality instantaneous depths. 
We introduce a system that maintains a low-cost global representation of 3D geometry as a truncated signed distance function (TSDF). 
When predicting depth for a new frame, we render a depth map from the TSDF at the current camera pose and give that as input to our depth estimator (see Figure~\ref{fig:teaser}).

Our experiments show that, surprisingly, just naively passing in a depth render of a global mesh into a depth estimation network fails to bring performance gains. 
Instead, we introduce a carefully-designed architecture to make use of such prior geometry `hints' as input, and show that the same network can make use of these hints both from the short and long term.
We validate our approach via extensive experiments and ablations, and show that our method achieves a new state-of-the-art on depth estimation and reconstruction, as validated on the challenging ScanNetV2, 7Scenes, and 3RScan datasets.
Our contributions are: 
\begin{enumerate}
    \item 
        A system which can use prior estimations of geometry to improve instantaneous depth estimates.
        If a hint isn't available for a frame, our system gracefully falls back to the performance of a strong baseline model.
        We also extend this to include \emph{historical}, long-term hints into our framework, enabling the use of hints when revisiting a location after a period of time. 
    \item 
        As geometry is constructed in real-time, certain parts of it may still be unreliable. 
        We therefore propose an architecture which incorporates this geometry alongside a measure of its confidence. 
        This combined information is integrated with multi-view stereo cost volume data using a `Hint MLP'.
    \item 
        We introduce a new evaluation protocol that pays more careful consideration to the limitations of the ground-truth mesh, and evaluate ours and several baselines with this new evaluation.
\end{enumerate}

\section{Related Work}
\label{sec:related_work}

\paragraph{Multi-view Stereo.} 
Depth estimation from posed monocular videos is a long-standing problem in computer vision. 
Traditional patch-based solutions \cite{furukawa2015multi,schonberger2016pixelwise} have been outpaced by learning strategies \cite{yao2018mvsnet,huang2018deepmvs}, where a cost volume is built by warping features from multiple source frames at different depth hypotheses \cite{collins1996space}. 
The resulting 4D volume can be regularized by 3D convolutions \cite{yao2018mvsnet}, which are expensive in terms of memory. 
To tackle this, pyramidal approaches \cite{yang2020cost,gu2020cascade} compute multiple cost volumes at different resolutions with a narrow set of hypotheses, and depths computed at lower scales can provide geometric priors to following computation \cite{zhang2023geomvsnet,cai2023riav}.
SimpleRecon \cite{sayed2022simplerecon} shows that accurate depths can be achieved only via 2D convolutions, using cost volumes enriched by additional \emph{metadata}, while FineRecon \cite{Stier_2023_ICCV} further improves the results with a resolution-agnostic 3D training loss and a depth guidance strategy.
These multi-view stereo (MVS) approaches have several failure cases: (i) unmatched surfaces, 
\eg empty areas in the cost volume where no source frames have a matching view; (ii) depth planes are sparser at greater depths; 
(iii) greater depths require wider baselines and many more keyframes. 
Our proposed approach helps overcome these problems by injecting prior information into the network based on previously computed geometry data.



\paragraph{Use of Input Sequences.} For example,
\cite{Stereo-Matching-in-cheng-2023,Raft-stereo-Multilevel-lipson-2021,song2023prior} warp previous predictions or features and use these as input to the current prediction, in some cases together with a confidence estimate \cite{song2023prior}.
We also use an estimate of confidence, but our geometry priors are from a global reconstruction, enabling priors from long before in time.
Instead of regularizing the cost volume, CER-MVS \cite{ma2022multiview} iteratively updates a disparity field using recurrent units. 
Recurrent layers \cite{duzceker2021deepvideomvs} and Gaussian Processes \cite{hou2019multi} have also been used to ingest sequence data.
Our work uses sequences but we do not rely on generic layers to encode the prior information. 
Instead, we directly extract prior knowledge from a 3D reconstruction of the scene to guide the model.
An alternative use of sequences is to update predicted depth maps or network weights at \emph{test time} by optimizing image reconstruction losses \eg~\cite{casser2018depth,chen2019self,luo2020consistent,mccraith2020monocular,shu2020feature,kuznietsov2021comoda,izquierdo2023sfm} .
These methods require the entire sequence to be seen ahead of time, and aren't suitable for interactive applications.

\paragraph{Priors for Depth Estimation.} 
When available, additional knowledge about the scene might be injected to boost depth estimation methods. 
In autonomous driving \cite{geiger2012we,Menze2015CVPR}, for instance, LiDAR scans are often completed into a dense depth map \cite{ma2018sparse,ma2019self,Depth-Completion-using-Du-2022,choe2021volumetric}. 
Similarly, sparse point clouds from Simultaneous Localization And Mapping (SLAM) algorithms can be used as input to improve depth estimators \cite{wong2021unsupervised,Xin2023ISMAR}. 
However, these priors might be unevenly scattered or unavailable.
To tackle these issues, sparsity-agnostic methods \cite{uhrig2017sparsity,conti2023sparsity} and networks capable of both predicting and completing depth maps \cite{guizilini2021sparse} have been proposed. 
mvgMVS \cite{poggi2022multi} uses depth data from a depth sensor to directly modulate values in the cost volume. 
In contrast, our approach uses prior geometry generated from our own estimates as an additional input to an MLP alongside the cost volume.
Khan \etal~\cite{khan2023temporally} convert per-frame depth estimates to temporally consistent depth via a globally fused point cloud which is input into their final depth network.
Their focus is primarily temporal consistency; we compare with \cite{khan2023temporally} and find our approach achieves higher quality depths.
Like us, 3DVNet~\cite{rich20213dvnet} maintain a global reconstruction to improve depth maps. 
However, their reconstruction and depths are iteratively refined, precluding online use.
We also employ priors to improve current depth predictions.
Unlike these prior works, we argue that previous predictions of the model itself, if properly managed, are strong hints to improve current estimates for online depth estimation. 
Furthermore, we also show that our model is robust even in the absence of such information.



\paragraph{3D Scene Reconstruction.} 
Traditional methods for 3D reconstruction estimate depths (\eg from MVS), fuse them into a TSDF and extract a mesh via Marching Cubes \cite{lorensen1998marching}. 
In contrast, feedforward volumetric methods \cite{murez2020atlas,sun2021neuralrecon,gao2023visfusion,zuo2023incremental} directly estimate the volume occupancy (usually encoded as a TSDF), often leveraging expensive 3D CNNs. 
Implicit representations \cite{peng2020convolutional,yu2022monosdf,li2023neuralangelo} can generate high quality reconstructions via test-time optimization. These solutions are computationally expensive as they typically require a \textit{per-scene} optimization. 
Neural Radiance Fields (NeRF) \cite{mildenhall2020nerf} and Gaussian Splatting \cite{kerbl20233d} enable novel view synthesis, but their meshes tend to be noisy and additional post-processing steps \cite{rakotosaona2023nerfmeshing,guedon2023sugar} or a different scene representation \cite{Kulhanek_2023_ICCV} might be necessary to generate consistent 3D representations. 
Extensions to NeRFs use structure-from-motion point clouds and depth estimation to improve view synthesis~\cite{roessle2022dense,kangle2021dsnerf,scade-nerfs-from-Uy-2023} or reconstructions \cite{fu2022geo,yu2022monosdf}.
These offline approaches require the whole video to be seen ahead of time and are typically slow, making them unsuitable for interactive applications.
Conversely, our method efficiently estimates depth maps using a 2D MVS model boosted by its own predictions, unlocking accurate reconstructions and depth estimates with a low overhead in computation. 






\section{Method}

Our method takes as input a live online sequence of RGB images along with their poses and intrinsics. The goal of our method is to predict a depth map for the current frame given all frames that have come before it. To train our method we assume we have access to a ground truth depth map for each training image.


\begin{figure}[t]
    \centering
    \includegraphics[width=0.98\linewidth]{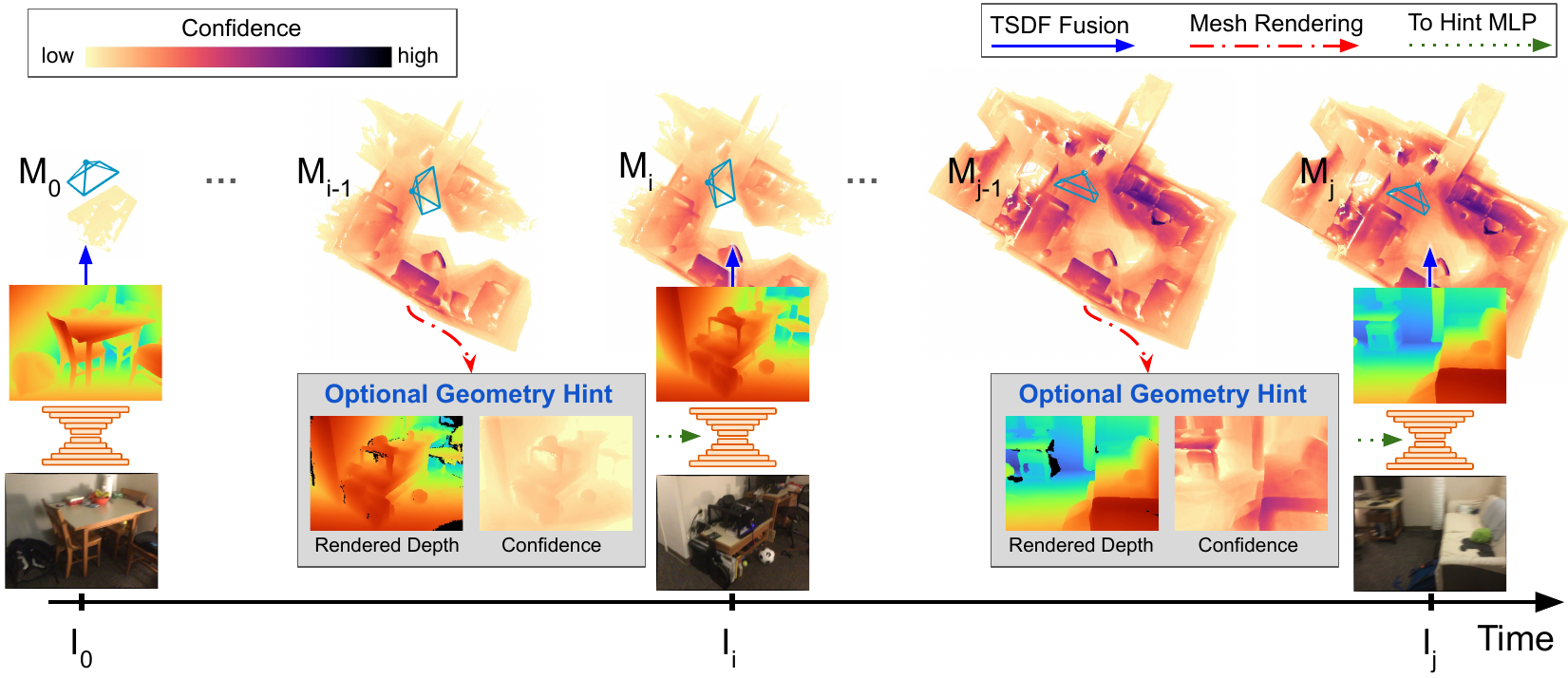}
    \caption{
        \textbf{Method overview.} 
            Like other MVS methods, we take as input a sequence of RGB frames.
            Optionally, our network can additionally ingest a depth map rendered from the 3D representation of the scene built up so far, encoded in a TSDF volume. 
            Alongside the rendered depth map, we include an estimate of how confident the global geometry is at each point, visualized as vertex colors (purple is higher confidence) on the \incremental mesh here.
            The depth predicted by our model is fused back into the TSDF to update the 3D geometry incrementally. 
            When no such rendered depth map is available, our network gracefully falls back to our baseline model's performance.
    }
    \label{fig:method-overview}
\end{figure}

Similar to previous MVS methods \eg \cite{yao2018mvsnet,sayed2022simplerecon}, we rely on features matched between the current frame and a few recent source frames to build a feature-based cost volume. 
Our key contribution is to show how a 3D representation of the scene, constructed from previous depth estimates, can be used as additional input to our network to improve depth predictions for each new frame. 
In particular, this geometric `hint' complements the likelihood depth estimated via multi-view-stereo on the current and source frames.
We incorporate this extra source of information by combining it with the information provided by the cost volume using a ``Hint MLP.''
An overview of our method can be seen in Fig.~\ref{fig:method-overview}.

\subsection{MVS background}

Our method builds on SimpleRecon~\cite{sayed2022simplerecon}, a state-of-the-art depth estimation method from posed images.  
Similar to other MVS methods \eg~\cite{collins1996space,kendall2017end,yao2018mvsnet}, SimpleRecon constructs a cost volume using multiple viewpoints of the same scene.
To compute the cost volume, SimpleRecon starts by extracting feature maps for the current image as well as associated source frames. 
Extracted source-frame features are then warped to the current camera at each hypothesis depth plane and compared against features for the current frame. 
This comparison, together with additional \emph{metadata}, produces a $C {\times} D {\times} H {\times} W$, feature volume with depth likelihood estimates at each location, where $C$ is the channel dimension, $D$ the number of depth planes and $H {\times} W$ the dimensions of the feature maps.
A \emph{matching MLP} is applied to each aggregated feature vector in parallel to obtain the final cost volume with dimensions $D {\times} H {\times} W$.
The advantage of this reduction step is that the cost volume can then be processed using 2D convolutions, in contrast to methods that bypass this step and require expensive 3D convolutions.
A follow-up monocular prior head regularizes the volume using 2D CNNs and a depth decoder finally produces the output depth map.


\subsection{Hint MLP}

In addition to using source and target frames, our method allows the input of a \emph{rendered depth} and a \emph{confidence map} from a prior geometric representation of the scene.
Note that this geometry could be from the distant past, but it could also be from the recent past \eg as constructed incrementally from previous frames we have recently seen.
The rendered depth map is a 2D image of depth values, where each pixel represents the rendered depth from the current camera position to our prior estimate of geometry along that camera ray.
The confidence map is a 2D image where each pixel gives an estimate of how certain we are that the prior depth estimate is correct at that pixel, where values of $1.0$ indicate that we have extremely high confidence in the prior depth at this pixel and values of $0.0$ mean a very low confidence in the rendered geometry at this pixel.

Following~\cite{sayed2022simplerecon} we  create a feature volume and apply a \emph{matching MLP} to each combined feature vector individually to give a matching score, creating a cost volume with dimensions $D {\times} H {\times} W$.
Inspired by this, we use an additional \emph{Hint MLP} to combine the information from the cost volume with the rendered depth and confidence images.
Like the matching MLP, the Hint MLP is applied in parallel at each spatial location and depth plane in the feature volume.
Our Hint MLP takes as input a vector with three elements: (i) the matching score from the cost volume, (ii) the geometry hint, defined as the absolute difference between the TSDF rendered depth and the current depth plane and (iii) the confidence value at that pixel.
For pixels where there isn't a rendered depth value, for example because they haven't been seen before, we set the confidence to $0$ and the geometry hint to $-1$. This is done both at training and test time.

\begin{figure}[t]
    \centering
    \includegraphics[width=0.95\linewidth]{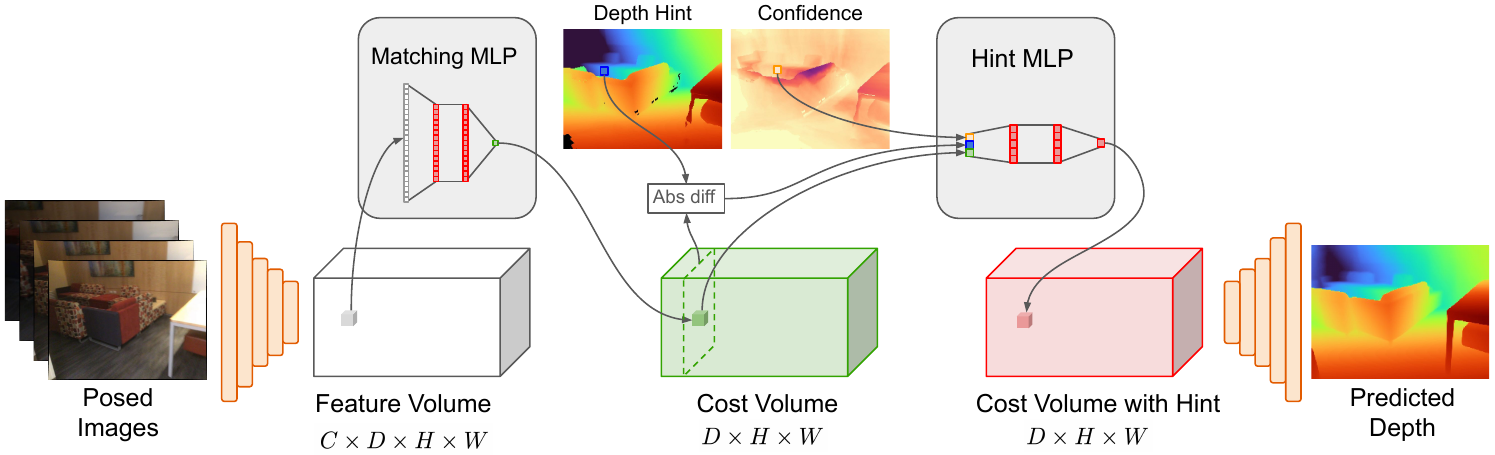}
    \caption{
        \textbf{Method detail.} 
            Our feature volume is reduced to a cost volume via a \emph{matching MLP}.  
            Our \emph{Hint MLP} then combines the multi-view-stereo cost volume with an estimate of previously predicted geometry.  
            For every location in the cost volume, the \emph{Hint MLP} takes as input (i) the visual matching score, (ii) the geometry hint, formed as the absolute difference between the rendered depth hint and the depth plane at that cost volume position, and (iii) an estimate of the confidence of the hint at that pixel. 
    }
    \label{fig:method-detail}
\end{figure}

See Fig.~\ref{fig:method-detail} for an illustration of how the \emph{Hint MLP} is incorporated into the network architecture.
We validate our choice of how to combine the different sources of information in our ablations in Section~\ref{sec:ablation}.

\subsection{Maintaining persistent 3D geometry}

In our work, we encode our persistent geometry as a truncated signed distance function (TSDF). 
A TSDF is a volumetric representation of the scene that stores at each voxel the distance (truncated and signed) from the voxel to the nearest estimated 3D surface, together with a scalar confidence value. 
Our TSDF is built up using fused depth maps produced using our method at previous keyframes.

\paragraph{Rendered depth and rendered confidence.}
For each new frame, given the corresponding intrinsics and extrinsics, we render a depth map and a confidence map from the TSDF from the point of view of the camera. 
This is achieved using marching cubes~\cite{lorensen1998marching} to obtain a mesh, followed by a mesh rendering step.
We render the mesh using PyTorch3D \cite{ravi2020pytorch3d}, which runs on the GPU.
The marching cubes step takes 9.4ms and the mesh rendering step takes 9.2ms on average (see Section~\ref{sec:timings} for more timings). 
We obtain the confidence map by sampling the voxel confidence channel in the TSDF with coordinates from the backprojected rendered depth; this takes less than 1ms.

\paragraph{Updating the persistent 3D geometry.} 
Once we have estimated a depth map for a new frame we aggregate  it into the current geometry estimate. 
Our fusion follows InfiniTAM~\cite{InfiniTAM_ECCV_2016,InfiniTAM_ISMAR_2015}.
When updating the TSDF values we also update the confidence measures.
The confidence for a pixel with depth $d$ is $c = 0.025\times \max((1-\hat{d})^2, 0.25)$, 
where $\hat{d} = (d - \text{depth}_{\min}) / (\text{depth}_{\max} - \text{depth}_{\min})$ is the depth value normalized to $[0, 1]$. This gives pixels with higher depths a lower confidence values, while the clamp to a minimum value of 0.25 ensures that even distant predictions have some confidence assigned.


\paragraph{Motivating our representations.}
We use TSDFs and meshes for our persistent geometry as they are ubiquitous, lightweight in terms of memory and runtime, and easy and quick to update and extract geometry from.
This is unlike other geometry representations \eg~NeRFs or other implicit methods, where the geometry is more `baked-in' and harder to update and extract.

\vspace{-4pt}
\subsection{Sources of prior geometry}
\vspace{-4pt}
\label{sec:sources_of_geometry}

Our prior geometry can come from different sources.
\begin{itemize}[label=$\bullet$]

\item When we see an environment for the very first time, \ie we haven't previously scanned this environment, the TSDF is built up \incrementally as the camera moves in the new environment. 
When the camera views a completely unseen part of the environment, the network doesn't have access to a geometry hint and must rely only on the cost volume.
Over time, more and more of the environment will be present in the geometry hints.

\item Alternatively, when we \textbf{\revisit} a location we have previously seen, we can load previously generated geometry to use for geometry hints.
In this scenario, we assume the current camera position is in the same coordinate system as the loaded geometry. However, we may find that some items have moved since the original geometry was created.
Our experiments, with the 3RScan dataset \cite{Wald2019RIO}, evaluate precisely this scenario. 

\item Finally, if we want the best possible reconstruction from our feedforward model we can run \textbf{\offline}.
Here, we use a \emph{two-pass} approach: 
We first run the full RGB sequence through our system \emph{without} using hints to build an initial TSDF. 
We then run the full sequence a second time, where the TSDF generated on the first pass is used as the hint for every frame.

\end{itemize}
\subsection{Training data generation}

At test time, our \emph{Hint MLP} is likely to encounter a range of scenarios: at the start of sequences, for example, there may not be a geometry hint available. 
At the other extreme, after a whole sequence has been observed, subsequent predictions may have access to a hint for almost every input pixel.
To ensure our network is robust to different test-time scenarios, we train with different randomly selected types of hints.

For 50\% of the training items we don't provide a geometry hint, instead feeding $-1$ as the rendered depth and $0$ as the confidence; in this scenario, the network must learn to rely only on the matching cost.
For the other 50\% of the training items the network has access both to the matching cost and to the rendered depth and rendered confidence.
We generate a dataset of geometry hints by fusing depths from \cite{sayed2022simplerecon} into a training-time TSDF.
    50\% of the time that we give a training-time hint it is rendered from the full and complete TSDF, assuming the entire scene has been previously fused, and 
    50\% a partial TSDF fusing only frames up to the current training frame.

\subsection{Implementation details}

We train our model (and all our ablations) with a batch size of 16 on each of two Nvidia A100 GPUs. We use a learning rate of $1e{-}4$ that we drop by a factor of 10 at $70000$ and $80000$ steps. 
We train with an input dimension of $512\times384$, and an output dimension of $256\times192$.
During training, we color augment RGB images.
Our network follows that of \cite{sayed2022simplerecon}, with the addition of our `Hint MLP' as described above.
This network uses an EfficientNetV2 S~\cite{tan2021efficientnetv2} encoder for the monocular image prior and the first two blocks of a ResNet18~\cite{he2016deep} encoder for generating matching features for the cost volume.
The decoder follows UNet++ \cite{zhou2018unet++}.
Our \emph{Hint MLP} is small with two hidden layers, each comprising 12 neurons; this adds $2$ms to our runtime.
Full architecture details are given in the supplementary material.
Our training losses directly follow \cite{sayed2022simplerecon} and their all applied at the final network output. Our \emph{Hint MLP} does not require any intermediate supervision.
We select source frames for the cost volume with the strategy and hyperparameters from~\cite{duzceker2021deepvideomvs}.
For evaluation of online methods, all source frames come from the past in a sequence. 
For evaluation of offline methods, source frames may come from anywhere in a sequence. 
For both `ours', SimpleRecon \cite{sayed2022simplerecon} and DeepVideoMVS \cite{duzceker2021deepvideomvs} we use the same TSDF fusion method. We fuse depth maps to a maximum depth of $3.5$m into a TSDF volume of $2$cm resolution. 
For the FineRecon \cite{Stier_2023_ICCV} baseline we use a $1$cm voxel resolution.

\section{Evaluation}

We evaluate with three challenging 3D datasets, all acquired with a handheld RGBD sensor.
We train and evaluate on \textbf{ScanNetV2~\cite{dai2017scannet}}, which comprises 1,201 training scans, 312 validation scans, and 100 testing scans of indoor scenes. 
We additionally evaluate our ScanNetV2 models on \textbf{7-Scenes~\cite{shotton2013scene}} without fine-tuning,  following \eg~\cite{sayed2022simplerecon} and with the test split from \cite{duzceker2021deepvideomvs}.
We also evaluate on \textbf{3RScan~\cite{Wald2019RIO}}.
This dataset captures the same environment in multiple separate scans, between which objects' positions have changed. 
This tests our ability to use scans captured in the past as `hints' for instantaneous depth estimates.

\begin{figure}[t]
  \centering
  \includegraphics[width=0.9\textwidth]{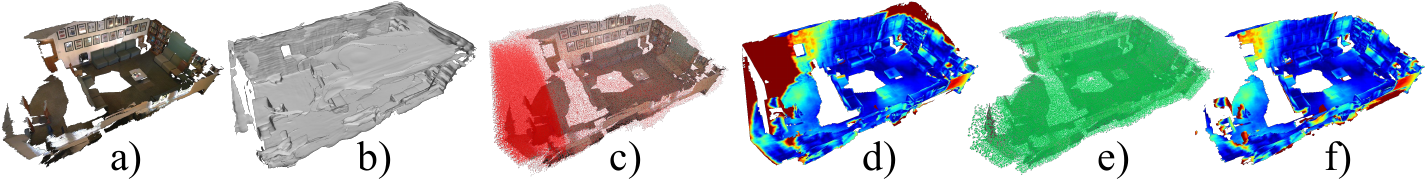}
  \caption{
    \textbf{We introduce a more accurate mesh evaluation.}
        (a) shows the ground truth mesh, which contains many holes. 
        (b) shows an example predicted mesh, here from \cite{Stier_2023_ICCV}. 
        This is punished for being too complete, as \cite{bozic2021transformerfusion}'s visibility mask (c) extends beyond the ground truth, giving high \emph{Acc} error on their prediction (d). 
        Our new masking (e) is tighter to the ground truth mesh, giving a more meaningful error (f).}
  \label{fig:new_masks}
\end{figure}

\begin{table*}[!bt]
    \begin{center}
    \setlength{\tabcolsep}{0.3em} 
    \resizebox{0.975\columnwidth}{!}{  
        \begin{tabular}{lccccc|ccccc}
            & \multicolumn{5}{c}{\textbf{ScanNetV2}} & \multicolumn{5}{c}{\textbf{7Scenes}} \\
            \cmidrule(lr){2-6}\cmidrule(lr){7-11}
            & Abs Diff$\downarrow$  & Abs Rel$\downarrow$ & Sq Rel$\downarrow$ & $\delta < 1.05$$\uparrow$ &  $\delta < 1.25$$\uparrow$ & Abs Diff$\downarrow$  & Abs Rel$\downarrow$ & Sq Rel$\downarrow$ & $\delta < 1.05$$\uparrow$ &  $\delta < 1.25$$\uparrow$ \\
            \midrule
            DPSNet~\cite{im2019dpsnet} & .1552 & .0795 & .0299 & 49.36 & 93.27 & .1966 & .1147 & .0550 & 38.81 & 87.07 \\
            MVDepthNet~\cite{wang2018mvdepthnet} & .1648 & .0848 & .0343 & 46.71 & 92.77 & .2009 & .1161 & .0623 & 38.81 & 87.70 \\
            DELTAS~\cite{sinha2020deltas} & .1497 & .0786 & .0276 & 48.64 & 93.78 & .1915 & .1140 & .0490 & 36.36 & 88.13 \\
            GPMVS~\cite{hou2019multi} & .1494 & .0757 & .0292 & 51.04 & 93.96 & .1739 & .1003 & .0462 & 42.71 & 90.32 \\
            DeepVideoMVS, fusion~\cite{duzceker2021deepvideomvs}* & \cellcolor{thirdcolor}.1186 & \cellcolor{thirdcolor}.0583 & \cellcolor{thirdcolor}.0190 & \cellcolor{thirdcolor}60.20 & \cellcolor{thirdcolor}96.76 & \cellcolor{thirdcolor}.1448 & \cellcolor{thirdcolor}.0828 & \cellcolor{thirdcolor}.0335 & \cellcolor{thirdcolor}47.96 & \cellcolor{thirdcolor}93.79 \\
            SimpleRecon \cite{sayed2022simplerecon} & \cellcolor{secondcolor}.0873 & \cellcolor{secondcolor}.0430 & \cellcolor{secondcolor}.0128 & \cellcolor{secondcolor}74.12 & \cellcolor{secondcolor}98.05 & \cellcolor{secondcolor}.1045 & \cellcolor{secondcolor}.0575 & \cellcolor{secondcolor}.0156 & \cellcolor{secondcolor}60.12 & \cellcolor{firstcolor}97.33 \\
            \textbf{Ours} (\incremental) & \cellcolor{firstcolor}.0767 & \cellcolor{firstcolor}.0369 & \cellcolor{firstcolor}.0112 & \cellcolor{firstcolor}79.94 & \cellcolor{firstcolor}98.35 & \cellcolor{firstcolor}.0985 & \cellcolor{firstcolor}.0534 & \cellcolor{firstcolor}.0156 & \cellcolor{firstcolor}64.76 & \cellcolor{secondcolor}97.01 \\
            \midrule
            \textbf{Ours} (no hint) & .0870 & .0428 & .0128 & 74.35 & 98.02 & .1082 & .0593 & .0171 & 59.44 & 96.97 \\
            \textbf{Ours} (\offline) & .0627 & .0306 & .0092 & 86.46 & 98.62 & .0858 & .0466 & .0133 & 71.74 & 97.61 \\
            \bottomrule
        \end{tabular}
    }
    \end{center}
    \caption{\textbf{Depth evaluation on ScanNetV2 and 7Scenes.} 
    Unless stated otherwise, predictions are computed incrementally, without access to future frames or frames from previous scans of the scene.
    We highlight the \colorbox{firstcolor}{best}, \colorbox{secondcolor}{second-best} and \colorbox{thirdcolor}{third-best} methods per metric. 
    Previous scores are from~\cite{duzceker2021deepvideomvs,sayed2022simplerecon}. 
    *\cite{duzceker2021deepvideomvs} was boosted by computing depths using three inference frames instead of two; they also use a custom 90/10 split.
    }
    
    
    \label{table:depth_results}
\end{table*}

\subsection{Mesh evaluation metrics}
\label{sec:new_masks}


We follow existing works~\cite{bozic2021transformerfusion,murez2020atlas} and report reconstruction metrics based on point to point distances on sampled point clouds.
ScanNetV2 ground truth meshes are not complete, as the scan sequences don't have full coverage.
This means methods which overpredict geometry not in the ground truth get unfairly punished. 
TransformerFusion~\cite{bozic2021transformerfusion} use a visibility mask to trim predictions when computing prediction to ground-truth distances.
However, these masks are over-sized, and include large areas of geometry which aren't present in the ground truth mesh.
We propose new visibility volumes using rendered depth maps of the ground-truth meshes, which much tighter fit the ground truth meshes.
Figure~\ref{fig:new_masks} shows the difference between masks from \cite{bozic2021transformerfusion} and our new proposed masks.
More are shown in the supplementary.
We show results on both sets of masks in Tables~\ref{table:tf_mesh_evaluation} and \ref{table:our_mesh_evaluation}, and we will release our updated masks for reproducibility.


\subsection{Depth and reconstruction performance}



\paragraph{Depth estimation.}
Table~\ref{table:depth_results} shows our depth estimation performance on ScanNetV2 and 7Scenes in both \incremental and \offline modes (see Section~\ref{sec:sources_of_geometry} for details of these modes).
Our \incremental method outperforms all baseline methods across most scores.
Our \offline approach outperforms all competitors including our incremental method.

\paragraph{Reconstruction.}
For meshing results we separate all methods into `online' and `offline': `Online' methods can produce meshes instantaneously, while `offline' methods are designed with the assumption that all frames are available at once.
Tables~\ref{table:tf_mesh_evaluation} and \ref{table:our_mesh_evaluation} show reconstruction performance both `online' and `offline'.
Our \incremental model sets a new state-of-the-art for online reconstruction performance, validating our approach to depth and reconstruction.
See Figures~\ref{fig:qualitative-comparison} and \ref{fig:qualitative-comparison-online} for qualitative comparisons with previous state-of-the-art.
Ours (\offline) obtains even better results; we present these in the second section of the table, where we compare against other offline reconstruction approaches \eg~\cite{Stier_2023_ICCV,stier2021vortx,bozic2021transformerfusion}.
Our approach is first or second best in most metrics.
Figure~\ref{fig:qualitative-comparison-two-pass} shows the benefit running ours offline can bring. 


\begin{table*}[tb]
\begin{center}
    \setlength{\tabcolsep}{0.3em} 
    \resizebox{0.95\columnwidth}{!}{  
    \footnotesize
        \begin{tabular}{clcccccccc}
            & & Volumetric & Acc$\downarrow$ & Comp$\downarrow$ & Chamfer$\downarrow$ & Prec$\uparrow$ & Recall $\uparrow$ & F-Score $\uparrow$\\
            \midrule
 \parbox[t]{2mm}{\multirow{4}{*}{\rotatebox[origin=c]{90}{Online}}} 
&DeepVideoMVS~\cite{duzceker2021deepvideomvs} & No & \cellcolor{thirdcolor}6.49 & \cellcolor{thirdcolor}6.97 & \cellcolor{thirdcolor}6.73 & \cellcolor{thirdcolor}.568 & \cellcolor{thirdcolor}.595 & \cellcolor{thirdcolor}.579 \\
&NeuralRecon~\cite{sun2021neuralrecon} & Yes & 7.31 & 10.81 & 9.06 & .453 & .592 & .511 \\
&SimpleRecon~\cite{sayed2022simplerecon} (online)  & No & \cellcolor{secondcolor}5.56 & \cellcolor{firstcolor}5.02 & \cellcolor{secondcolor}5.29 & \cellcolor{secondcolor}.631 & \cellcolor{firstcolor}.712 & \cellcolor{secondcolor}.668 \\
&\textbf{Ours} (\incremental) & No & \cellcolor{firstcolor}4.92 & \cellcolor{secondcolor}5.49 & \cellcolor{firstcolor}5.20 & \cellcolor{firstcolor}.685 & \cellcolor{secondcolor}.701 & \cellcolor{firstcolor}.692 \\
    \midrule
    &ATLAS~\cite{murez2020atlas} & Yes & 5.59 & 7.52 & 6.55 & .671 & .610 & .637 \\
    \parbox[t]{2mm}{\multirow{4}{*}{\rotatebox[origin=c]{90}{Offline}}} &TransformerFusion~\cite{bozic2021transformerfusion} & Yes & \cellcolor{thirdcolor}4.68 & 8.27 & 6.48 & \cellcolor{thirdcolor}.698 & .600 & .644 \\
    &VoRTX~\cite{stier2021vortx} & Yes & \cellcolor{secondcolor}4.38 & 7.23 & 5.80 & \cellcolor{secondcolor}.726 & .651 & .685 \\
    &SimpleRecon~\cite{sayed2022simplerecon} (offline) $\dagger$ & No & 5.25 & \cellcolor{secondcolor}4.86 & \cellcolor{thirdcolor}5.05 & .654 & \cellcolor{thirdcolor}.725 & \cellcolor{thirdcolor}.687 \\
    &FineRecon~\cite{Stier_2023_ICCV} & Yes & 4.92 & \cellcolor{thirdcolor}5.06 & \cellcolor{secondcolor}4.99 & .687 & \cellcolor{firstcolor}.737 & \cellcolor{secondcolor}.710 \\
    &\textbf{Ours} (\offline) & No & \cellcolor{firstcolor}4.15 & \cellcolor{firstcolor}4.85 & \cellcolor{firstcolor}4.50 & \cellcolor{firstcolor}.743 & \cellcolor{secondcolor}.734 & \cellcolor{firstcolor}.738 \\
            \bottomrule
        \end{tabular}
    }
    \end{center}
    \caption{\textbf{Mesh Evaluation on ScanNetV2 with new visibility masks}.
    We evaluate methods using the evaluation from \cite{bozic2021transformerfusion} using our visibility masks that more accurately represent the ground-truth mesh (Sec.~\ref{sec:new_masks}). The `Volumetric' category indicates whether a technique involves volumetric 3D reconstruction.
    Following \cite{sayed2022simplerecon}, for other Multi-View Stereo (MVS) methods that generate solely depth maps, we applied conventional TSDF fusion for reconstruction. 
    \textbf{Chamfer} distance is the mean of accuracy and completion and \textbf{F-Score} is the harmonic mean of precision and recall.
    $\dagger$~Like Ours (offline),
    SimpleRecon \cite{sayed2022simplerecon} (offline) uses source frames from both past and future.}
    \label{table:our_mesh_evaluation}
\end{table*}

\begin{table*}[t]
    \begin{center}
    \setlength{\tabcolsep}{0.3em} 
    \resizebox{0.99\columnwidth}{!}{  
        \begin{tabular}{llcccccc}
            & & Abs Diff$\downarrow$ & Abs Rel$\downarrow$ & Sq Rel$\downarrow$ & RMSE$\downarrow$ & $\delta < 1.05$$\uparrow$ & $\delta < 1.25$$\uparrow$\\
            \midrule
\row{row:ours_no_hint_mlp} & \textbf{Ours} without Hint MLP (as in SimpleRecon~\cite{sayed2022simplerecon}) & .0873 & .0430 & .0128 & .1483 & 74.12 & 98.05 \\
\row{row:ours_no_confidence} & \textbf{Ours} w/ online hint, without confidence & .0863 & .0392 & .0129 & .1529 & 77.81 & 98.02 \\
\row{row:ours_hint_in_cv_enc} & \textbf{Ours} w/ hint \& confidence to cost volume encoder & .0890 & .0438 & .0130 & .1506 & 73.39 & 97.95 \\
\row{row:ours_warped_depth} & \textbf{Ours} w/ warped depth as hint & .0889 & .0436 & .0130 & .1492 & 73.26 & 98.06 \\
\row{row:ours_variable_depth_planes} & \textbf{Ours} w/ hint-based variable depth planes \cite{Xin2023ISMAR} & .0821 & .0405 & .0121 & .1432 & 76.67 & 98.19 \\
\row{row:tocd} & SimpleRecon~\cite{sayed2022simplerecon} w/ TOCD \cite{khan2023temporally} & .0880 & .0437 & .0134 & .1505 & 74.19 & 97.83 \\
\row{row:ours_cost_volume_modulation} & \textbf{Ours} w/ hint cost volume modulation \cite{poggi2022multi} & .0773 & .0372 & .0112 & .1381 & 79.52 & 98.34 \\
\row{row:ours_single_mlp} & \textbf{Ours} w/ single MLP for matching and hints & .0773 & .0371 & .0112 & .1381 & 79.56 & 98.35 \\
\row{row:ours_no_hint} & \textbf{Ours} (no hint) & .0870 & .0428 & .0128 & .1477 & 74.35 & 98.02 \\
\row{row:ours_fast} & \textbf{Ours} (\incremental, fast) & .0826 & .0400 & .0125 & .1473 & 77.14 & 98.05 \\
\row{row:ours} & \textbf{Ours} (\incremental) & .0767 & .0369 & .0112 & .1377 & 79.94 & 98.35 \\
\hline
\row{row:ours_guided_cv_offline} & \textbf{Ours} offline w/ hint cost volume modulation  \cite{poggi2022multi} & .0651 & .0320 & .0094 & .1242  & 84.92 & 98.61 \\
\row{row:ours_offline} & \textbf{Ours} (\offline) & .0627 & 0306 & .0092 & .1225  & 86.46 & 98.62 \\
            \bottomrule
        \end{tabular}
    }
    \end{center}
    \caption{
        \textbf{Incremental ablation evaluation.} 
        Scores are depth metrics on ScanNetV2.
        See the text for descriptions of these variants, and the supplementary for full metrics. 
    }
    \label{tab:ablations}
\end{table*}

\begin{table*}[tb]
\begin{center}
    \setlength{\tabcolsep}{0.3em} 
    \resizebox{0.98\columnwidth}{!}{  
    \footnotesize
        \begin{tabular}{clcccccccc}
            & & Volumetric & Acc$\downarrow$ & Comp$\downarrow$ & \textbf{Chamfer}$\downarrow$ & Prec$\uparrow$ & Recall$\uparrow$ & \textbf{F-Score}$\uparrow$\\
            \midrule
\parbox[t]{2mm}{\multirow{6}{*}{\rotatebox[origin=c]{90}{Online}}} &DPSNet~\cite{im2019dpsnet} & No & 11.94 & 7.58 & 9.77 & .474 & .519 & .492 \\
&DELTAS~\cite{sinha2020deltas} & No & 11.95 & 7.46 & 9.71 & .478 & .533 & .501 \\
&DeepVideoMVS~\cite{duzceker2021deepvideomvs} & No & 5.84 & \cellcolor{thirdcolor}6.97 & \cellcolor{thirdcolor}6.41 & \cellcolor{thirdcolor}.639 & .595 & .615 \\
&NeuralRecon~\cite{sun2021neuralrecon} & Yes & \cellcolor{secondcolor}5.09 & 9.13 & 7.11 & .630 & \cellcolor{thirdcolor}.612 & \cellcolor{thirdcolor}.619 \\
&SimpleRecon~\cite{sayed2022simplerecon} (online)  & No & \cellcolor{thirdcolor}5.72 & \cellcolor{firstcolor}5.02 & \cellcolor{secondcolor}5.37 & \cellcolor{secondcolor}.682 & \cellcolor{firstcolor}.712 & \cellcolor{secondcolor}.696 \\
&\textbf{Ours} (\incremental) & No & \cellcolor{firstcolor}4.70 & \cellcolor{secondcolor}5.49 & \cellcolor{firstcolor}5.09 & \cellcolor{firstcolor}.730 & \cellcolor{secondcolor}.701 & \cellcolor{firstcolor}.714 \\
    \midrule
    \parbox[t]{2mm}{\multirow{8}{*}{\rotatebox[origin=c]{90}{Offline}}} &COLMAP~\cite{schonberger2016pixelwise,schoenberger2016sfm} & No & 10.22 & 11.88 & 11.05 & .509 & .474 & .489 \\
    &ATLAS~\cite{murez2020atlas} & Yes & 7.11 & 7.52 & 7.31 & .679 & .610 & .640 \\
    &3DVNet~\cite{rich20213dvnet} & Yes & 6.73 & 7.72 & 7.22 & .655 & .596 & .621 \\
    &TransformerFusion~\cite{bozic2021transformerfusion} & Yes & 5.52 & 8.27 & 6.89 & .729 & .600 & .655 \\
    &VoRTX~\cite{stier2021vortx} & Yes & \cellcolor{firstcolor}4.31 & 7.23 & 5.77 & \cellcolor{secondcolor}.767 & .651 & .703 \\
    &SimpleRecon~\cite{sayed2022simplerecon} (offline)$\dagger$  & No & 5.37 & \cellcolor{secondcolor}4.86 & \cellcolor{secondcolor}5.12 & .702 & \cellcolor{thirdcolor}.725 & \cellcolor{thirdcolor}.712 \\
    &FineRecon~\cite{Stier_2023_ICCV} & Yes & \cellcolor{thirdcolor}5.25 & \cellcolor{thirdcolor}5.06 & \cellcolor{thirdcolor}5.16 & \cellcolor{firstcolor}.779 & \cellcolor{firstcolor}.737 & \cellcolor{firstcolor}.756 \\
    &\textbf{Ours} (\offline) & No & \cellcolor{secondcolor}4.96 & \cellcolor{firstcolor}4.85 & \cellcolor{firstcolor}4.90 & \cellcolor{thirdcolor}.752 & \cellcolor{secondcolor}.734 & \cellcolor{secondcolor}.742 \\
            \bottomrule
        \end{tabular}
    }
    \end{center}
    \caption{\textbf{Mesh Evaluation on ScanNetV2}.
    Here we use the evaluation and visibility masks from \cite{bozic2021transformerfusion}. Note that
VoRTX wins on accuracy, but its very sparse predictions give a poor completion score.
    }
    \label{table:tf_mesh_evaluation}
\end{table*}

\begin{figure*}[t]
    \centering
    \newcommand{\qualimwidth}{0.14\textwidth}
    \renewcommand{\tabcolsep}{2pt}
    \scriptsize

    \newcommand{\addArrowToImage}[5]{ 
        \begin{tikzpicture}
            \node[anchor=south west,inner sep=0] (image) at (0,0) {\includegraphics[width=\qualimwidth]{#1}};
            \begin{scope}[x={(image.south east)},y={(image.north west)}]
                \draw[-latex, ultra thick, white] (#2,#3) -- (#4,#5);
            \end{scope}
        \end{tikzpicture}
    }

    \newcommand{\addBunkBedArrows}[5]{ 
        \begin{tikzpicture}
            \node[anchor=south west,inner sep=0] (image) at (0,0) {\includegraphics[width=\qualimwidth]{#1}};
            \begin{scope}[x={(image.south east)},y={(image.north west)}]
                \draw[-latex, ultra thick, white] (#2,#3) -- (#4,#5);
                \draw[-latex, ultra thick, white] (0.4,0.2) -- (0.2,0.15);
            \end{scope}
        \end{tikzpicture}
    }

    \begin{tabular}{ccccccc}
        \centering

         \raisebox{1.1\height}{\parbox[t]{2mm}{\rotatebox[origin=c]{90}{\textbf{Image}}}}  & \includegraphics[width=\qualimwidth]{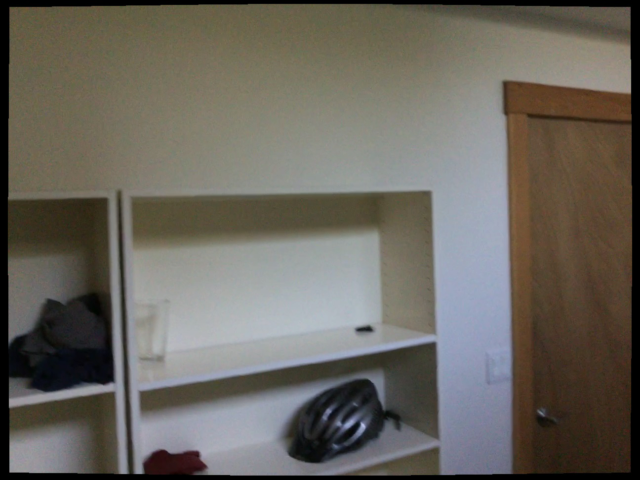} & \includegraphics[width=\qualimwidth]{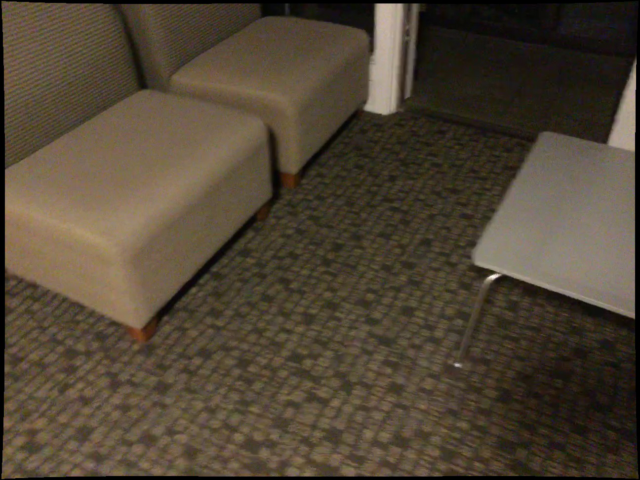} & \includegraphics[width=\qualimwidth]{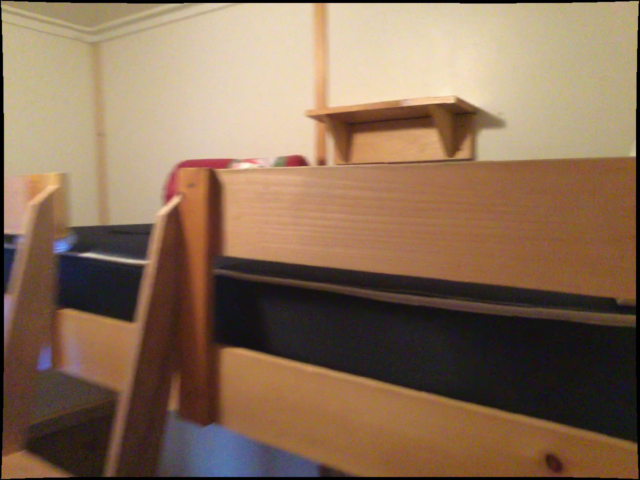} & \includegraphics[width=\qualimwidth]{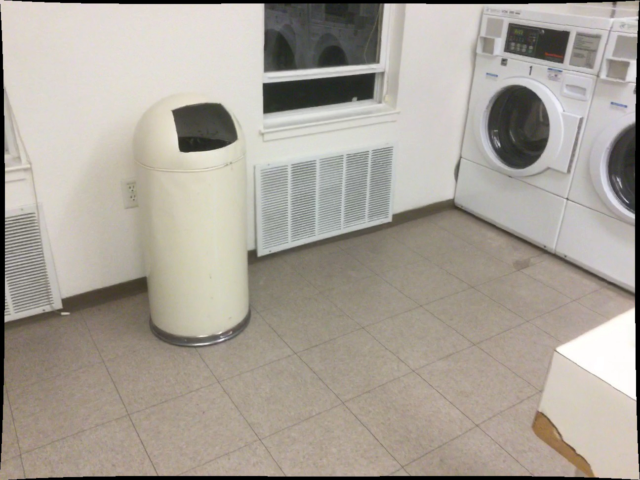} & \includegraphics[width=\qualimwidth]{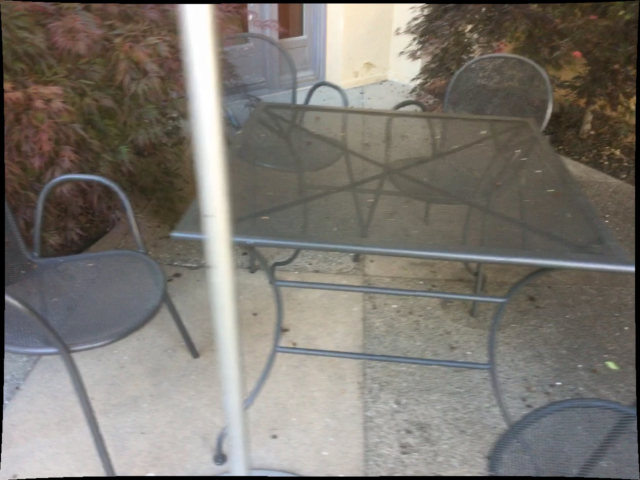} & \includegraphics[width=\qualimwidth]{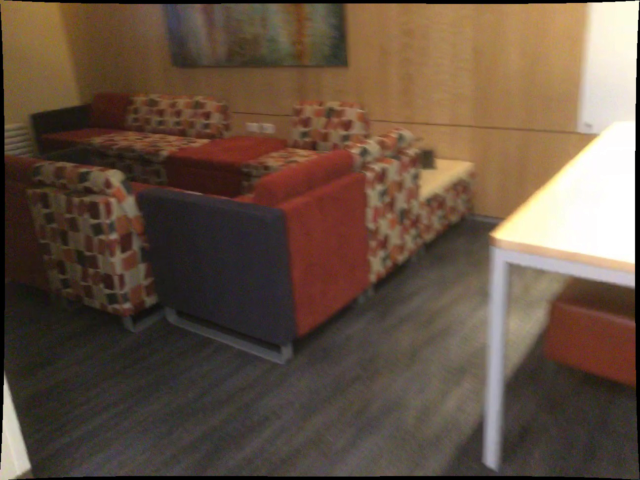} \\
        
        \raisebox{1.5\height}{\parbox[t]{2mm}{\rotatebox[origin=c]{90}{\textbf{GT}}}} & \includegraphics[width=\qualimwidth]{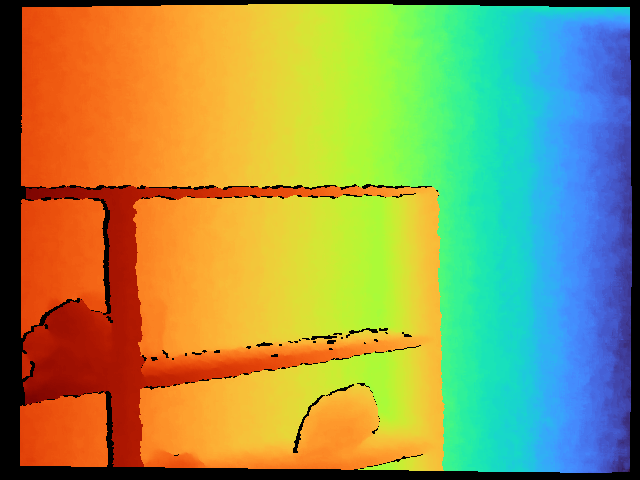} & \includegraphics[width=\qualimwidth]{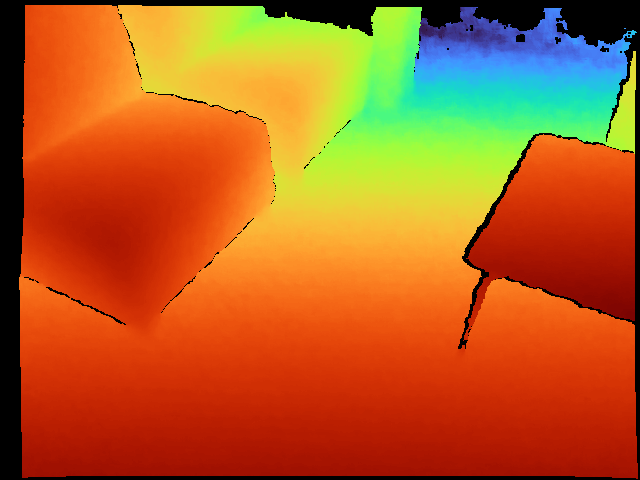} & \includegraphics[width=\qualimwidth]{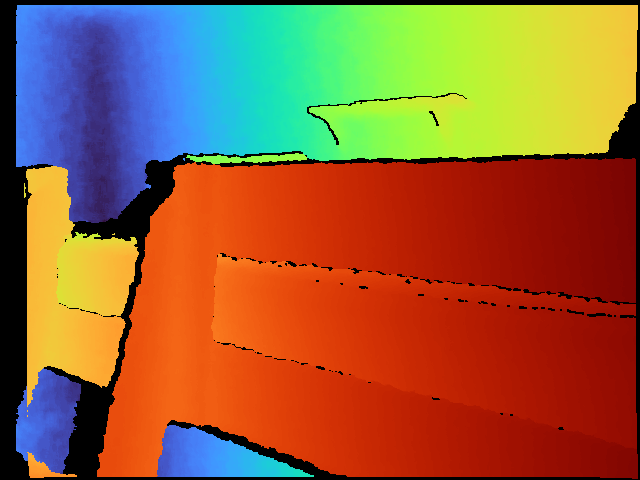} & \includegraphics[width=\qualimwidth]{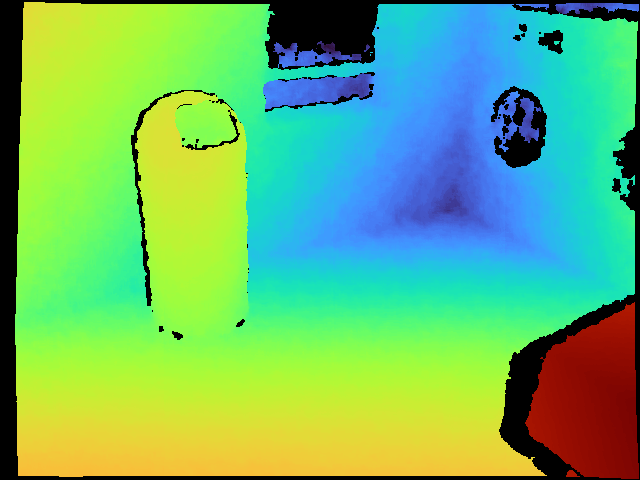} & \includegraphics[width=\qualimwidth]{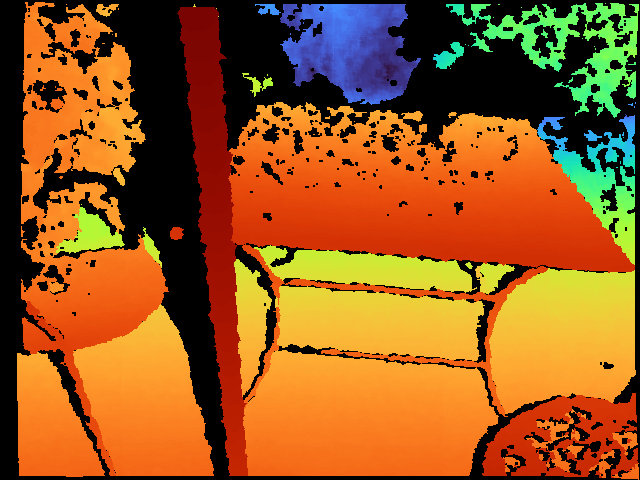} & \includegraphics[width=\qualimwidth]{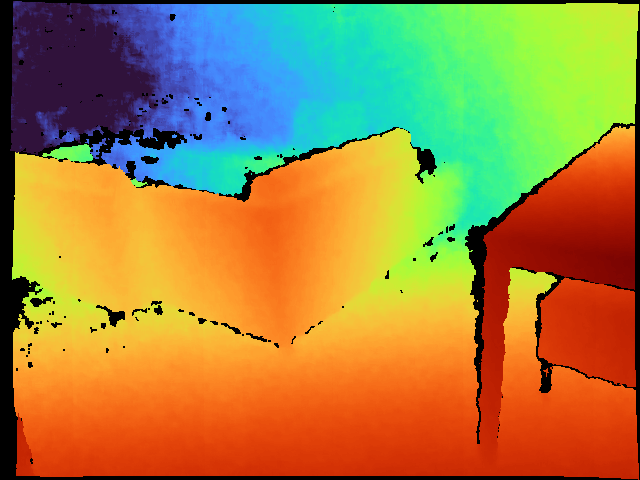} \\

        \raisebox{1.35\height}{\parbox[t]{2mm}{\rotatebox[origin=c]{90}{\textbf{Ours}}}} & \addArrowToImage{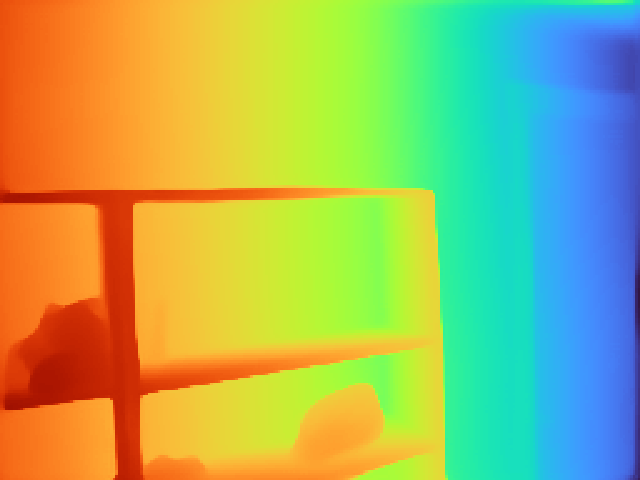}{0.7}{0.85}{0.58}{0.63} & \addArrowToImage{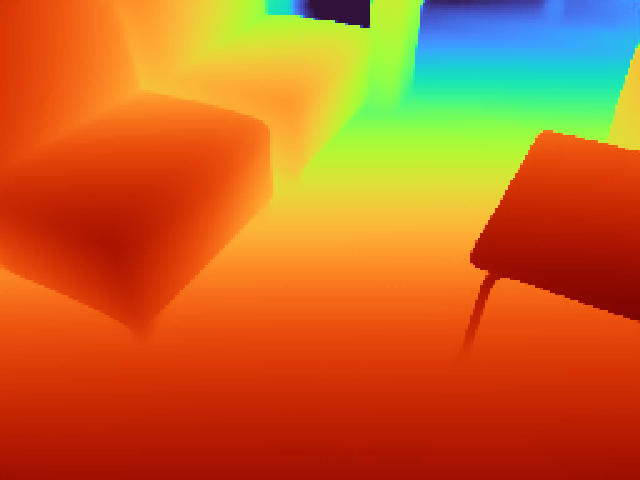}{0.5}{0.5}{0.7}{0.4} & \addBunkBedArrows{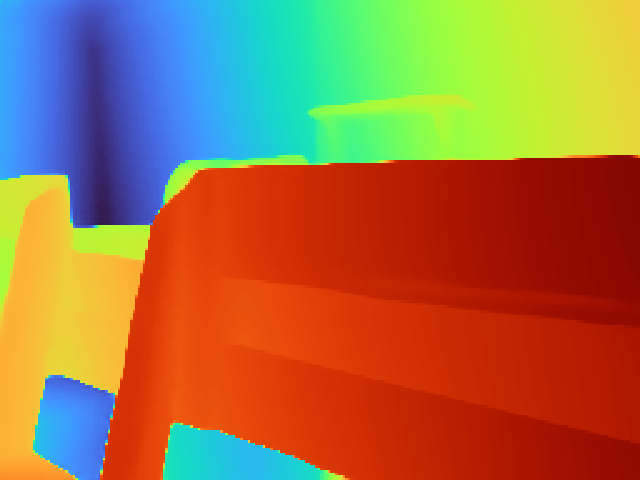}{0.4}{0.85}{0.2}{0.83} & \addArrowToImage{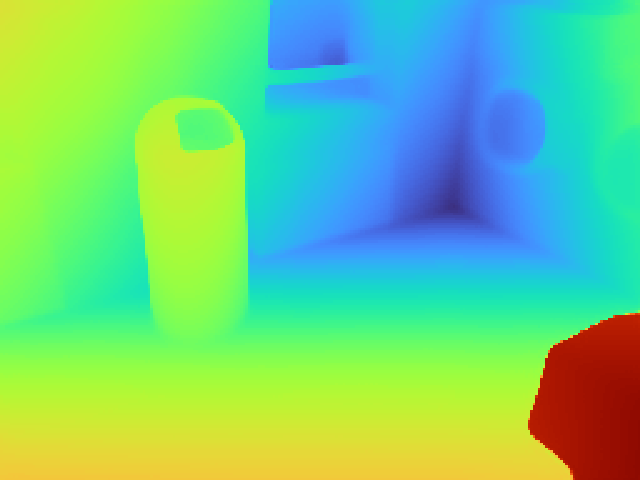}{0.6}{0.23}{0.8}{0.20} & \addArrowToImage{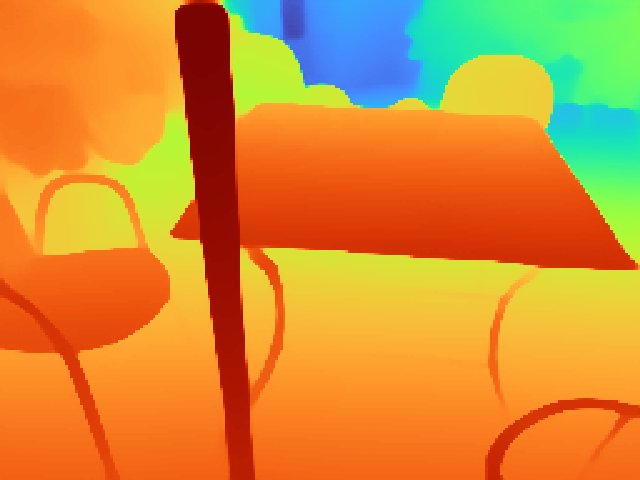}{0.70}{0.75}{0.55}{0.65} & \addArrowToImage{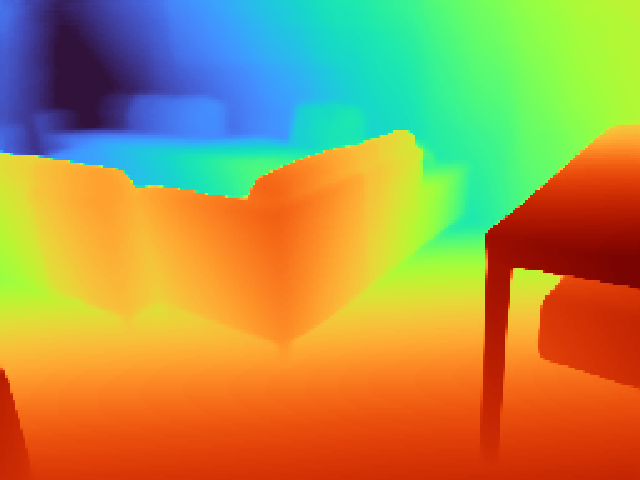}{0.40}{0.85}{0.20}{0.85} \\

        \raisebox{1.1\height}{\parbox[t]{2mm}{\rotatebox[origin=c]{90}{\textbf{SR~\cite{sayed2022simplerecon}}}}} & \addArrowToImage{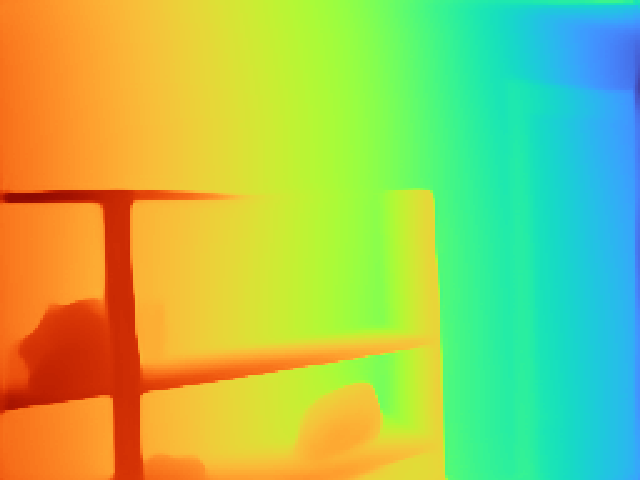}{0.7}{0.85}{0.58}{0.63} & \addArrowToImage{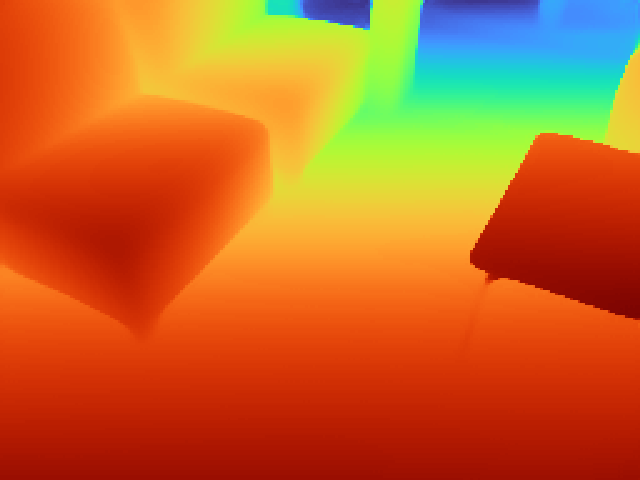}{0.5}{0.5}{0.7}{0.4} & \addBunkBedArrows{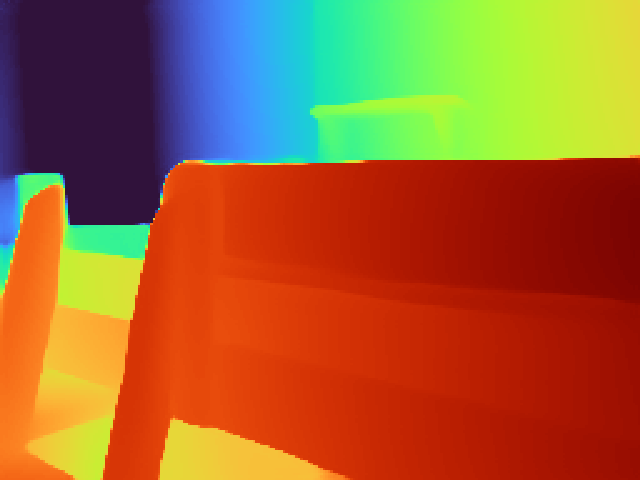}{0.4}{0.85}{0.2}{0.83} & \addArrowToImage{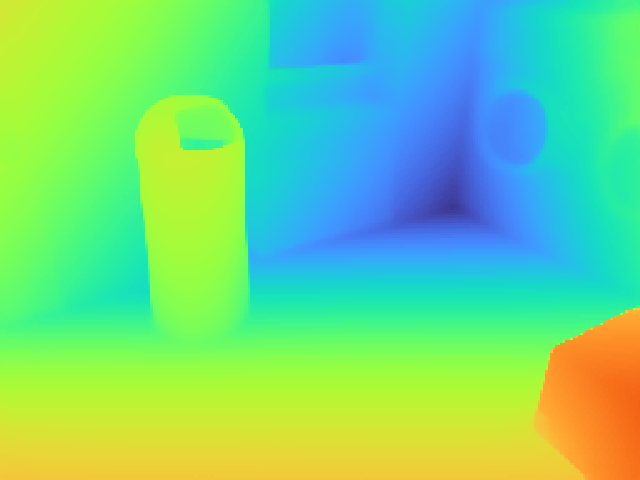}{0.6}{0.23}{0.8}{0.20} & \addArrowToImage{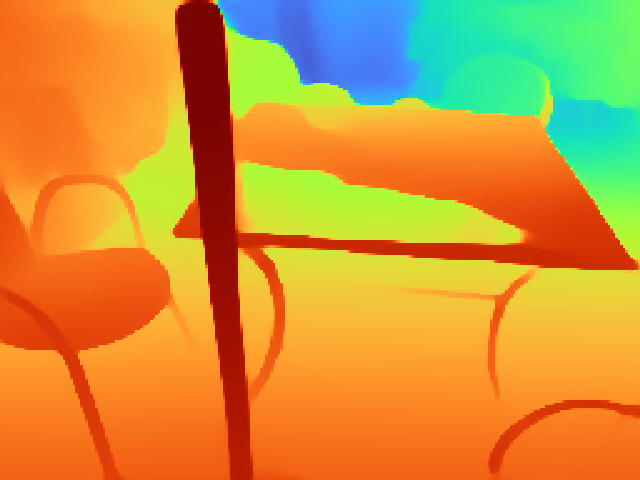}{0.70}{0.75}{0.55}{0.65} & \addArrowToImage{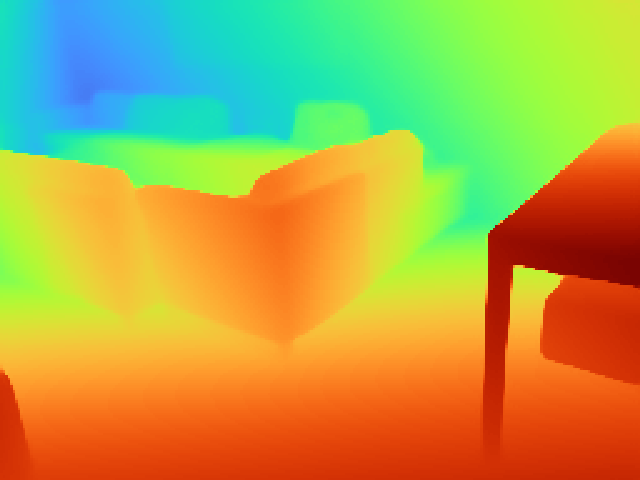}{0.40}{0.85}{0.20}{0.85} \\
        
        \raisebox{0.75\height}{\parbox[t]{2mm}{\rotatebox[origin=t]{90}{\textbf{~\cite{duzceker2021deepvideomvs}}}}} & \includegraphics[width=\qualimwidth]{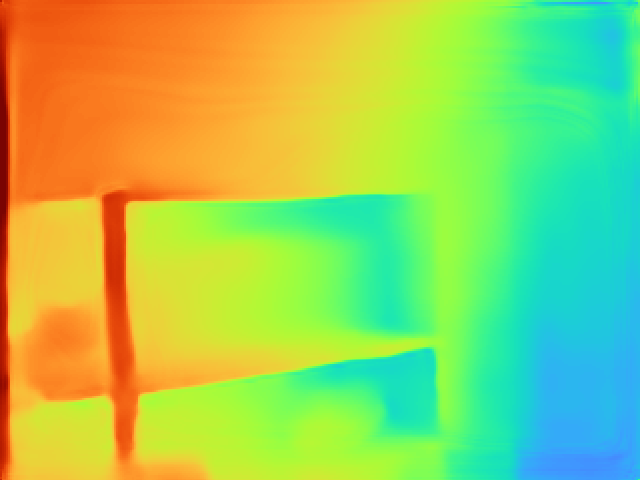} & \includegraphics[width=\qualimwidth]{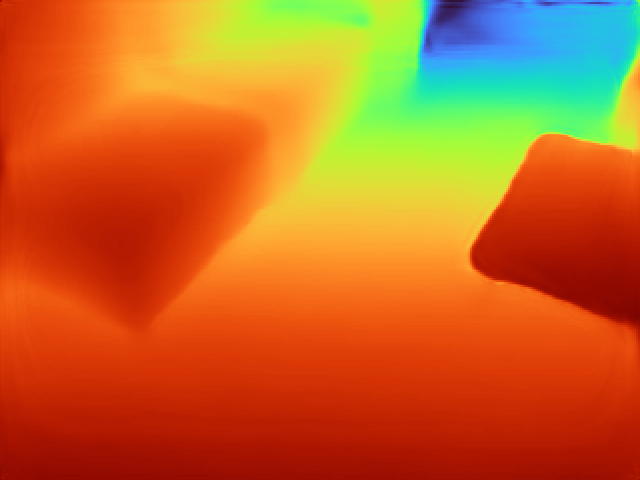} & \includegraphics[width=\qualimwidth]{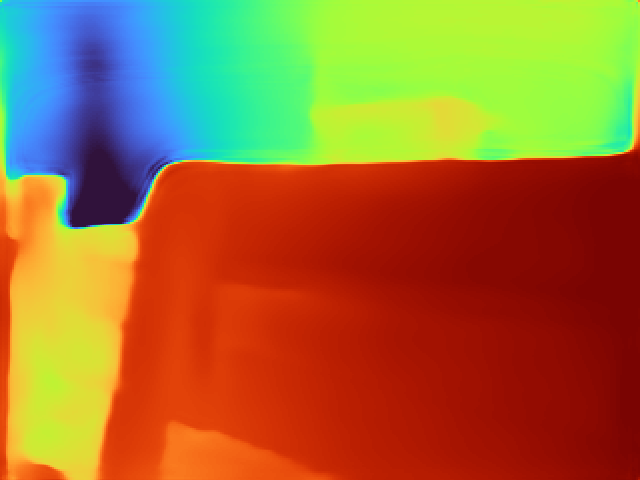} & \includegraphics[width=\qualimwidth]{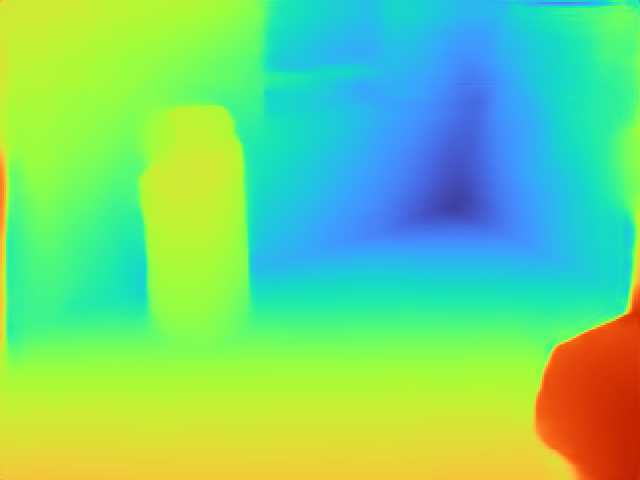} & \includegraphics[width=\qualimwidth]{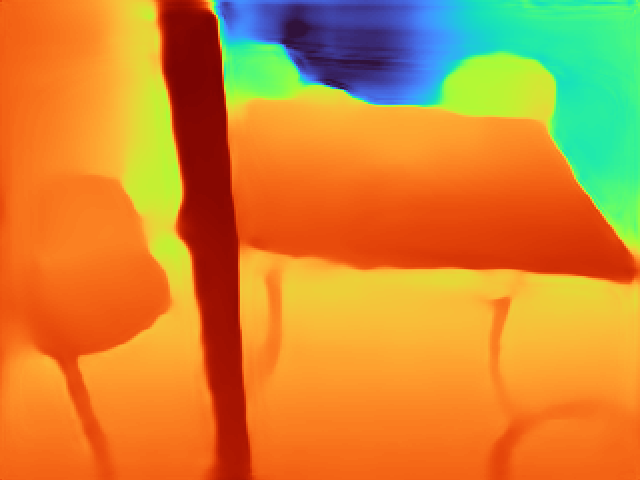} & \includegraphics[width=\qualimwidth]{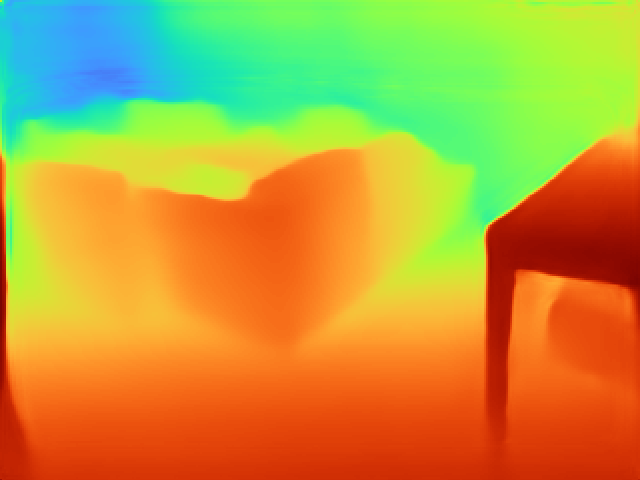} \\

    \end{tabular}
    \vspace{-10pt}
    \caption{
        \textbf{Qualitative depth results on ScanNetV2.}
        All methods are run \incrementally, where we run at interactive speeds only with access to previous frames.
        We compare with \cite{sayed2022simplerecon} and \cite{duzceker2021deepvideomvs}, the two closest-performing baselines.
        Our depth maps are more accurate with better small details (\eg top) and overall geometry (\eg bottom).
        \label{fig:qualitative-comparison}
    }
\end{figure*}

\paragraph{Long-term hints on 3RScan.}
Our system can use long-term hints \eg where we \textbf{\revisit} an environment we have first observed at some time in the past.
We use the 3RScan dataset~\cite{Wald2019RIO} for this scenario, as this includes multiple scans of the same location at different times.
We use the TSDF generated from a previous visit as the hint for the current, online depth estimates.
See Table~\ref{table:threerscan} for our results, where we see the benefit of using long-term hints vs baselines which don't have access to hints, or an incremental version of our method.
The table also includes figures showing how our method can gracefully cope with the situation where the scene has changed since the hint TSDF was generated.


\paragraph{Timings.}
\label{sec:timings}
Our online, \incremental system takes just 76.6ms to compute depth for a single frame, as measured on an Nvidia A100.
The majority of this time (52.8ms) is running the forward pass of the depth network, while the remainder is generating the hint and updating the TSDF. A version of our model with smaller networks than~\cite{sayed2022simplerecon} runs at 50.4ms per frame, see \ref{row:ours_fast} in Table~\ref{tab:ablations}. This compares to \eg 58ms per frame update of \cite{sayed2022simplerecon} or 90ms of \cite{sun2021neuralrecon}.
{\textcolor{offline}{Offline}}, we are faster than close competitor FineRecon \cite{Stier_2023_ICCV}, taking on average 13.8s per scene vs.~\cite{Stier_2023_ICCV}'s 48.1s.
See Supplementary Material for a full breakdown of timings.

\begin{table*}[tb]
\begin{center}
    \setlength{\tabcolsep}{0.3em} 
    \begin{minipage}[b]{0.75\linewidth}
        \scriptsize
        \resizebox{1.05\linewidth}{!}{  
        \begin{tabular}{lccccc}
            & Abs Diff$\downarrow$ & Sq Rel$\downarrow$ & RMSE$\downarrow$ & $\delta < 1.05$$\uparrow$ & $\delta < 1.25$$\uparrow$\\
            \midrule
            Rendered depth from TSDF & .2506 & .1293 & .3338 & 23.53 & 67.97 \\
            Densified rendered depth & .1763 & .0629 & .2264 & 30.61 & 81.69 \\
            SimpleRecon~\cite{sayed2022simplerecon} & .1350 & .0437 & .1879 & 46.88 & 89.32\\
            \midrule
            \textbf{Ours} (no hint) & .1346 & .0449 & .1879 & 47.28 & 89.39\\
            \textbf{Ours} (\incremental) & .1255 & .0395 & .1787 & 48.76 & 90.47 \\
            \textbf{Ours} (\revisit) & .1182 & .0368 & .1710 & 50.24 & 91.78 \\ 
            \textbf{Ours} (\revisit, pose noise) & .1199 & .0372 & .1725 & 49.46 & 91.56 \\ 
            \bottomrule
        \end{tabular}
        }
    \end{minipage}
    \begin{minipage}[b]{0.22\linewidth}
        \centering
        \newcommand{\qualimwidth}{0.3\textwidth}
        \renewcommand{\tabcolsep}{2pt}
        \small
        \begin{tabular}{cc}
            \centering
            \textbf{\scriptsize Input} & \textbf{\scriptsize GT} \\
            \includegraphics[width=\qualimwidth]{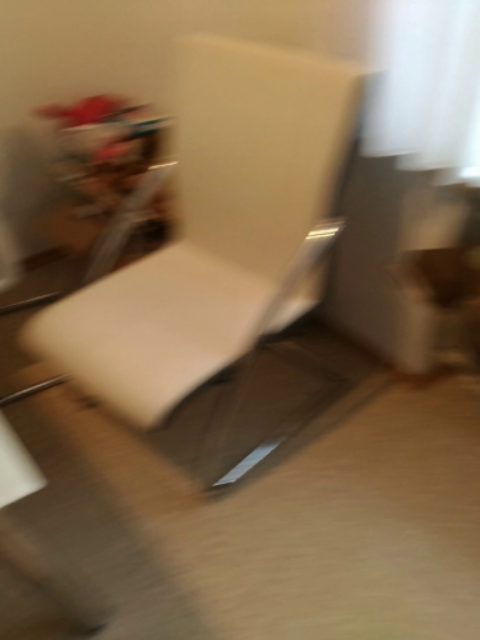} & \includegraphics[width=\qualimwidth]{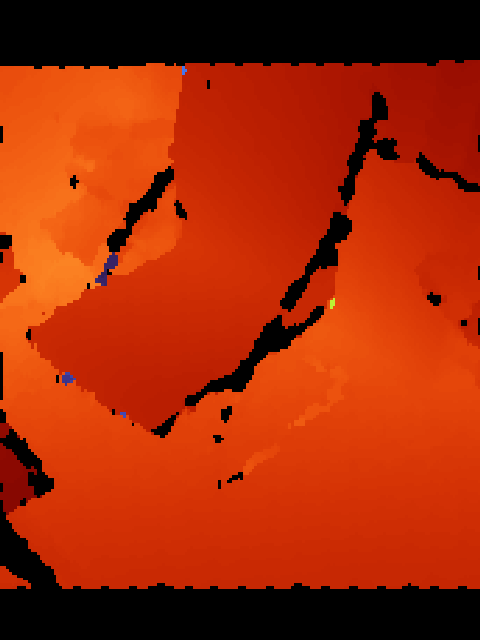} \\
            \textbf{\scriptsize Hint} & \textbf{\scriptsize Ours} \\
            \includegraphics[width=\qualimwidth]{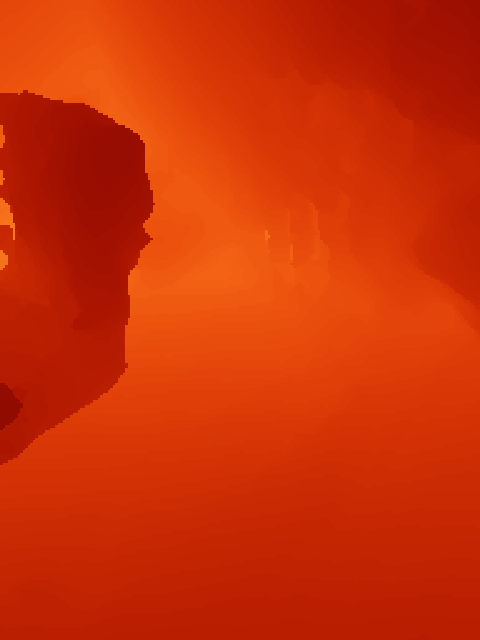} & \includegraphics[width=\qualimwidth]{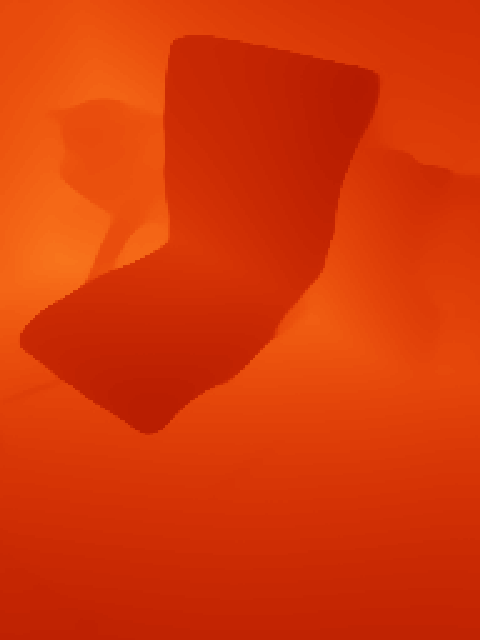} \\
        \end{tabular}
    \end{minipage}
    \end{center}
    \caption{
        \textbf{Long-term hints on 3RScan.} 
        Ours (\revisit) shows depth scores when we use the geometry from a \emph{previous visit} as `hints' for our current depth estimates.
        This mode beats previous baselines for this dataset, validating our proposal to retain long-term hints.
        \emph{Rendered depth} uses the prior TSDF's render as the current depth estimate.
        \emph{Densified rendered depth} improves upon the above baseline with a network to fill in missing geometry.
        Full metrics and descriptions are in the supplementary.
        On the right we show we are robust to stale geometry: the chair has moved position since the hint TSDF was generated, but our system gracefully recovers.
        \label{table:threerscan}
    }
\end{table*}

\begin{figure}[t]
    \centering
    \newcommand{\qualimwidth}{0.12\textwidth}
    \renewcommand{\tabcolsep}{3pt}
    \small
    \begin{tabular}{cccccc}
        \centering
        \textbf{\scriptsize Input} & \textbf{\scriptsize GT} & \textbf{\scriptsize Hint} & \textbf{\scriptsize Confidence} & \textbf{\scriptsize Ours} &  \textbf{\scriptsize SR~\cite{sayed2022simplerecon}} \\
        \includegraphics[width=\qualimwidth]{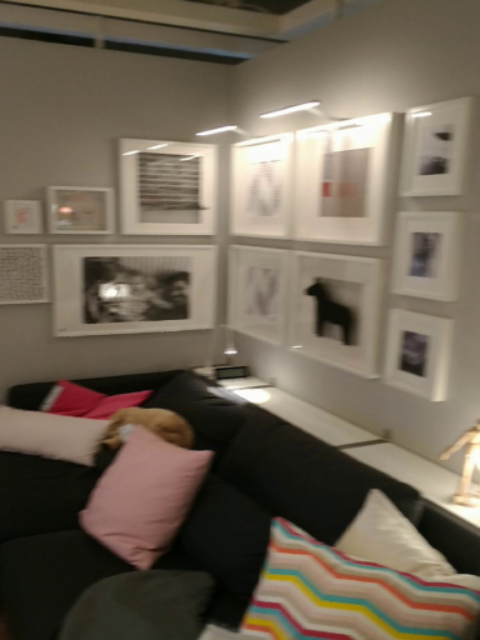} & 
        \includegraphics[width=\qualimwidth]{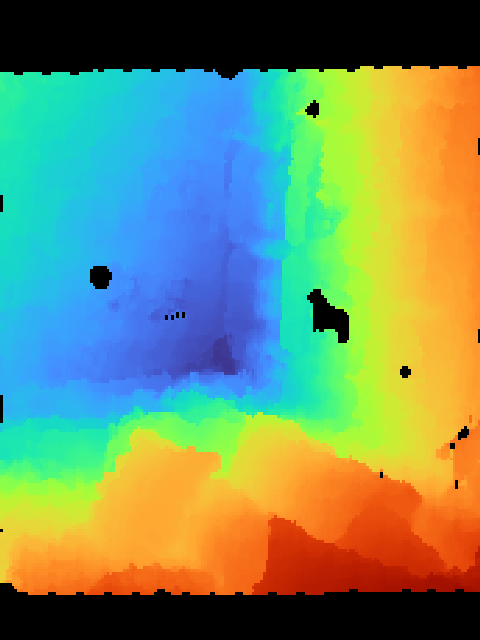} & 
        \includegraphics[width=\qualimwidth]{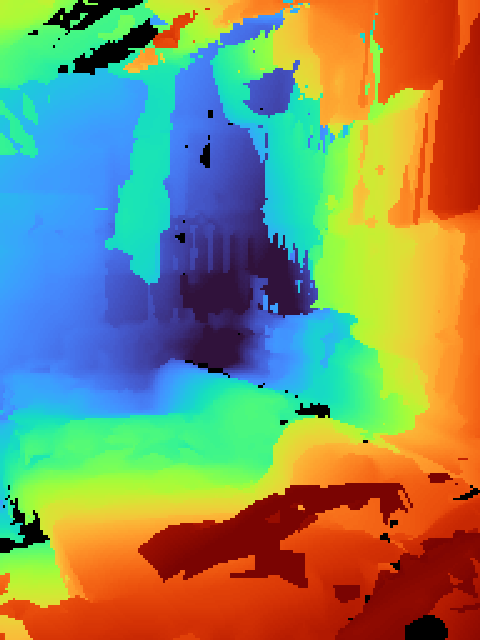} &
        \includegraphics[width=\qualimwidth]{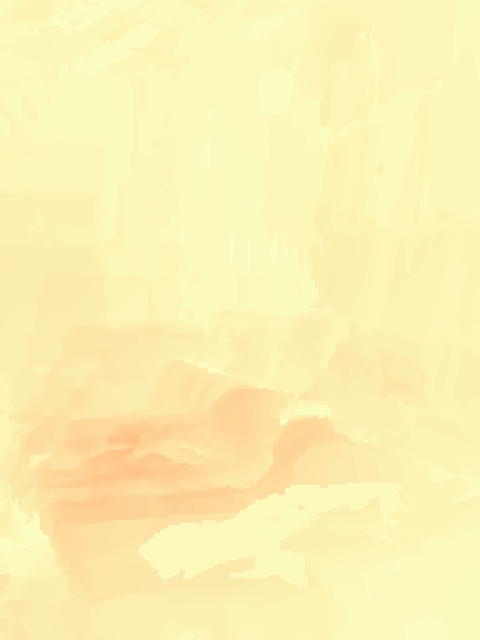} &
        \includegraphics[width=\qualimwidth]{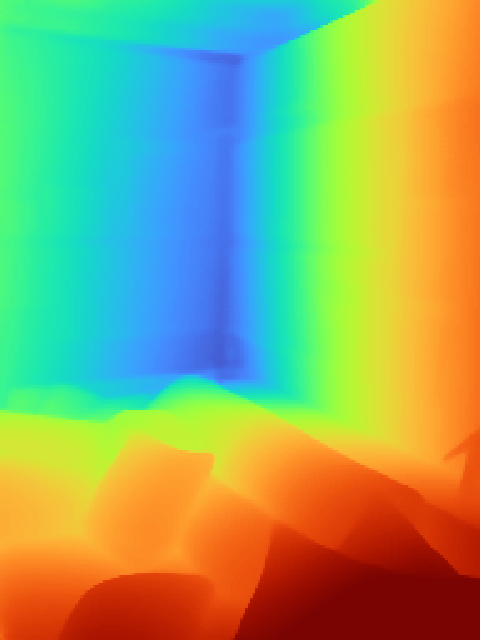} & 
        \includegraphics[width=\qualimwidth]{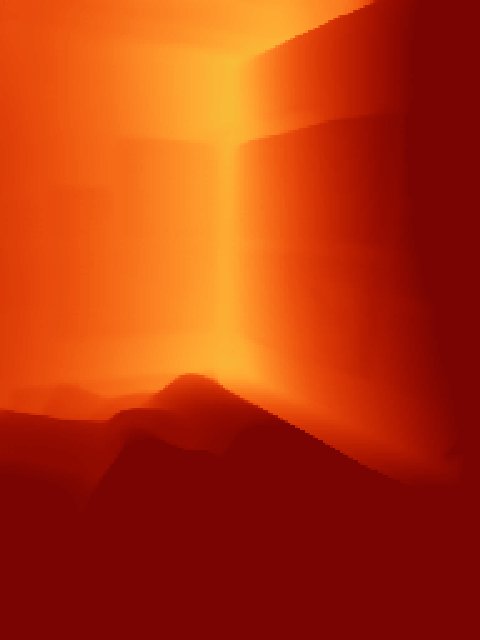} \\
        \includegraphics[width=\qualimwidth]{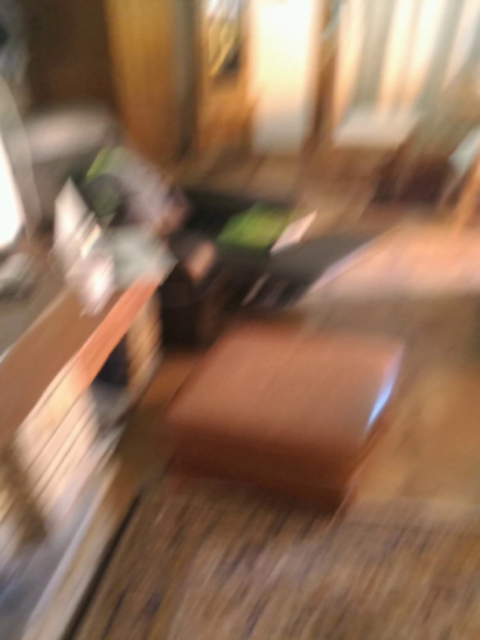} & 
        \includegraphics[width=\qualimwidth]{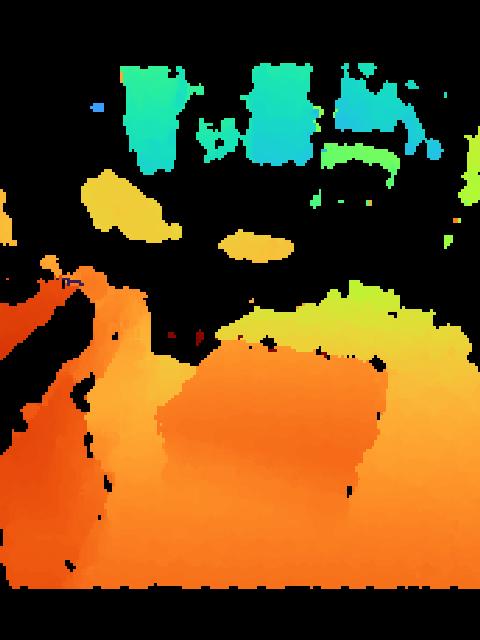} & 
        \includegraphics[width=\qualimwidth]{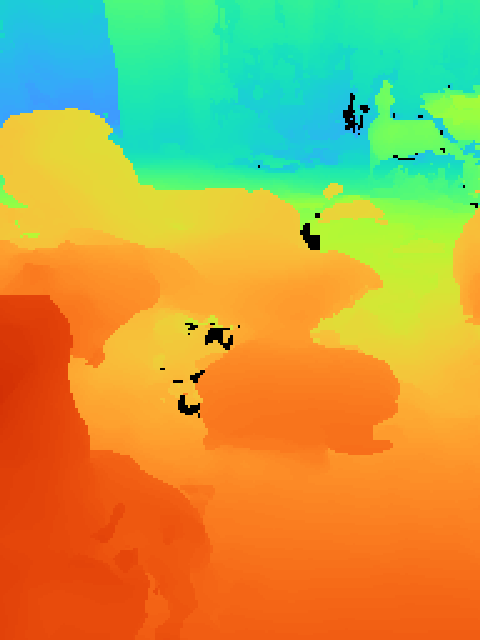} &
        \includegraphics[width=\qualimwidth]{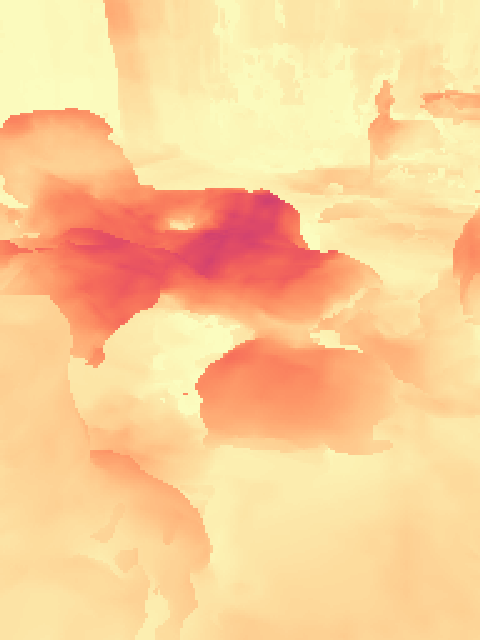} &
        \includegraphics[width=\qualimwidth]{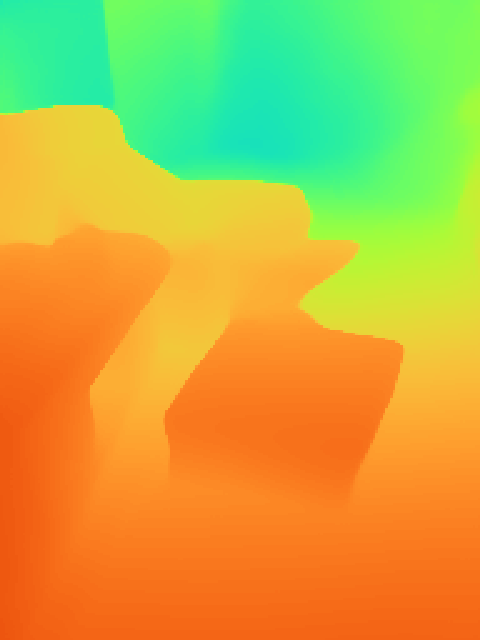} & 
        \includegraphics[width=\qualimwidth]{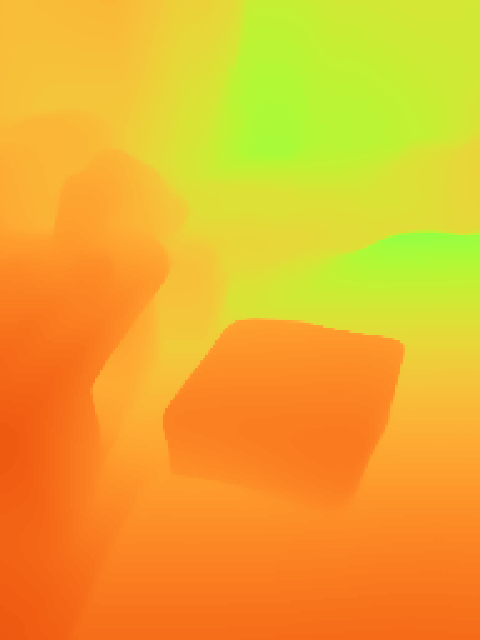} \\
    \end{tabular}
    \caption{\textbf{Qualitative results on 3RScan.} 
        Note the major improvement of our model over \cite{sayed2022simplerecon} on this challenging dataset, and how we can recover from misleading hints.}
    \label{fig:qualitative-comparison-3RScan}    
\end{figure}

\subsection{Ablations and variants}
\label{sec:ablation}

Table~\ref{tab:ablations} shows ablations and variants of our approach in \incremental mode.
Row \ref{row:ours_no_hint_mlp} doesn't use a geometry hint or a Hint MLP at all in training or evaluation, so it is functionally equivalent to \cite{sayed2022simplerecon}.
Ablations \ref{row:ours_hint_in_cv_enc} and \ref{row:ours_single_mlp} replace our Hint MLP with alternatives methods; both of these score worse than \textbf{ours} (\ref{row:ours}).
Our use of a separate Hint MLP has an additional advantage that we can cache the cost volume output for the second pass of \offline mode.
In \ref{row:ours_no_confidence} the network has no access to confidences, so it may incorrectly rely on under-construction subpar geometry. 
We then vary the format the geometry hint takes: ablation \ref{row:ours_warped_depth} forward-warps previous depth predictions to the current viewpoint.
\ref{row:ours_variable_depth_planes} and \ref{row:tocd} are implementations in the spirit of previous publications \cite{Xin2023ISMAR,khan2023temporally} which allow depth as input to an MVS system; see supplementary for full details.
\ref{row:ours_no_hint} is our full model but where we avoid giving a hint for any test-time frames; this performs similarly to \ref{row:ours_no_hint_mlp}, showing we are not disadvantaged in situations where no hint is available. \ref{row:ours_guided_cv_offline} and \ref{row:ours_cost_volume_modulation} uses our TSDF depth render to modulate cost volume values as in~\cite{poggi2022multi} instead of our Hint MLP.
Finally, \textbf{Ours} (\ref{row:ours}) outperforms all ablations.

\paragraph{Sensitivity to pose errors.}
In the \revisit scenario, to simulate errors in the relocalization algorithm, we add noise to the one-off alignment between the previous capture and the current capture.
The result is shown in Table~\ref{table:threerscan}, and shows our tolerance to noisy alignments.
See the supplementary for details.

\paragraph{Non-static scenes.}
Figure~\ref{fig:qualitative-comparison-3RScan} shows our system is robust to scene geometry changes after the initial visit.
Please see supplementary material for results and details showing our robustness to moving objects

\begin{figure}[t]
    \centering
    \newcommand{\qualimwidth}{0.21\textwidth}

    \newcommand{\addArrowToImage}[5]{ 
        \begin{tikzpicture}
            \node[anchor=south west,inner sep=0] (image) at (0,0) {\includegraphics[width=\qualimwidth]{#1}};
            \begin{scope}[x={(image.south east)},y={(image.north west)}]
                \draw[-latex, ultra thick, bluearrow] (#2,#3) -- (#4,#5);
            \end{scope}
        \end{tikzpicture}
    }
    
    \renewcommand{\tabcolsep}{2pt}
    \begin{tabular}{cccc}
        \scriptsize
        \centering
        \textbf{NeuralRecon~\cite{sun2021neuralrecon}} & \textbf{\scriptsize SR~\cite{sayed2022simplerecon} (online) } & \textbf{\scriptsize Ours (\incremental) } & \textbf{\scriptsize Ground truth}\\
        \includegraphics[width=\qualimwidth]{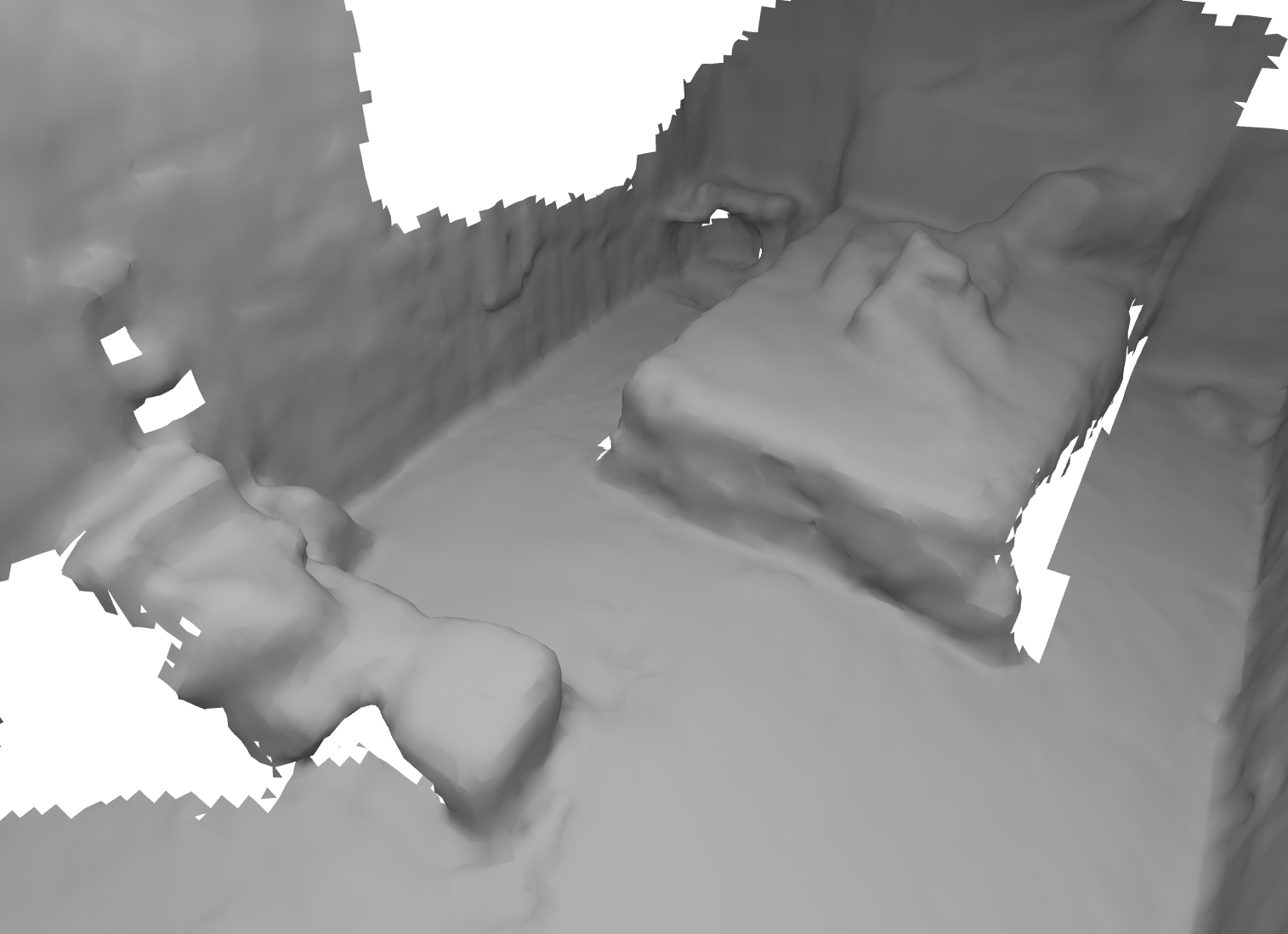} & 
        \addArrowToImage{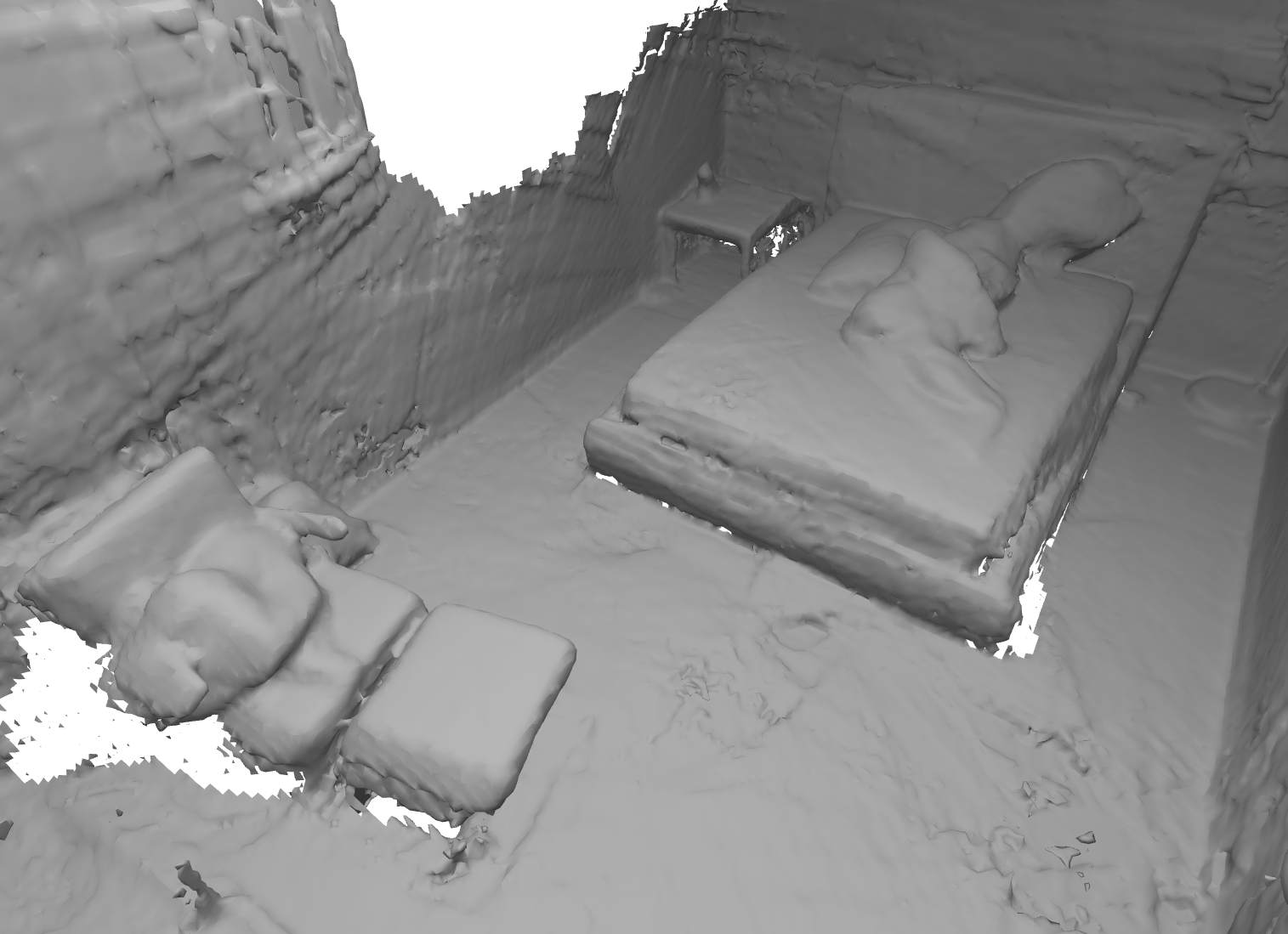}{0.4}{0.6}{0.2}{0.7} & 
        \addArrowToImage{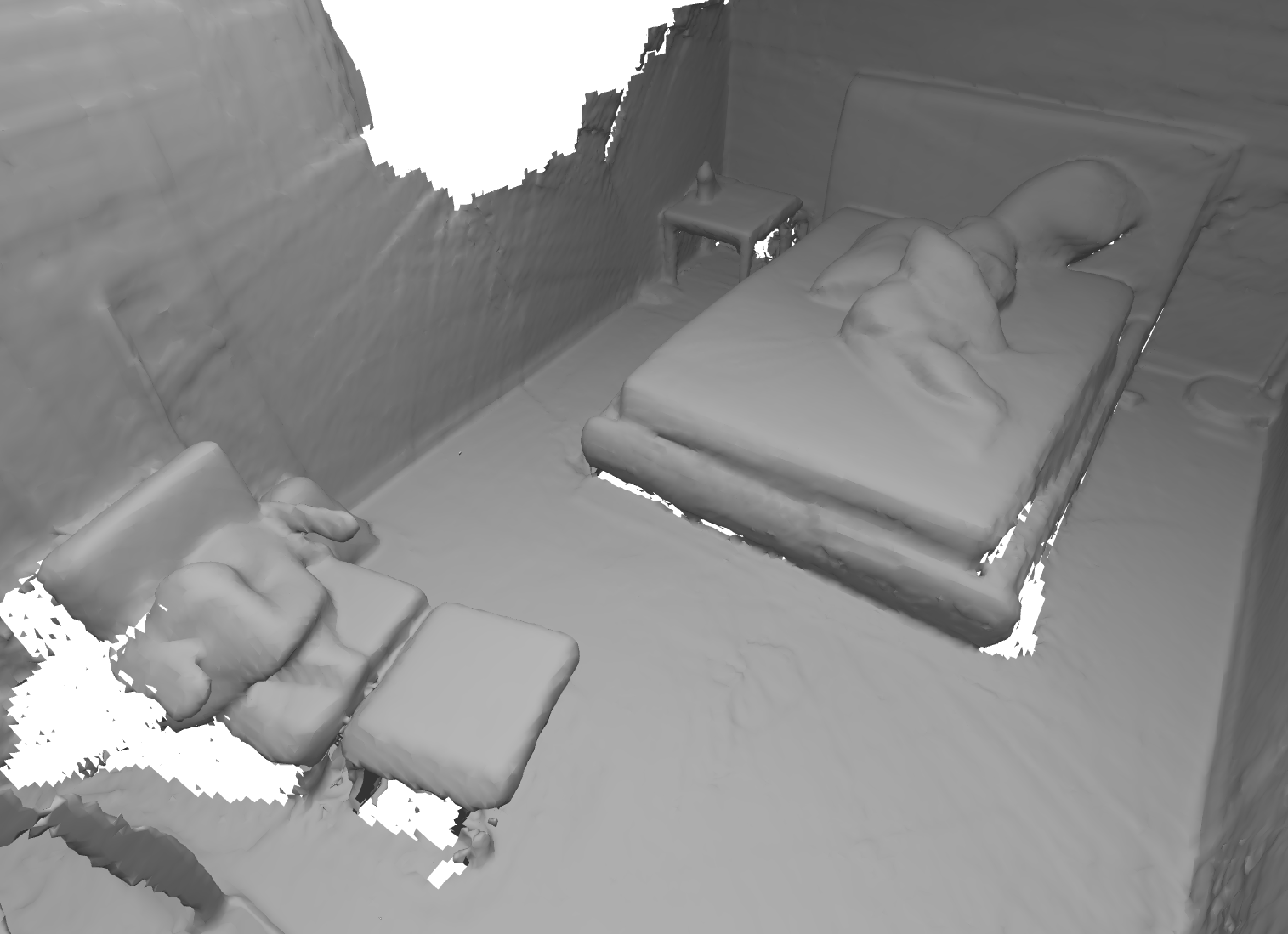}{0.4}{0.6}{0.2}{0.7} & 
        \includegraphics[width=\qualimwidth]{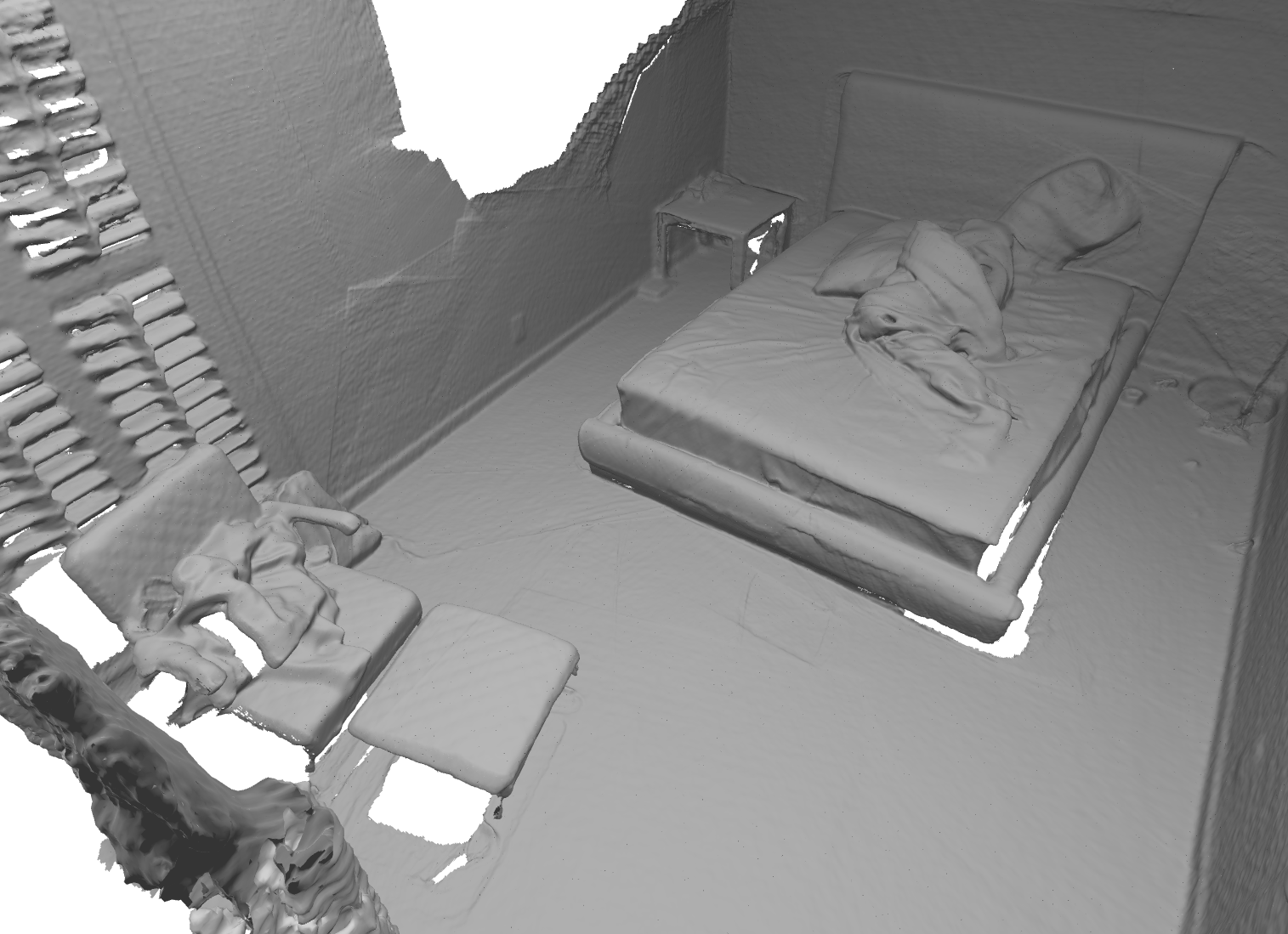} \\

        \includegraphics[width=\qualimwidth]{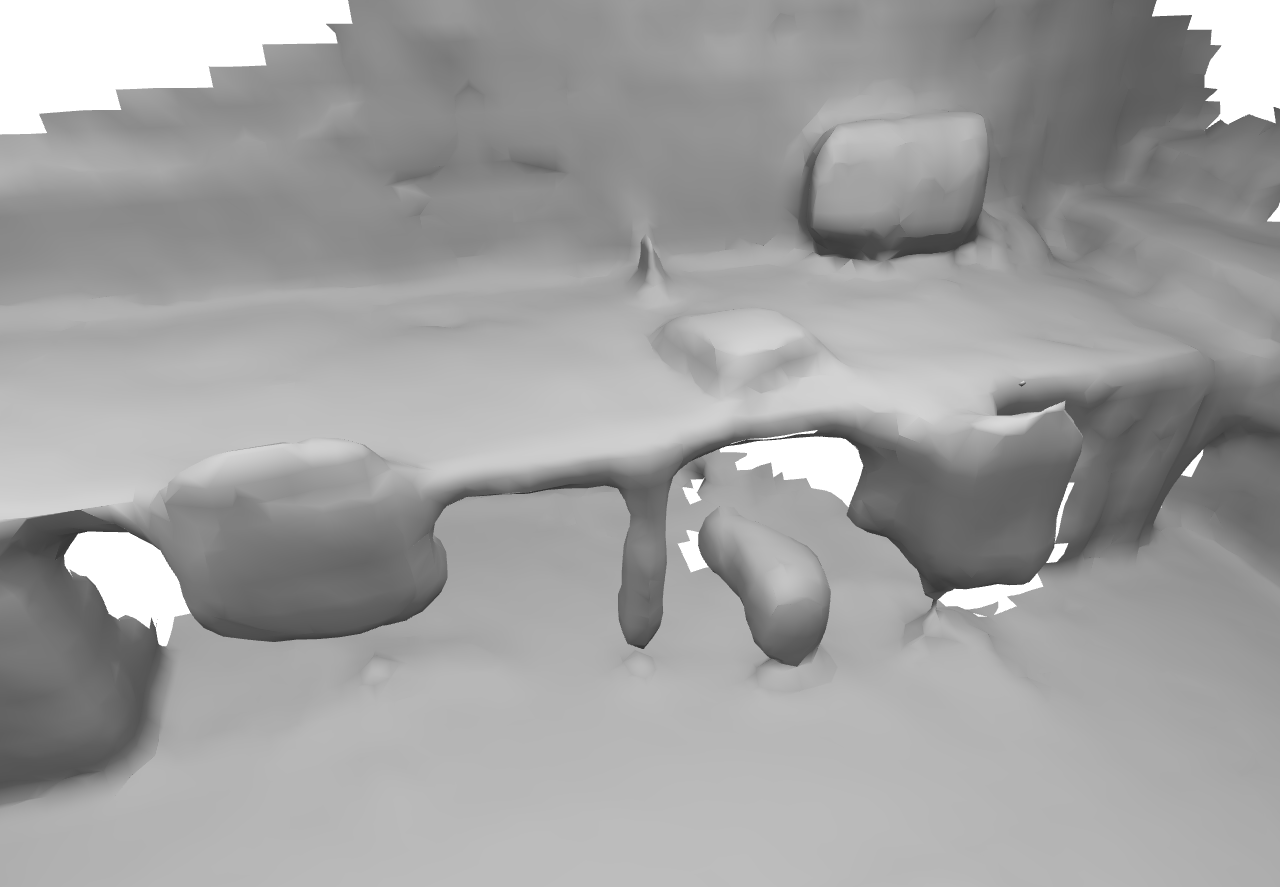} & 
        \addArrowToImage{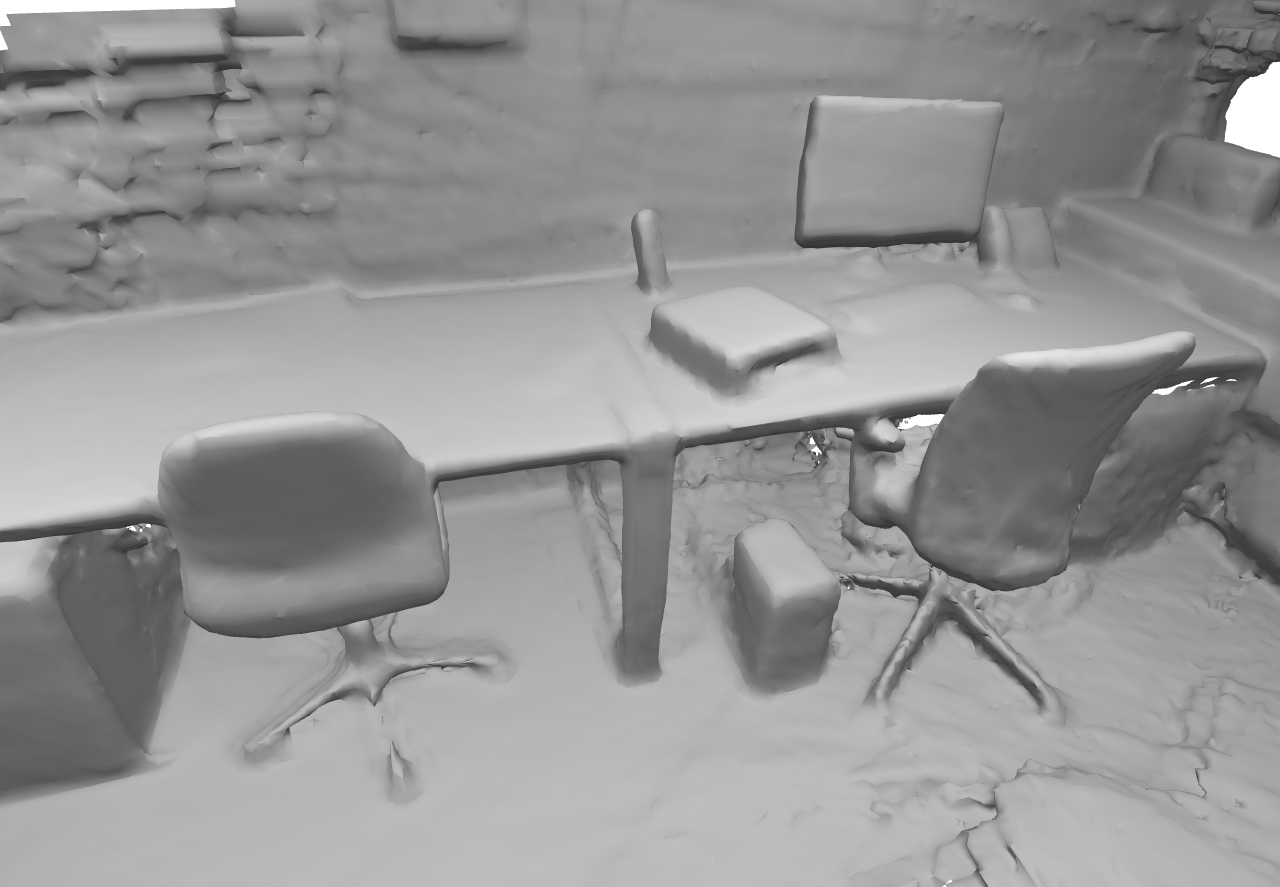}{0.52}{0.15}{0.32}{0.15} & 
        \addArrowToImage{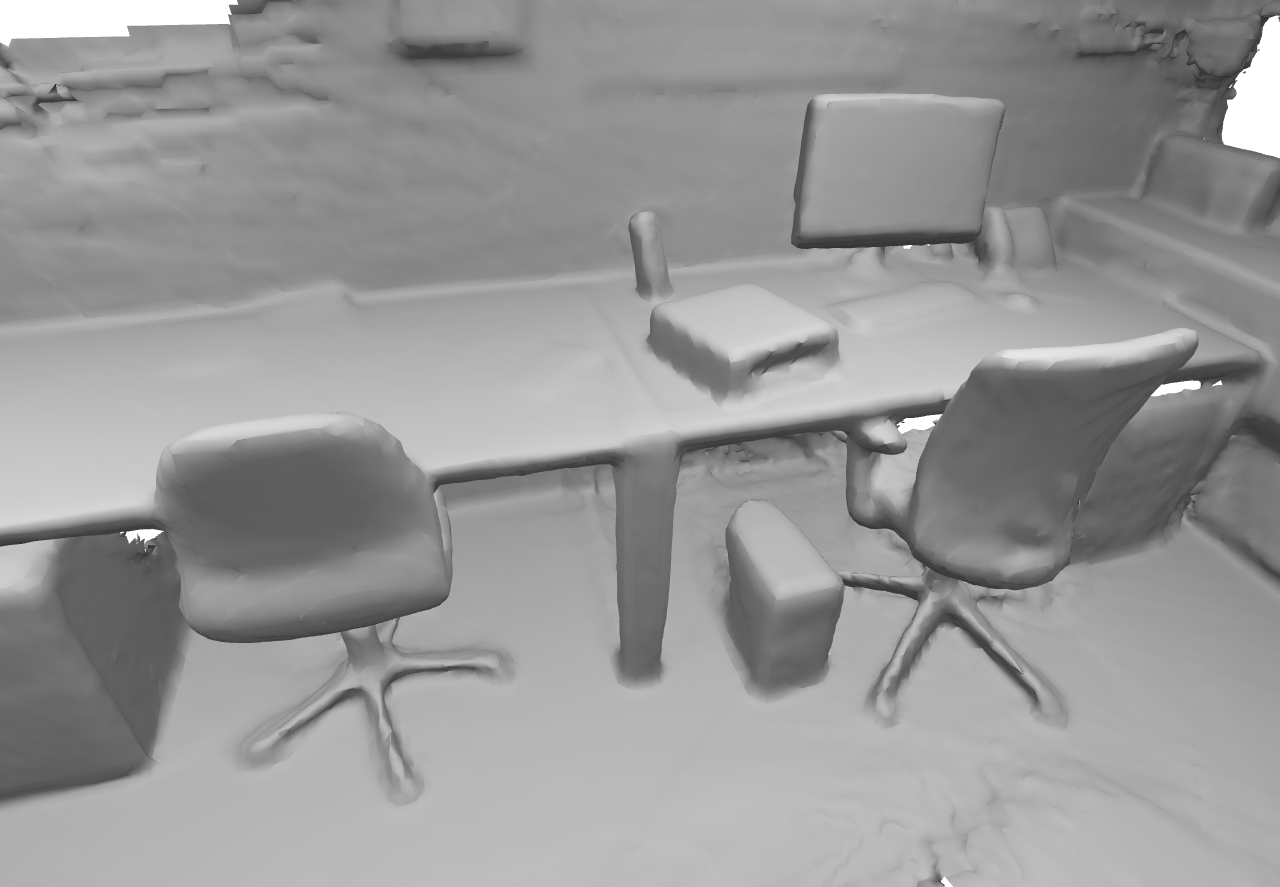}{0.52}{0.15}{0.32}{0.15} & 
        \includegraphics[width=\qualimwidth]{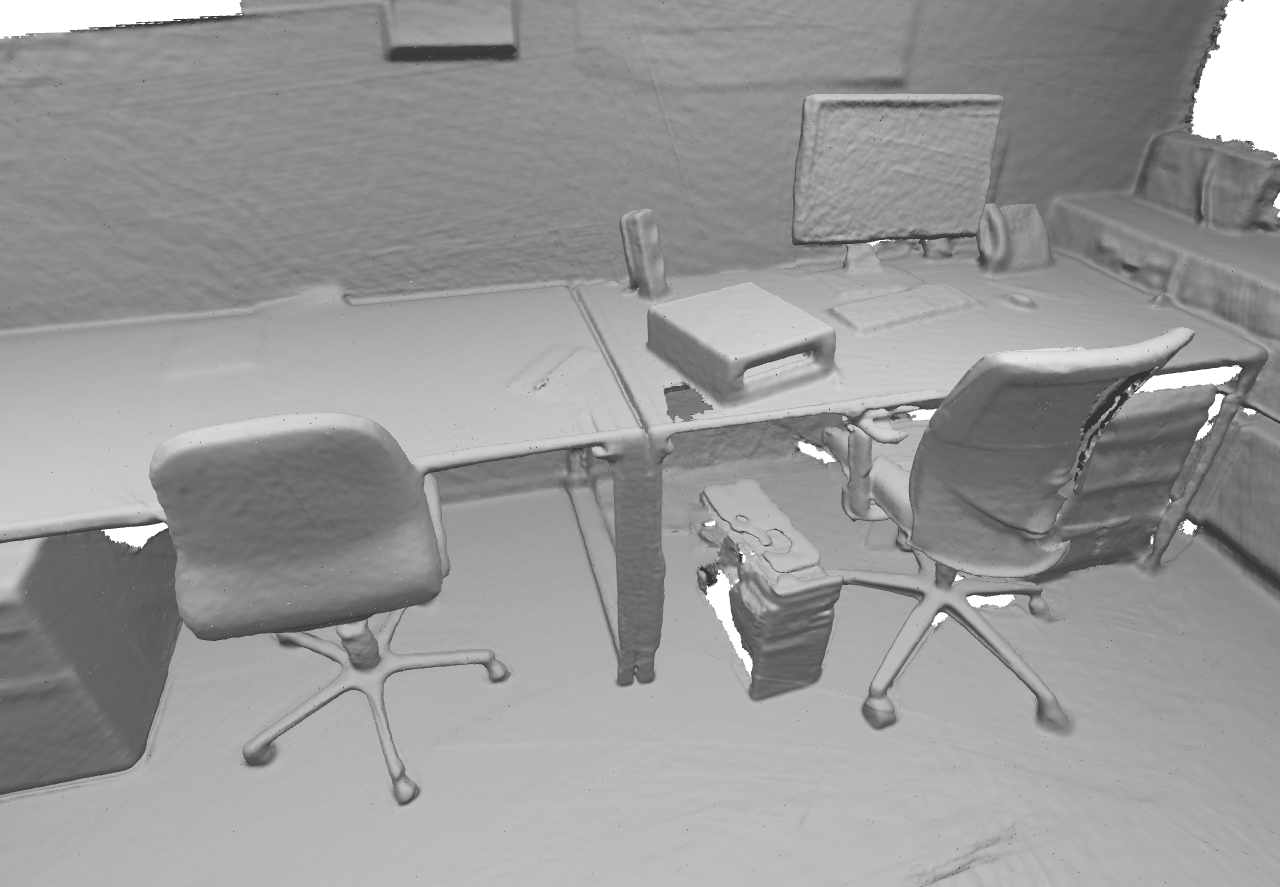} \\

        \includegraphics[width=\qualimwidth]{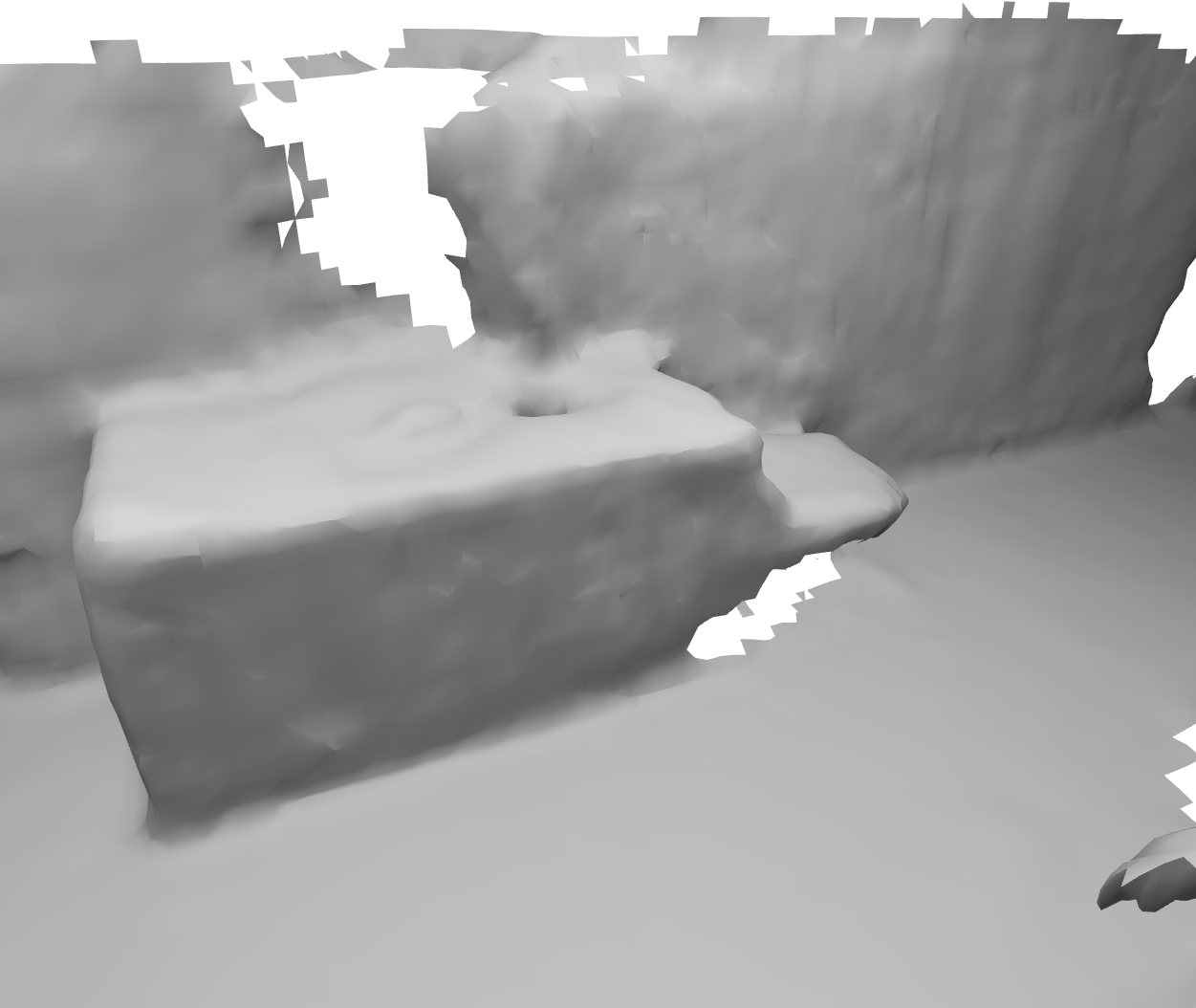} & 
        \addArrowToImage{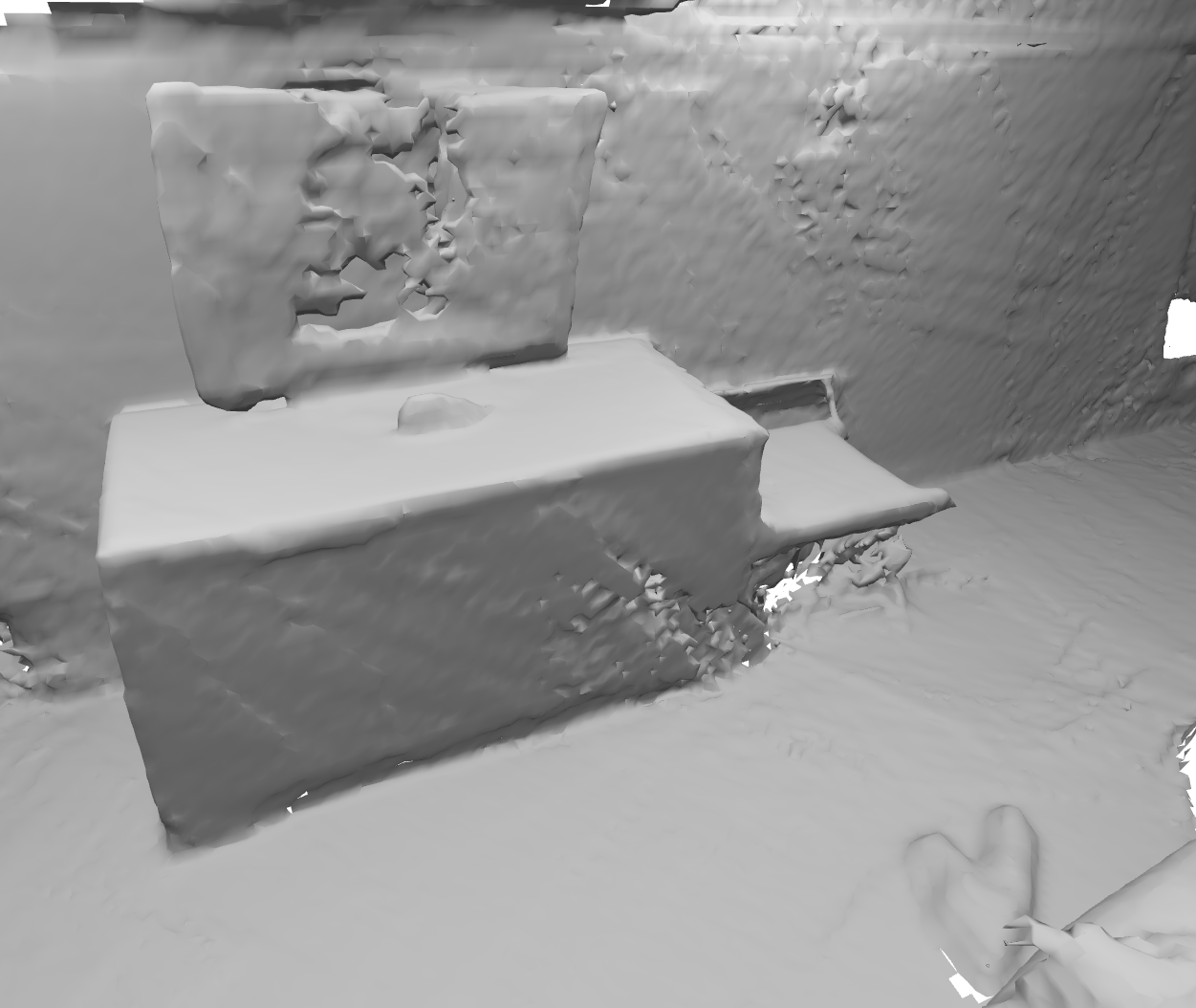}{0.65}{0.8}{0.45}{0.8} & 
        \addArrowToImage{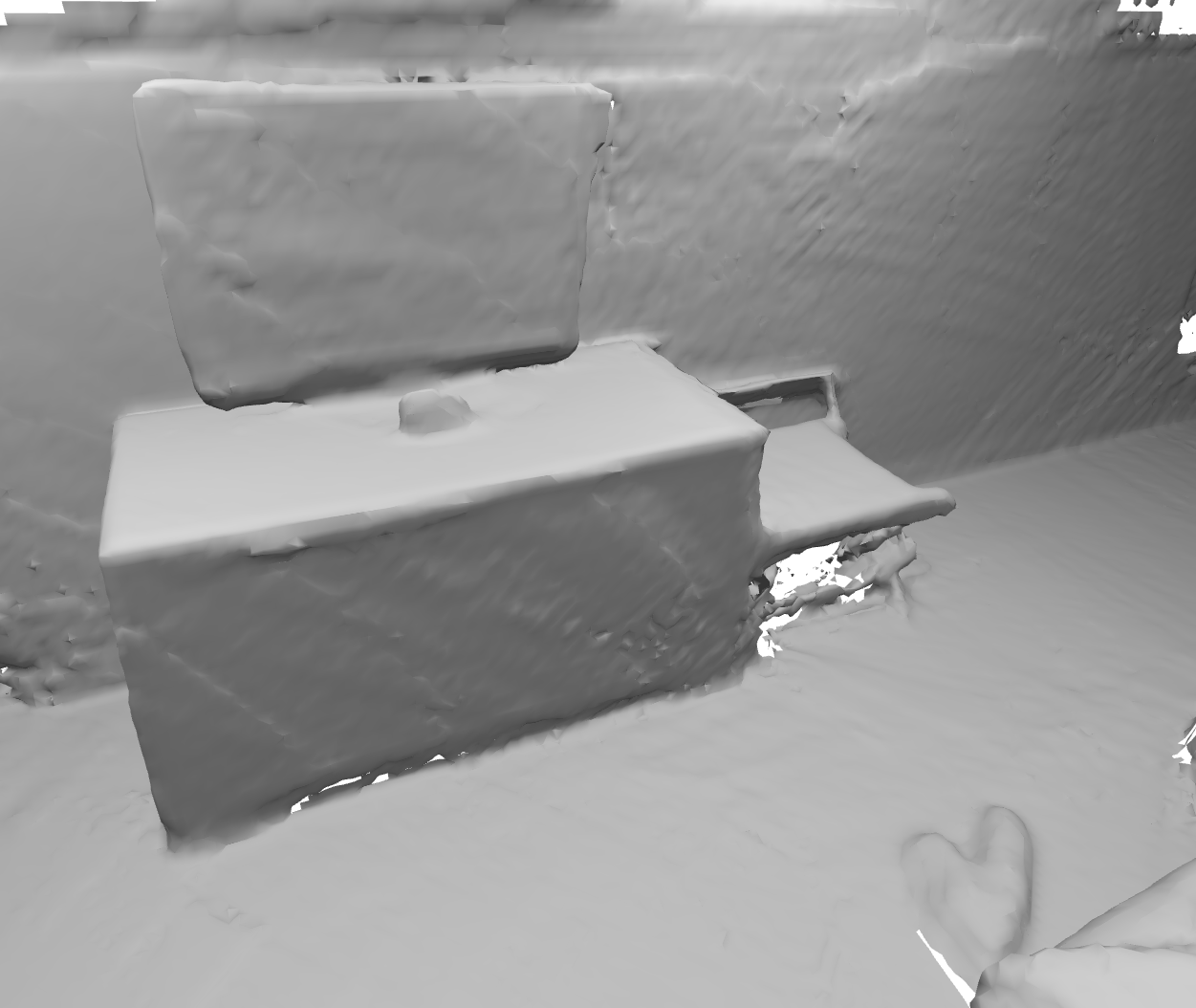}{0.65}{0.8}{0.45}{0.8} & 
        \addArrowToImage{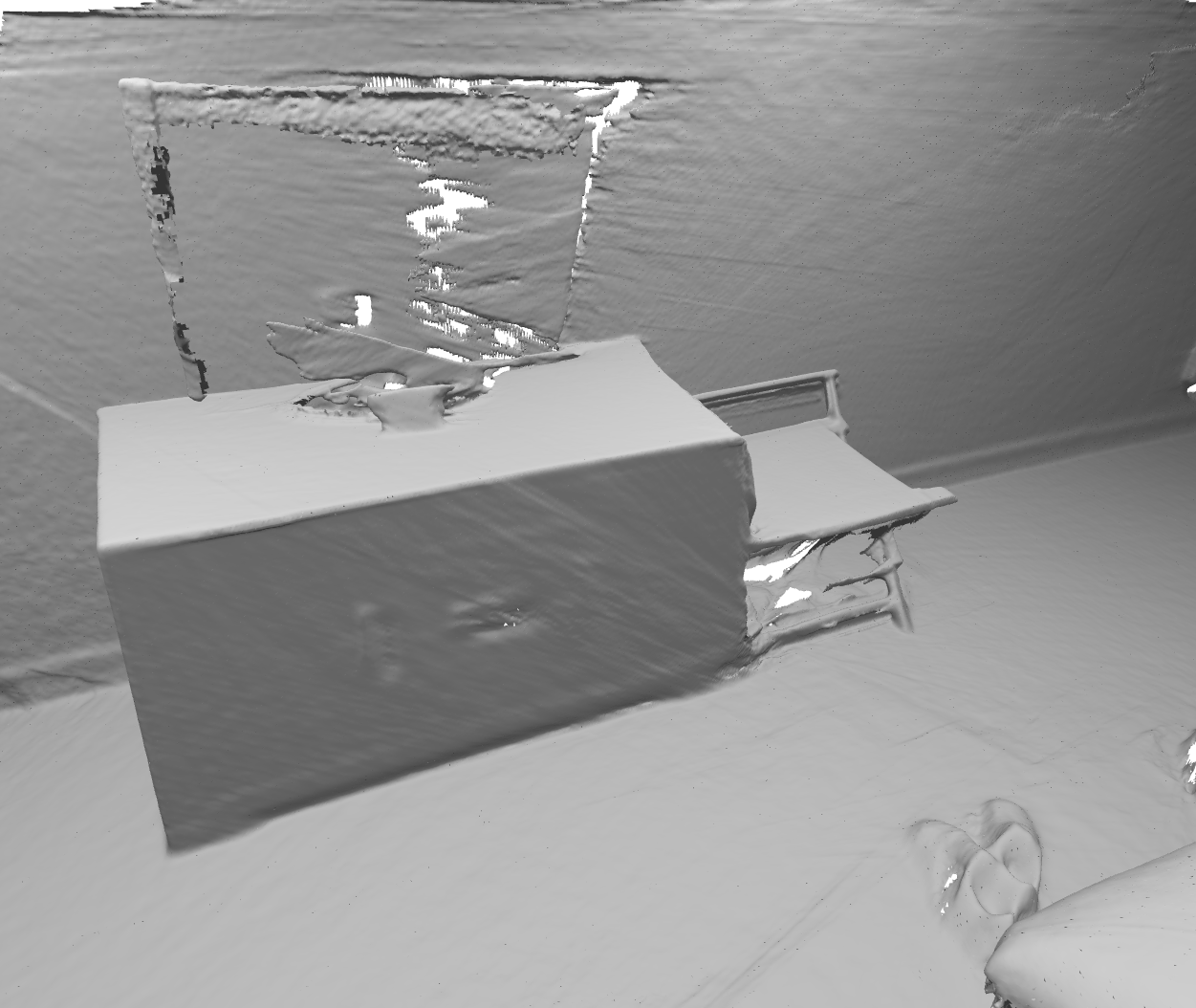}{0.65}{0.8}{0.45}{0.8} \\
    \end{tabular}
    \caption{\textbf{Qualitative results for meshing (\incremental).} Our method gives higher quality meshes than baselines \eg~\cite{sayed2022simplerecon} by using existing geometry for future predictions. 
    }
    \label{fig:qualitative-comparison-online} 
\end{figure}

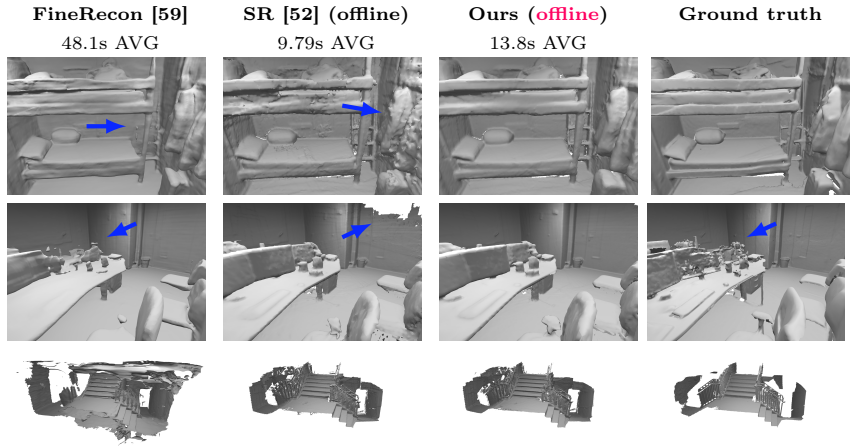
\begin{figure}[h]
    \centering
    \newcommand{\qualimwidth}{0.215\textwidth}
    \renewcommand{\tabcolsep}{2pt}

    \newcommand{\addArrowToImage}[5]{ 
        \begin{tikzpicture}
            \node[anchor=south west,inner sep=0] (image) at (0,0) {\includegraphics[width=\qualimwidth]{#1}};
            \begin{scope}[x={(image.south east)},y={(image.north west)}]
                \draw[-latex, ultra thick, bluearrow] (#2,#3) -- (#4,#5);
            \end{scope}
        \end{tikzpicture}
    }
    \begin{tabular}{cccc}
        \centering
        \textbf{\scriptsize FineRecon~\cite{Stier_2023_ICCV}} & \textbf{\scriptsize SR~\cite{sayed2022simplerecon} (offline)} & \textbf{\scriptsize Ours (\offline)} & \textbf{\scriptsize Ground truth}\\
        \scriptsize 48.1s AVG & \scriptsize 9.79s AVG & \scriptsize  13.8s AVG & \\
        \addArrowToImage{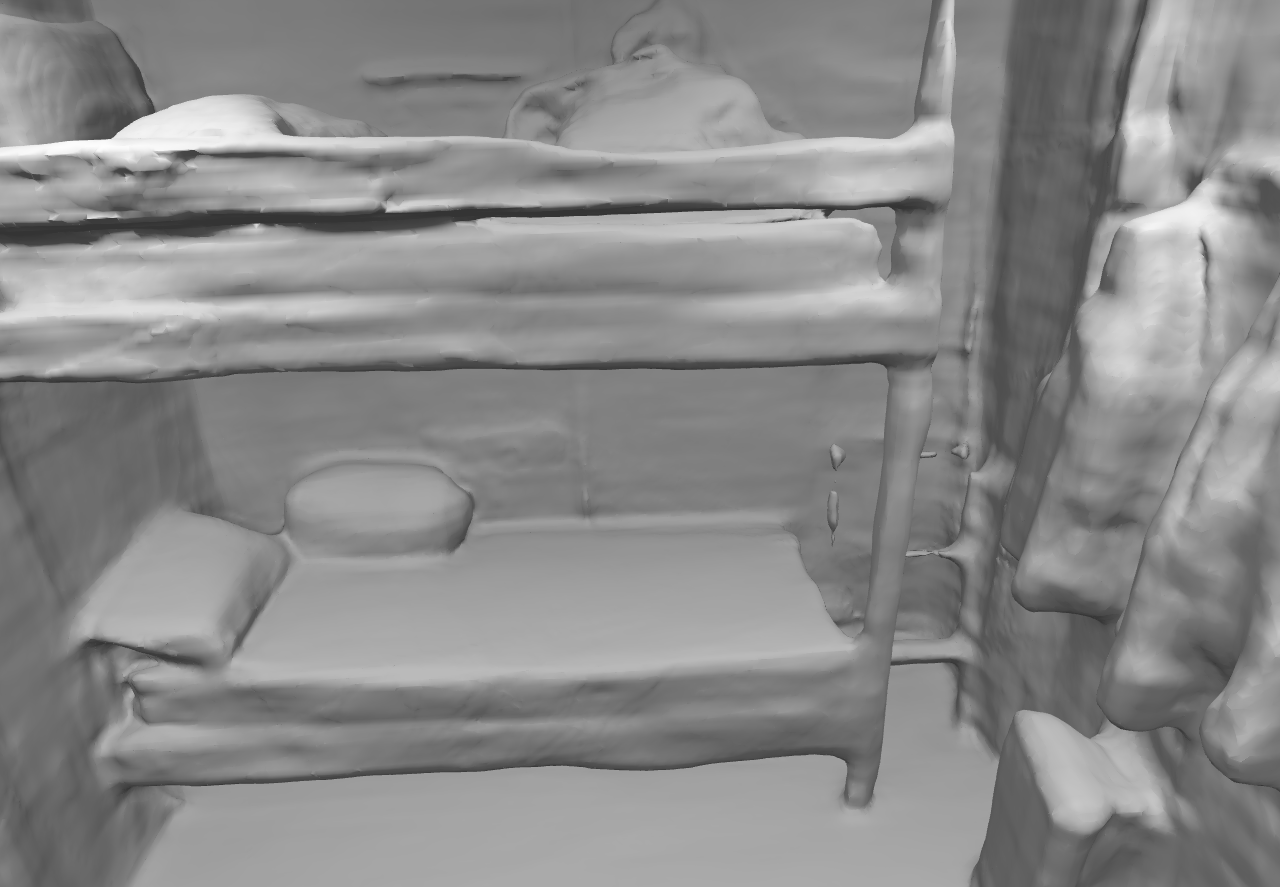}{0.4}{0.5}{0.6}{0.5} & 
        \addArrowToImage{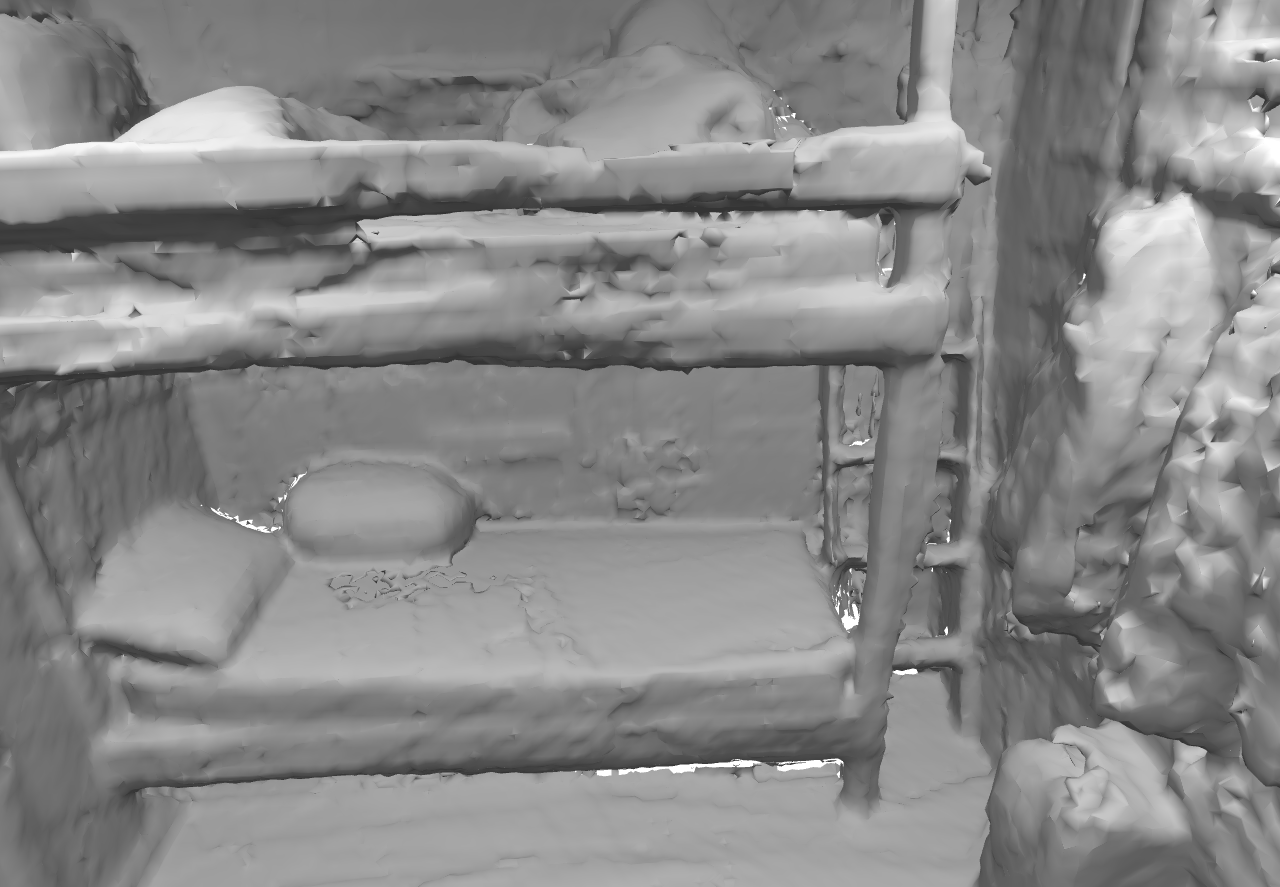}{0.60}{0.65}{0.8}{0.6} & 
        \includegraphics[width=\qualimwidth]{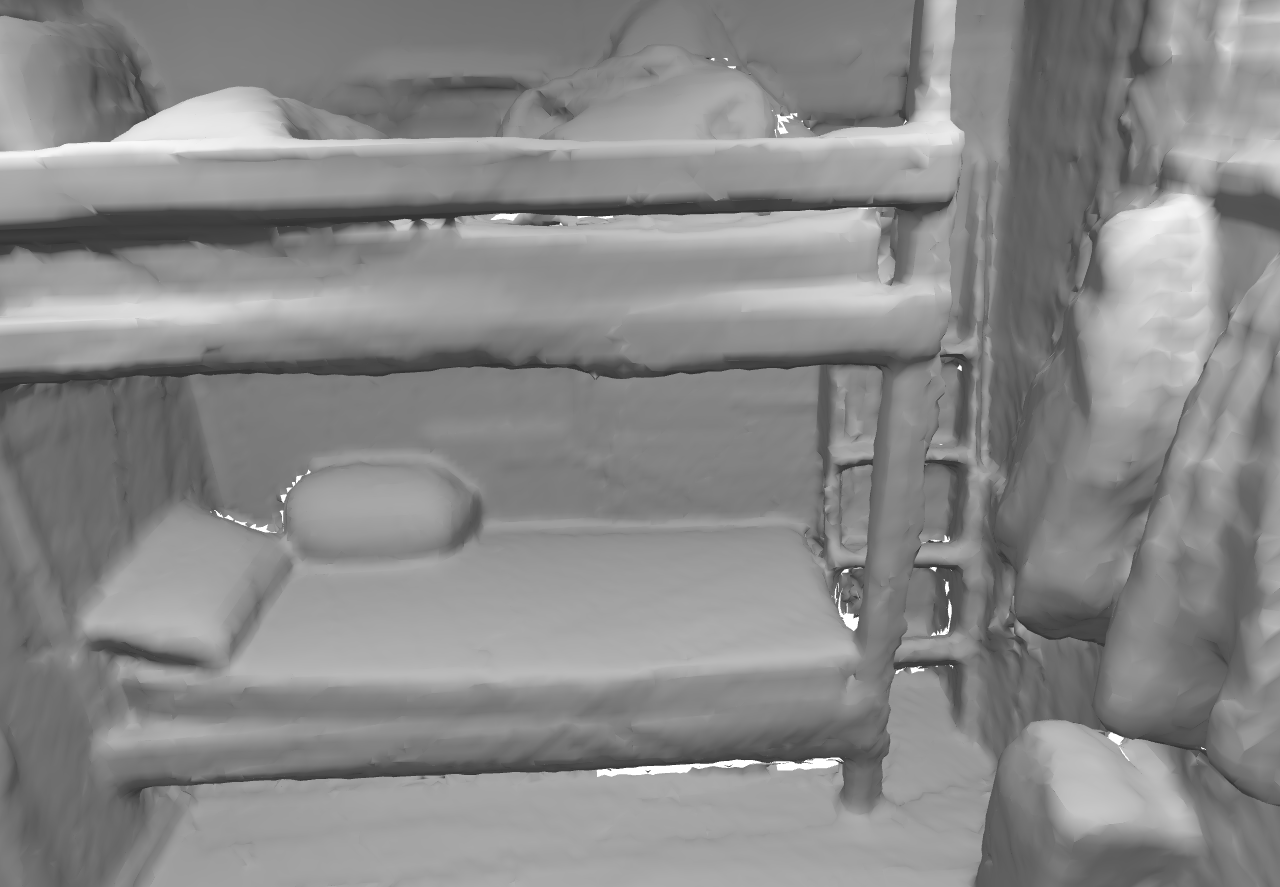} & 
        \includegraphics[width=\qualimwidth]{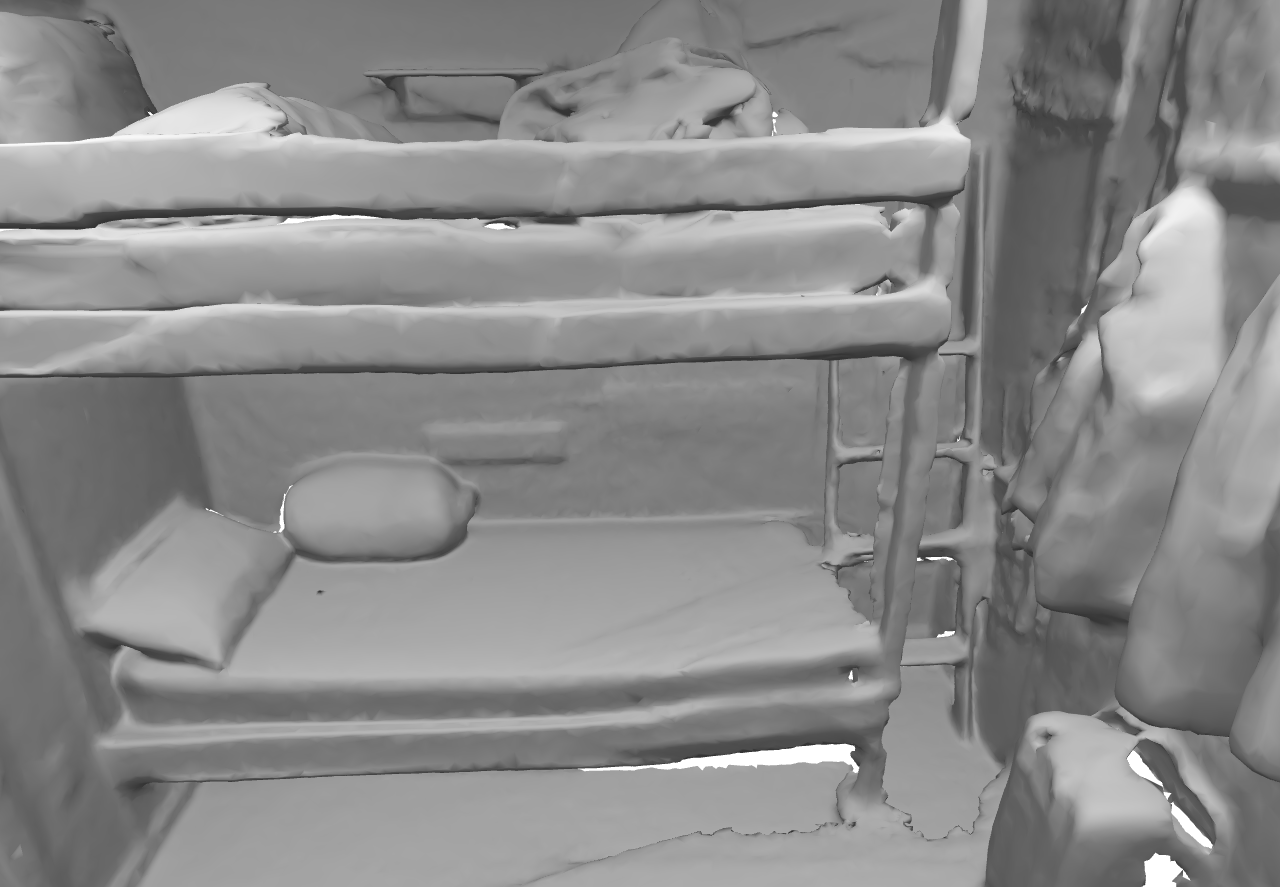}\\

        \addArrowToImage{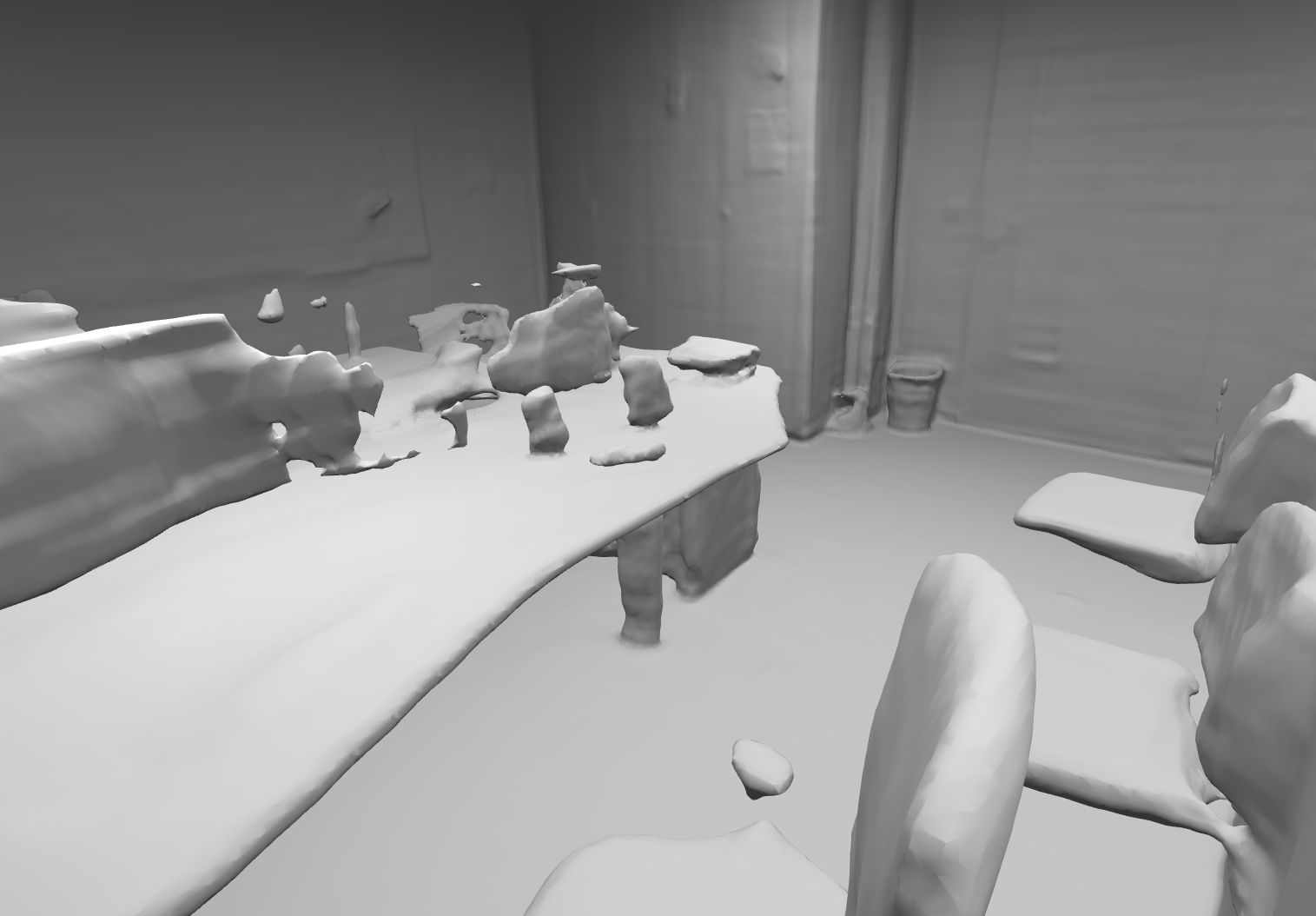}{0.65}{0.85}{0.5}{0.75} & 
        \addArrowToImage{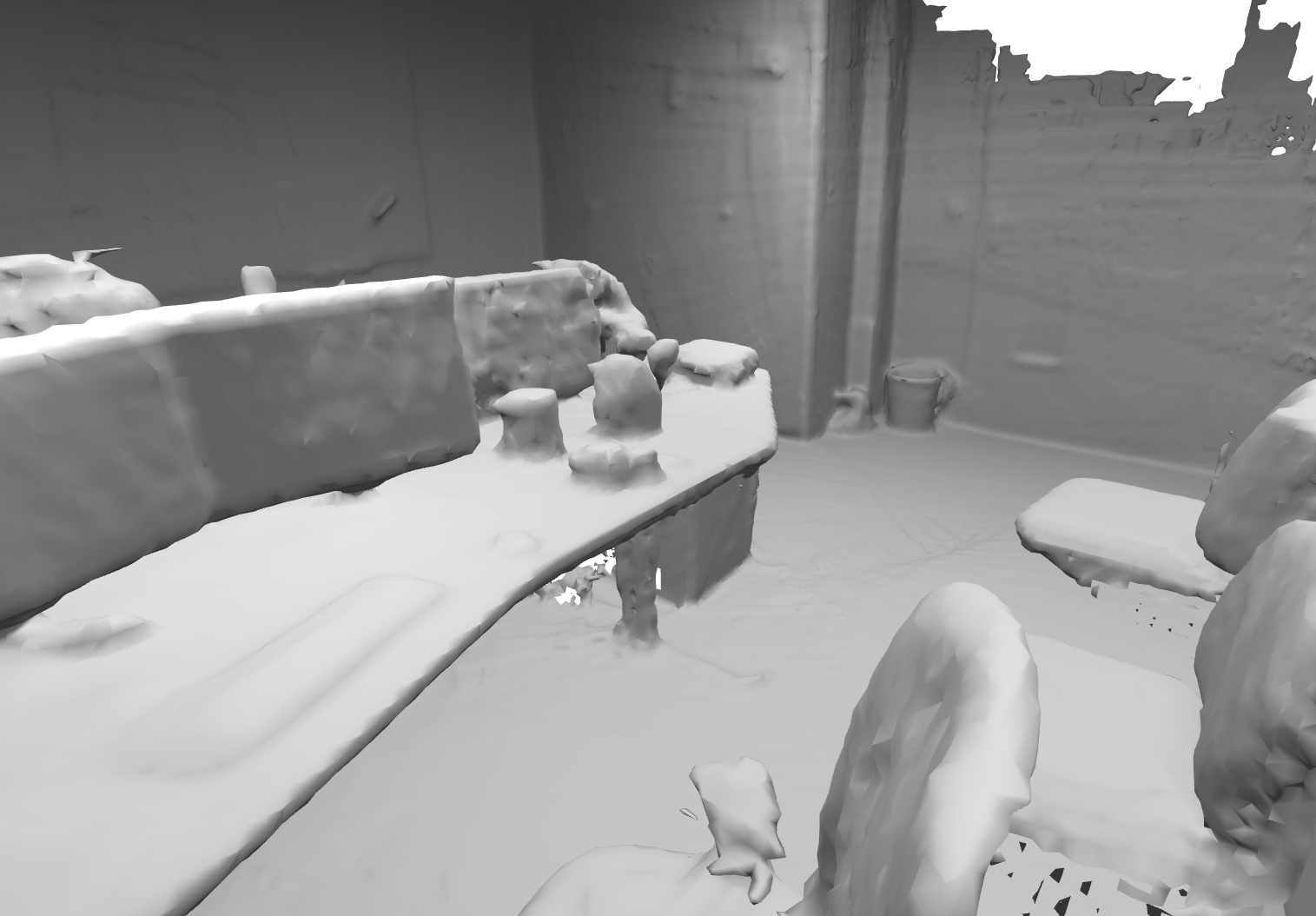}{0.6}{0.75}{0.75}{0.85} & 
        \includegraphics[width=\qualimwidth]{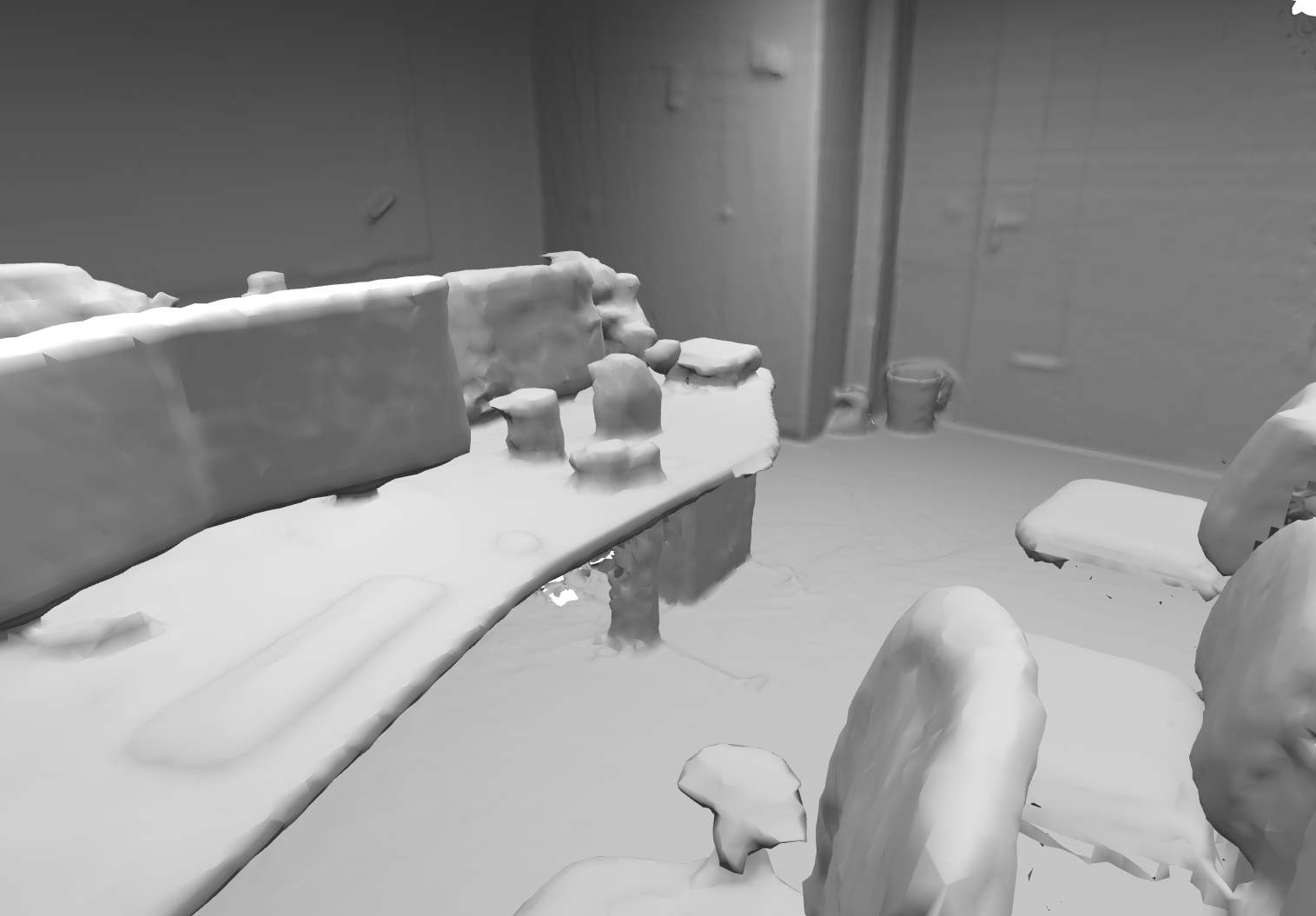} & 
        \addArrowToImage{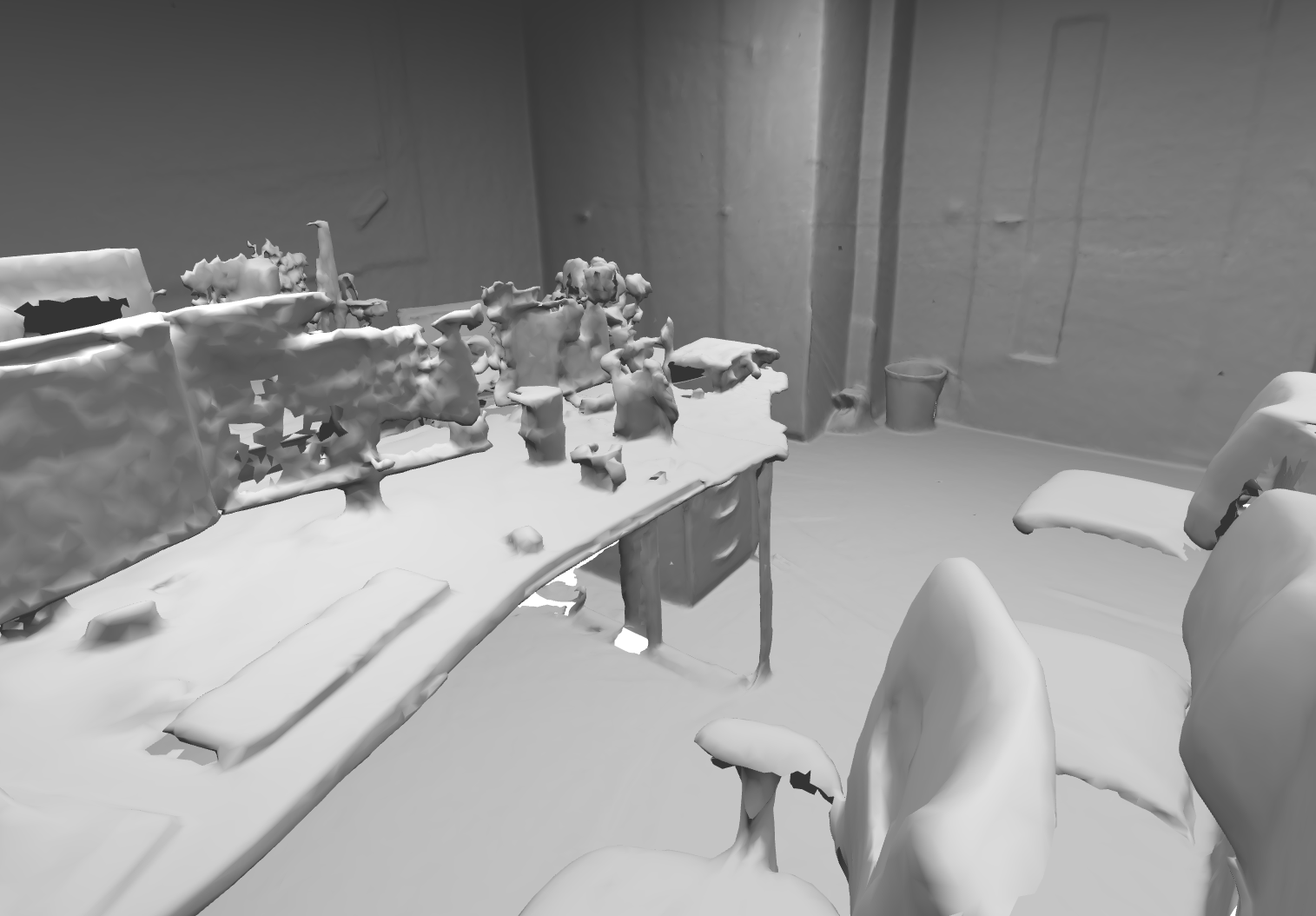}{0.65}{0.85}{0.5}{0.75} \\

        \includegraphics[width=\qualimwidth]{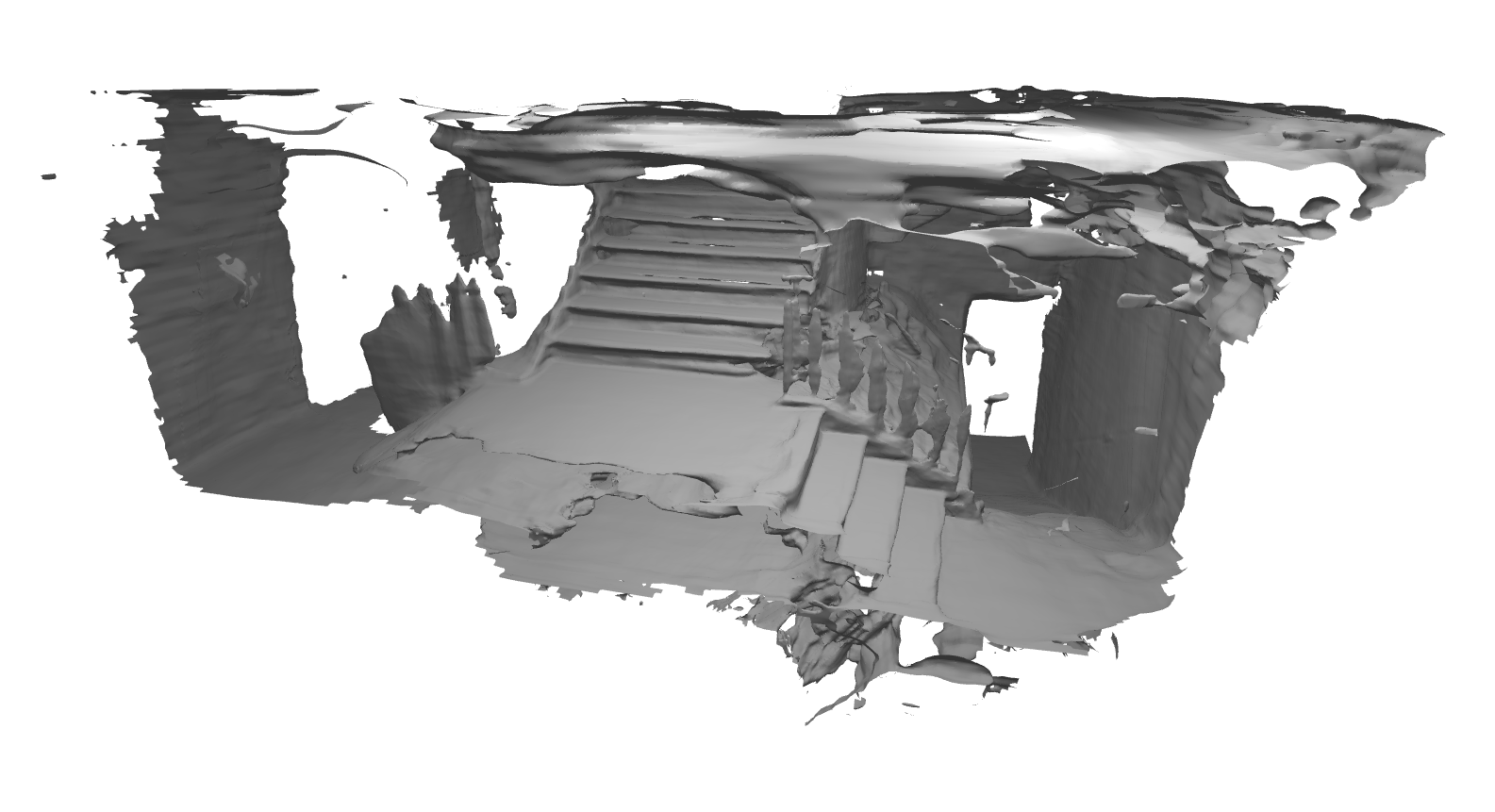} & 
        \includegraphics[width=\qualimwidth]{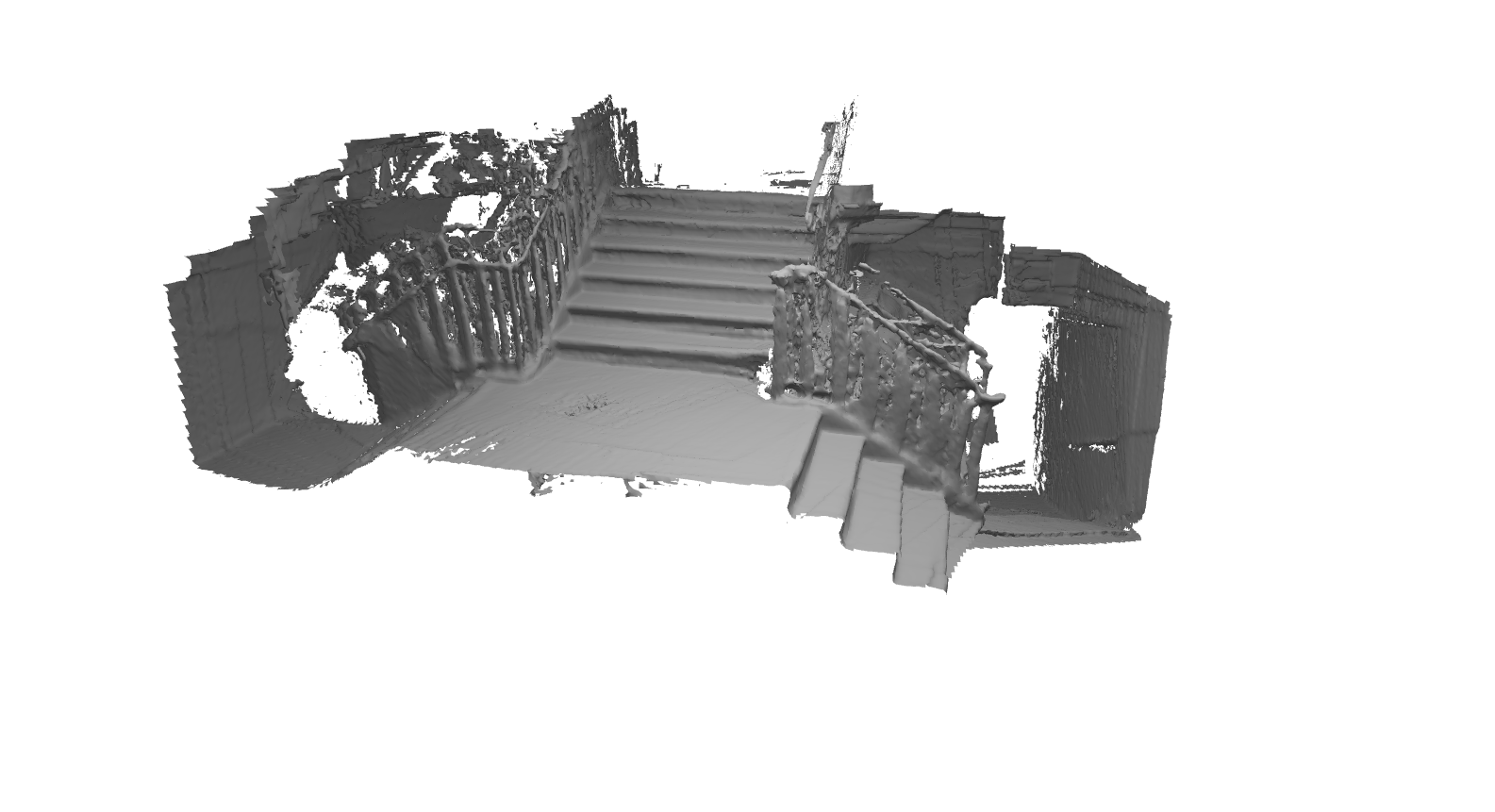} & 
        \includegraphics[width=\qualimwidth]{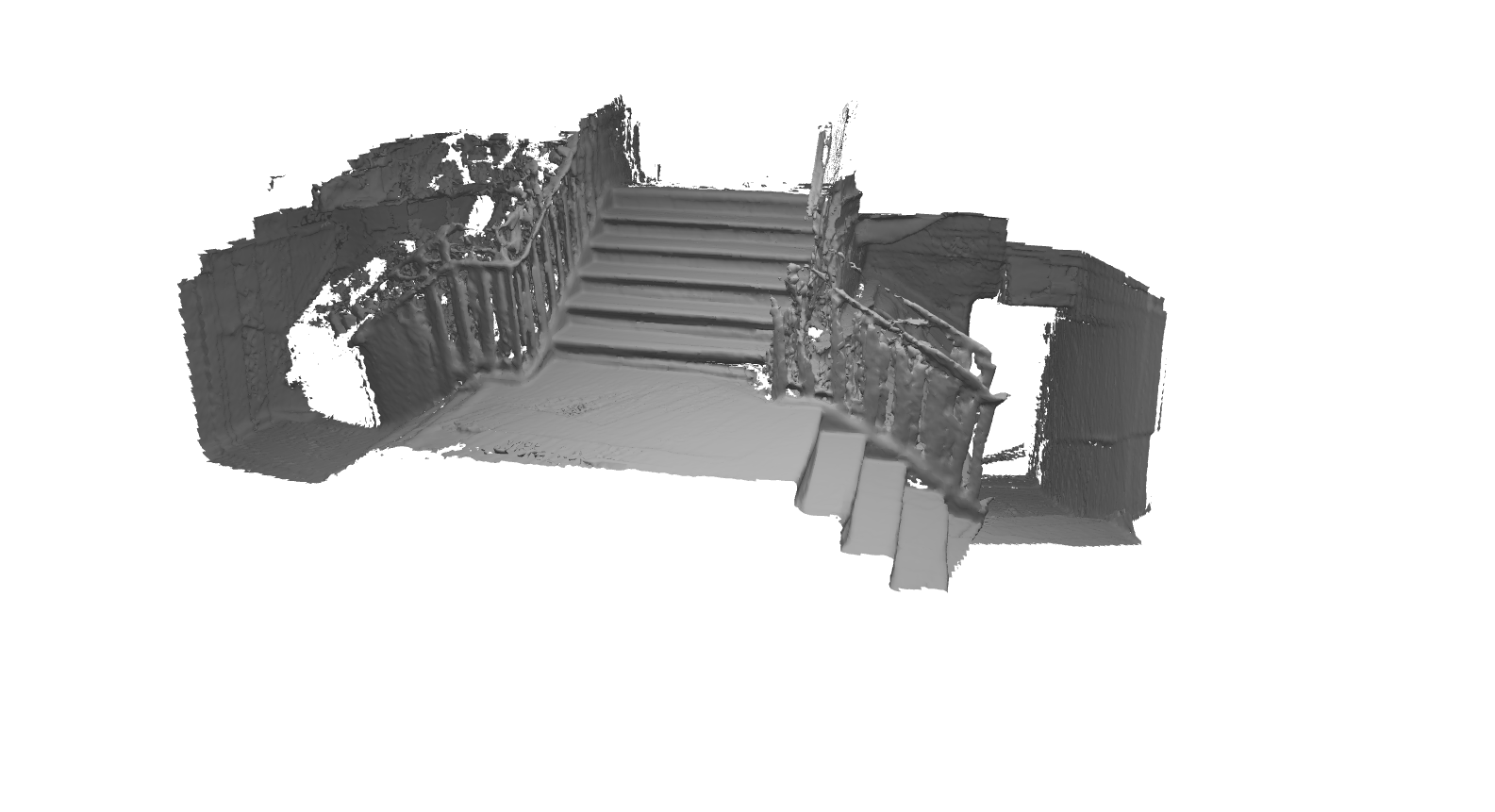} & 
        \includegraphics[width=\qualimwidth]{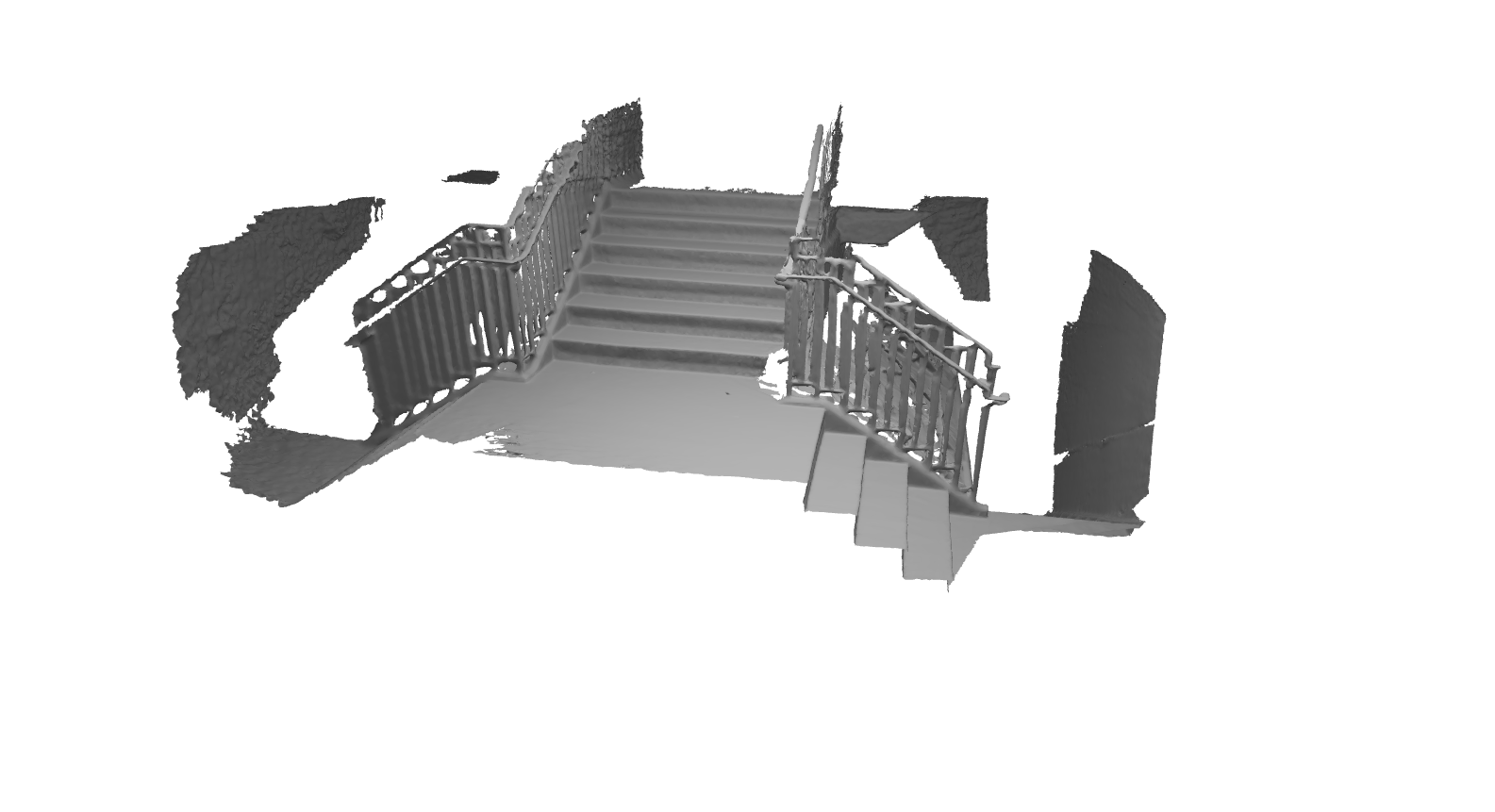} \\

    \end{tabular}
    \caption{\textbf{Qualitative results for meshing (\offline, with two passes).} Our method enables higher quality meshes than baselines \eg~\cite{sayed2022simplerecon} by running it twice.}
    \label{fig:qualitative-comparison-two-pass}   
\end{figure}

\begin{SCfigure}[1.2][h]
  \centering
  \includegraphics[width=0.7\textwidth]{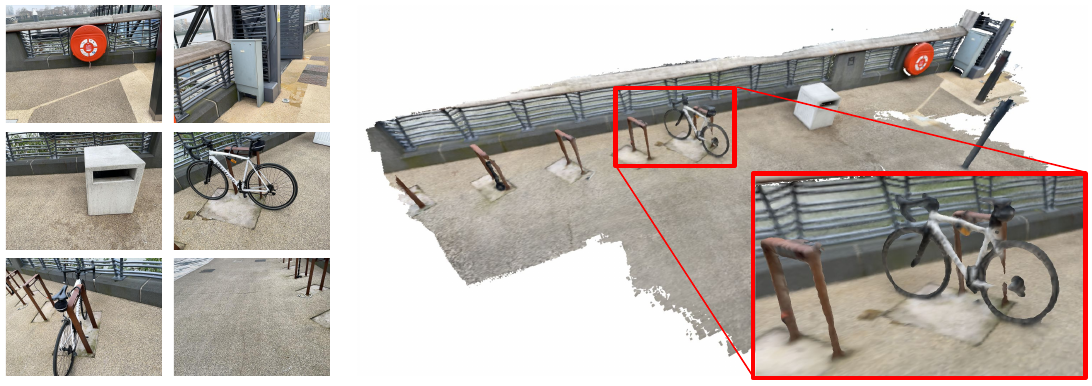}
  \caption{Our method generalizes to new domains
  \eg this outside scene, casually captured with a smartphone. Poses are from ARKit~\cite{arkit}.}
\end{SCfigure}

\subsection{Limitations}

Like other methods which reconstruct 3D scenes by fusing depth maps \eg \cite{sayed2022simplerecon}, we only reconstruct geometry we have directly observed, and do not complete any hidden or out-of-frustum surfaces.
Since we limit to $3.5$m the depths fused into the TSDF, our method has the most benefits when revisiting an area that has been previously observed close by. 
For example, if the camera is moving forward through a long corridor, our proposed geometric hints would only cover a small region of the image.
Similarly to other multi-view stereo based methods, our method struggles with transparent and reflective surfaces.


\section{Conclusion}

We introduced a system for depth estimation and 3D reconstruction which can take as input, where available, previously-made estimates of the scene's geometry.
Our carefully-designed architecture takes as input geometrical hints, and makes high quality depth estimates even when these hints are not available.
We showed how our method can make use of hints from the near-term for instantaneous depths in new environments, and also hints from the past when estimating depths in previously visited locations.
We evaluated on a range of challenging datasets where we have obtained state-of-the-art depths and reconstructions. 




\clearpage
\section*{Acknowledgements}
We'd like to thank the Niantic Raptor R\&D infrastructure team - Saki
Shinoda, Jakub Powierza, and Stanimir Vichev - for their valuable infrastructure support.
%
%
\bibliographystyle{splncs04}
\bibliography{egbib}

\begin{thebibliography}{10}
\providecommand{\url}[1]{\texttt{#1}}
\providecommand{\urlprefix}{URL }
\providecommand{\doi}[1]{https://doi.org/#1}

\bibitem{arkit}
Apple: {ARKit} (2023), \url{https://developer.apple.com/documentation/arkit}, {Accessed: 5 October 2023}

\bibitem{bozic2021transformerfusion}
Bozic, A., Palafox, P., Thies, J., Dai, A., Nie{\ss}ner, M.: {TransformerFusion}: Monocular {RGB} scene reconstruction using transformers. NeurIPS  (2021)

\bibitem{cai2023riav}
Cai, C., Ji, P., Yan, Q., Xu, Y.: {RIAV-MVS}: Recurrent-indexing an asymmetric volume for multi-view stereo. In: CVPR (2023)

\bibitem{casser2018depth}
Casser, V., Pirk, S., Mahjourian, R., Angelova, A.: Depth prediction without the sensors: Leveraging structure for unsupervised learning from monocular videos. In: AAAI (2019)

\bibitem{chang2018pyramid}
Chang, J.R., Chen, Y.S.: Pyramid stereo matching network. In: CVPR (2018)

\bibitem{chen2019self}
Chen, Y., Schmid, C., Sminchisescu, C.: Self-supervised learning with geometric constraints in monocular video: Connecting flow, depth, and camera. In: ICCV (2019)

\bibitem{cheng2019learning}
Cheng, X., Wang, P., Yang, R.: Learning depth with convolutional spatial propagation network. PAMI  (2019)

\bibitem{Stereo-Matching-in-cheng-2023}
Cheng, Z., Yang, J., Li, H.: Stereo matching in time: 100+ {FPS} video stereo matching for extended reality. In: WACV (2023)

\bibitem{choe2021volumetric}
Choe, J., Joo, K., Imtiaz, T., Kweon, I.S.: Volumetric propagation network: Stereo-lidar fusion for long-range depth estimation. IEEE Robotics and Automation Letters  (2021)

\bibitem{collins1996space}
Collins, R.T.: A space-sweep approach to true multi-image matching. In: CVPR (1996)

\bibitem{conti2023sparsity}
Conti, A., Poggi, M., Mattoccia, S.: Sparsity agnostic depth completion. In: WACV (2023)

\bibitem{dai2017scannet}
Dai, A., Chang, A.X., Savva, M., Halber, M., Funkhouser, T., Nie{\ss}ner, M.: {ScanNet}: Richly-annotated {3D} reconstructions of indoor scenes. In: CVPR (2017)

\bibitem{kangle2021dsnerf}
Deng, K., Liu, A., Zhu, J.Y., Ramanan, D.: Depth-supervised {NeRF}: Fewer views and faster training for free. In: CVPR (2022)

\bibitem{du2020depthlab}
Du, R., Turner, E., Dzitsiuk, M., Prasso, L., Duarte, I., Dourgarian, J., Afonso, J., Pascoal, J., Gladstone, J., Cruces, N., et~al.: {DepthLab}: Real-time {3D} interaction with depth maps for mobile augmented reality. In: ACM Symposium on User Interface Software and Technology (2020)

\bibitem{duzceker2021deepvideomvs}
Duzceker, A., Galliani, S., Vogel, C., Speciale, P., Dusmanu, M., Pollefeys, M.: {DeepVideoMVS}: Multi-view stereo on video with recurrent spatio-temporal fusion. In: CVPR (2021)

\bibitem{fu2022geo}
Fu, Q., Xu, Q., Ong, Y.S., Tao, W.: {Geo-Neus}: Geometry-consistent neural implicit surfaces learning for multi-view reconstruction. NeurIPS  (2022)

\bibitem{furukawa2015multi}
Furukawa, Y., Hern{\'a}ndez, C.: Multi-view stereo: A tutorial, foundations and trends{\textregistered} in computer graphics and vision (2015)

\bibitem{gao2023visfusion}
Gao, H., Mao, W., Liu, M.: {VisFusion}: Visibility-aware online {3D} scene reconstruction from videos. In: CVPR (2023)

\bibitem{geiger2012we}
Geiger, A., Lenz, P., Urtasun, R.: Are we ready for autonomous driving? the kitti vision benchmark suite. In: CVPR (2012)

\bibitem{gu2020cascade}
Gu, X., Fan, Z., Zhu, S., Dai, Z., Tan, F., Tan, P.: Cascade cost volume for high-resolution multi-view stereo and stereo matching. In: CVPR (2020)

\bibitem{guedon2023sugar}
Gu{\'e}don, A., Lepetit, V.: {SuGaR}: Surface-aligned {Gaussian} splatting for efficient {3D} mesh reconstruction and high-quality mesh rendering. arXiv preprint arXiv:2311.12775  (2023)

\bibitem{guizilini2021sparse}
Guizilini, V., Ambrus, R., Burgard, W., Gaidon, A.: Sparse auxiliary networks for unified monocular depth prediction and completion. In: CVPR (2021)

\bibitem{he2016deep}
He, K., Zhang, X., Ren, S., Sun, J.: Deep residual learning for image recognition. In: Proceedings of the IEEE conference on computer vision and pattern recognition. pp. 770--778 (2016)

\bibitem{hou2019multi}
Hou, Y., Kannala, J., Solin, A.: Multi-view stereo by temporal nonparametric fusion. In: ICCV (2019)

\bibitem{huang2018deepmvs}
Huang, P.H., Matzen, K., Kopf, J., Ahuja, N., Huang, J.B.: {DeepMVS}: Learning multi-view stereopsis. In: CVPR (2018)

\bibitem{im2019dpsnet}
Im, S., Jeon, H.G., Lin, S., Kweon, I.S.: {DPSNet}: End-to-end deep plane sweep stereo. ICLR  (2019)

\bibitem{izquierdo2023sfm}
Izquierdo, S., Civera, J.: {SfM-TTR}: Using structure from motion for test-time refinement of single-view depth networks. In: CVPR (2023)

\bibitem{InfiniTAM_ISMAR_2015}
{K{\"a}hler}, O., {Prisacariu}, V.A., {Ren}, C.Y., {Sun}, X., {Torr}, P.H.S., {Murray}, D.W.: {Very High Frame Rate Volumetric Integration of Depth Images on Mobile Device}. {IEEE Transactions on Visualization and Computer Graphics (Proceedings International Symposium on Mixed and Augmented Reality 2015}  \textbf{22}(11) (2015)

\bibitem{InfiniTAM_ECCV_2016}
K{\"{a}}hler, O., Prisacariu, V.A., Murray, D.W.: Real-time large-scale dense 3d reconstruction with loop closure. In: Computer Vision - {ECCV} 2016 - 14th European Conference, Amsterdam, The Netherlands, October 11-14, 2016, Proceedings, Part {VIII}. pp. 500--516 (2016)

\bibitem{kendall2017end}
Kendall, A., Martirosyan, H., Dasgupta, S., Henry, P., Kennedy, R., Bachrach, A., Bry, A.: End-to-end learning of geometry and context for deep stereo regression. In: ICCV (2017)

\bibitem{kerbl20233d}
Kerbl, B., Kopanas, G., Leimk{\"u}hler, T., Drettakis, G.: 3d gaussian splatting for real-time radiance field rendering. ACM Transactions on Graphics  \textbf{42}(4) (2023)

\bibitem{khan2023temporally}
Khan, N., Penner, E., Lanman, D., Xiao, L.: Temporally consistent online depth estimation using point-based fusion. In: CVPR (2023)

\bibitem{Kulhanek_2023_ICCV}
Kulhanek, J., Sattler, T.: {Tetra-NeRF}: Representing neural radiance fields using tetrahedra. In: ICCV (2023)

\bibitem{kuznietsov2021comoda}
Kuznietsov, Y., Proesmans, M., Van~Gool, L.: {CoMoDA}: Continuous monocular depth adaptation using past experiences. In: WACV (2021)

\bibitem{li2023neuralangelo}
Li, Z., M\"uller, T., Evans, A., Taylor, R.H., Unberath, M., Liu, M.Y., Lin, C.H.: Neuralangelo: High-fidelity neural surface reconstruction. In: CVPR (2023)

\bibitem{Raft-stereo-Multilevel-lipson-2021}
Lipson, L., Teed, Z., Deng, J.: Raft-stereo: Multilevel recurrent field transforms for stereo matching. In: 3DV (2021)

\bibitem{lorensen1998marching}
Lorensen, W.E., Cline, H.E.: Marching cubes: A high resolution {3D} surface construction algorithm. In: Seminal graphics: pioneering efforts that shaped the field (1998)

\bibitem{luo2020consistent}
Luo, X., Huang, J.B., Szeliski, R., Matzen, K., Kopf, J.: Consistent video depth estimation. In: ACM SIGGRAPH (2020)

\bibitem{ma2019self}
Ma, F., Cavalheiro, G.V., Karaman, S.: Self-supervised sparse-to-dense: Self-supervised depth completion from lidar and monocular camera. In: ICRA (2019)

\bibitem{ma2018sparse}
Ma, F., Karaman, S.: Sparse-to-dense: Depth prediction from sparse depth samples and a single image. In: ICRA (2018)

\bibitem{ma2022multiview}
Ma, Z., Teed, Z., Deng, J.: Multiview stereo with cascaded epipolar {RAFT}. In: ECCV (2022)

\bibitem{mccraith2020monocular}
McCraith, R., Neumann, L., Zisserman, A., Vedaldi, A.: Monocular depth estimation with self-supervised instance adaptation. arXiv:2004.05821  (2020)

\bibitem{Menze2015CVPR}
Menze, M., Geiger, A.: Object scene flow for autonomous vehicles. In: CVPR (2015)

\bibitem{mildenhall2020nerf}
Mildenhall, B., Srinivasan, P.P., Tancik, M., Barron, J.T., Ramamoorthi, R., Ng, R.: {NeRF}: Representing scenes as neural radiance fields for view synthesis. In: ECCV (2020)

\bibitem{murez2020atlas}
Murez, Z., van As, T., Bartolozzi, J., Sinha, A., Badrinarayanan, V., Rabinovich, A.: Atlas: End-to-end {3D} scene reconstruction from posed images. In: ECCV (2020)

\bibitem{peng2020convolutional}
Peng, S., Niemeyer, M., Mescheder, L., Pollefeys, M., Geiger, A.: Convolutional occupancy networks. In: ECCV (2020)

\bibitem{poggi2022multi}
Poggi, M., Conti, A., Mattoccia, S.: Multi-view guided multi-view stereo. In: IROS (2022)

\bibitem{rakotosaona2023nerfmeshing}
Rakotosaona, M.J., Manhardt, F., Arroyo, D.M., Niemeyer, M., Kundu, A., Tombari, F.: {NeRFMeshing}: Distilling neural radiance fields into geometrically-accurate {3D} meshes. In: 3DV (2023)

\bibitem{ravi2020pytorch3d}
Ravi, N., Reizenstein, J., Novotny, D., Gordon, T., Lo, W.Y., Johnson, J., Gkioxari, G.: Accelerating {3D} deep learning with {PyTorch3D}. arXiv:2007.08501  (2020)

\bibitem{rich20213dvnet}
Rich, A., Stier, N., Sen, P., H\"ollerer, T.: {3DVNet}: Multi-view depth prediction and volumetric refinement. In: 3DV (2021)

\bibitem{roessle2022dense}
Roessle, B., Barron, J.T., Mildenhall, B., Srinivasan, P.P., Nie{\ss}ner, M.: Dense depth priors for neural radiance fields from sparse input views. In: CVPR (2022)

\bibitem{sayed2022simplerecon}
Sayed, M., Gibson, J., Watson, J., Prisacariu, V., Firman, M., Godard, C.: {SimpleRecon}: {3D} reconstruction without {3D} convolutions. In: ECCV (2022)

\bibitem{schonberger2016pixelwise}
Sch{\"o}nberger, J.L., Zheng, E., Frahm, J.M., Pollefeys, M.: Pixelwise view selection for unstructured multi-view stereo. In: ECCV (2016)

\bibitem{schoenberger2016sfm}
Sch\"{o}nberger, J.L., Frahm, J.M.: Structure-from-motion revisited. In: CVPR (2016)

\bibitem{shotton2013scene}
Shotton, J., Glocker, B., Zach, C., Izadi, S., Criminisi, A., Fitzgibbon, A.: Scene coordinate regression forests for camera relocalization in {RGB-D} images. In: CVPR (2013)

\bibitem{shu2020feature}
Shu, C., Yu, K., Duan, Z., Yang, K.: Feature-metric loss for self-supervised learning of depth and egomotion. In: ECCV (2020)

\bibitem{sinha2020deltas}
Sinha, A., Murez, Z., Bartolozzi, J., Badrinarayanan, V., Rabinovich, A.: {DELTAS}: Depth estimation by learning triangulation and densification of sparse points. In: ECCV (2020)

\bibitem{song2023prior}
Song, S., Truong, K.G., Kim, D., Jo, S.: Prior depth-based multi-view stereo network for online {3D} model reconstruction. Pattern Recognition  (2023)

\bibitem{Stier_2023_ICCV}
Stier, N., Ranjan, A., Colburn, A., Yan, Y., Yang, L., Ma, F., Angles, B.: Finerecon: Depth-aware feed-forward network for detailed {3D} reconstruction. In: ICCV (2023)

\bibitem{stier2021vortx}
Stier, N., Rich, A., Sen, P., H{\"o}llerer, T.: {VoRTX}: Volumetric {3D} reconstruction with transformers for voxelwise view selection and fusion. In: 3DV (2021)

\bibitem{sun2021neuralrecon}
Sun, J., Xie, Y., Chen, L., Zhou, X., Bao, H.: {NeuralRecon}: Real-time coherent {3D} reconstruction from monocular video. In: CVPR (2021)

\bibitem{tan2021efficientnetv2}
Tan, M., Le, Q.: Efficientnetv2: Smaller models and faster training. In: ICML (2021)

\bibitem{uhrig2017sparsity}
Uhrig, J., Schneider, N., Schneider, L., Franke, U., Brox, T., Geiger, A.: Sparsity invariant {CNNs}. In: 3DV (2017)

\bibitem{scade-nerfs-from-Uy-2023}
Uy, M.A., Martin-Brualla, R., Guibas, L., Li, K.: {SCADE}: {NeRFs} from space carving with ambiguity-aware depth estimates. In: CVPR (2023)

\bibitem{valentin2018depth}
Valentin, J., Kowdle, A., Barron, J.T., Wadhwa, N., Dzitsiuk, M., Schoenberg, M., Verma, V., Csaszar, A., Turner, E., Dryanovski, I., et~al.: Depth from motion for smartphone {AR}. Transactions on Graphics  (2018)

\bibitem{Wald2019RIO}
Wald, J., Avetisyan, A., Navab, N., Tombari, F., Niessner, M.: {RIO}: {3D} object instance re-localization in changing indoor environments. In: ICCV (2019)

\bibitem{wang2018mvdepthnet}
Wang, K., Shen, S.: {MVDepthNet}: Real-time multiview depth estimation neural network. In: 3DV (2018)

\bibitem{Depth-Completion-using-Du-2022}
Wenchao~Du, Hu~Chen, H.Y., zhang, Y.: Depth completion using geometry-aware embedding. In: ICRA (2022)

\bibitem{wong2021unsupervised}
Wong, A., Soatto, S.: Unsupervised depth completion with calibrated backprojection layers. In: ICCV (2021)

\bibitem{Xin2023ISMAR}
Xin, Y., Zuo, X., Lu, D., Leutenegger, S.: {SimpleMapping: Real-Time Visual-Inertial Dense Mapping with Deep Multi-View Stereo}. In: ISMAR (2023)

\bibitem{yang2020cost}
Yang, J., Mao, W., Alvarez, J.M., Liu, M.: Cost volume pyramid based depth inference for multi-view stereo. In: CVPR (2020)

\bibitem{yao2018mvsnet}
Yao, Y., Luo, Z., Li, S., Fang, T., Quan, L.: {MVSNet}: Depth inference for unstructured multi-view stereo. In: ECCV (2018)

\bibitem{yariv2021volume}
Yariv, L., Gu, J., Kasten, Y., Lipman, Y.: Volume rendering of neural implicit surfaces. In: NeurIPS (2021)

\bibitem{yu2022monosdf}
Yu, Z., Peng, S., Niemeyer, M., Sattler, T., Geiger, A.: {MonoSDF}: Exploring monocular geometric cues for neural implicit surface reconstruction. NeurIPS  (2022)

\bibitem{Zhang2019GANet}
Zhang, F., Prisacariu, V., Yang, R., Torr, P.H.: {GA-Net}: Guided aggregation net for end-to-end stereo matching. In: CVPR (2019)

\bibitem{zhang2019domaininvariant}
Zhang, F., Qi, X., Yang, R., Prisacariu, V., Wah, B., Torr, P.: Domain-invariant stereo matching networks. In: ECCV (2020)

\bibitem{zhang2023geomvsnet}
Zhang, Z., Peng, R., Hu, Y., Wang, R.: {GeoMVSNet}: Learning multi-view stereo with geometry perception. In: CVPR (2023)

\bibitem{zhou2018unet++}
Zhou, Z., Rahman~Siddiquee, M.M., Tajbakhsh, N., Liang, J.: {UNet++}: A nested {U-Net} architecture for medical image segmentation. In: Deep learning in medical image analysis and multimodal learning for clinical decision support (2018)

\bibitem{zuo2023incremental}
Zuo, X., Yang, N., Merrill, N., Xu, B., Leutenegger, S.: Incremental dense reconstruction from monocular video with guided sparse feature volume fusion. IEEE Robotics and Automation Letters  (2023)

\end{thebibliography}
\end{document}